\documentclass[10pt,twocolumn,letterpaper]{article}

\usepackage{iccv}
\usepackage{times}
\usepackage{epsfig}
\usepackage{graphicx}
\usepackage{amsmath}
\usepackage{amssymb}
\usepackage{csquotes}

\usepackage{algorithm}
\usepackage[noend]{algpseudocode}
\usepackage[normalem]{ulem}

\usepackage[pagebackref=true,breaklinks=true,letterpaper=true,colorlinks,bookmarks=false]{hyperref}

\iccvfinalcopy 

\def\iccvPaperID{3359} 

\ificcvfinal\fi

\begin{document}

\title{Confidence-Aware Learning for Camouflaged Object Detection}

\author{Jiawei Liu\\
ANU\\
{\tt\small jiawei.liu3@anu.edu.au}
\and
Jing Zhang\\
ANU\\
{\tt\small jing.zhang@anu.edu.au}

\and
Nick Barnes\\
ANU\\
{\tt\small nick.barnes@anu.edu.au}
}

\maketitle
\ificcvfinal\thispagestyle{empty}\fi

\begin{abstract}
Confidence-aware learning is proven as an effective solution to prevent 
networks becoming overconfident.
We present a confidence-aware camouflaged object detection framework using dynamic supervision to produce both accurate camouflage map and meaningful \enquote{confidence} representing model awareness about the current prediction. A camouflaged object detection network is designed to produce our camouflage prediction. Then, we concatenate it with the input image and feed it to the confidence estimation network to produce an one channel confidence map.
We generate dynamic supervision for the confidence estimation network, representing the
agreement of camouflage prediction with the ground truth camouflage map. With the produced confidence map, we introduce confidence-aware learning with the confidence map as guidance to pay more attention to the hard/low-confidence pixels in the loss function.
We claim that, once trained, our confidence estimation network can evaluate pixel-wise accuracy of the prediction without relying on the ground truth camouflage map.
Extensive results on four camouflaged object detection testing datasets illustrate the superior performance of the proposed model in explaining the camouflage prediction.
   
\end{abstract}

\section{Introduction}
Camouflaged is defined as a state where the object has disguised appearance that is indiscernible from its surroundings. Camouflage is a widely applied technique in the world of animals to conceal themselves, deceiving predators into making false judgements. This is achieved through various camouflage techniques, \eg disruptive colouration, self-decoration, cryptic behaviour,
\etc
\cite{barbosa2008cuttlefish, cott1940adaptive, forbes2011dazzled}. This natural occurrence also inspires the development of artificial camouflage, such as military camouflage \cite{barbosa2008cuttlefish}.



\begin{figure}[tp]
  \begin{center}
  \begin{tabular}{{c@{ } c@{ } c@{ } c@{ } c@{ }}}
  {\includegraphics[width=0.185\linewidth]{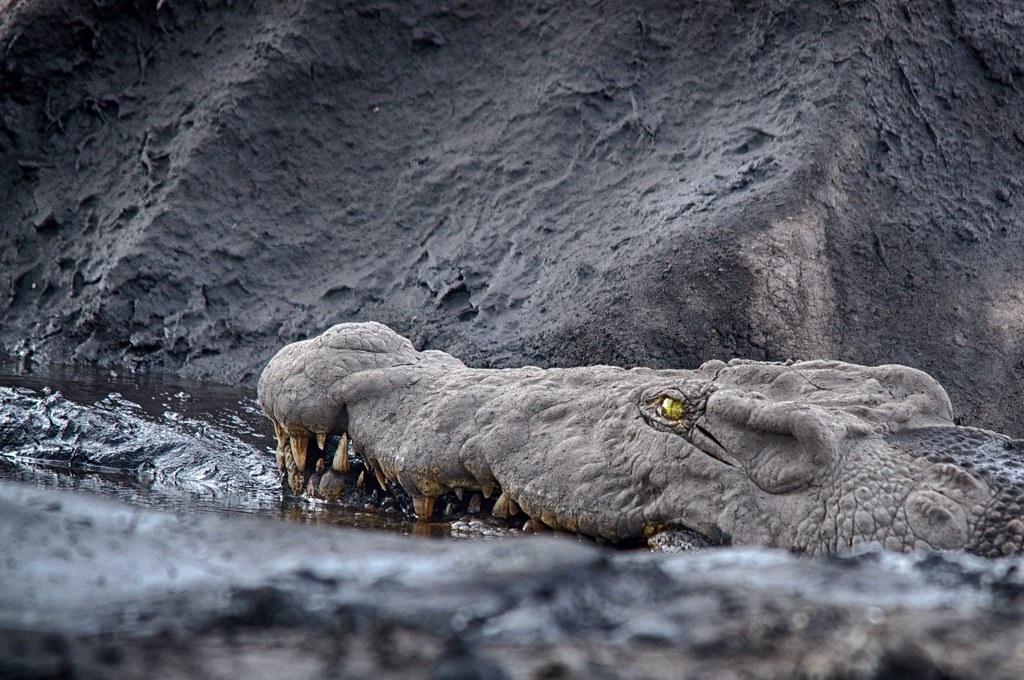}}&
     {\includegraphics[width=0.185\linewidth]{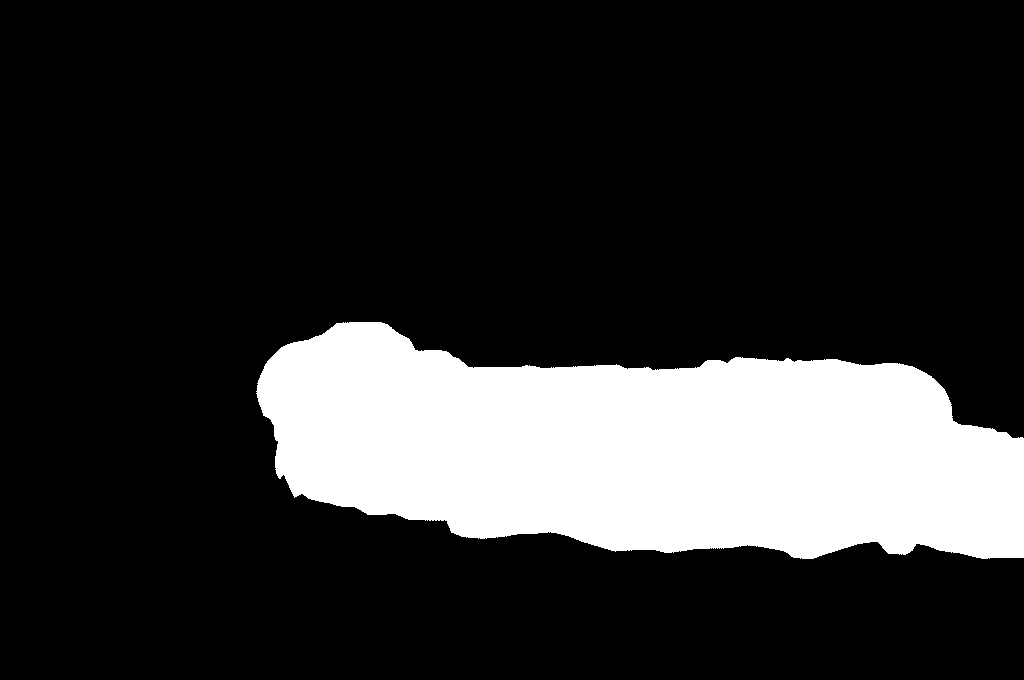}}&
     {\includegraphics[width=0.185\linewidth]{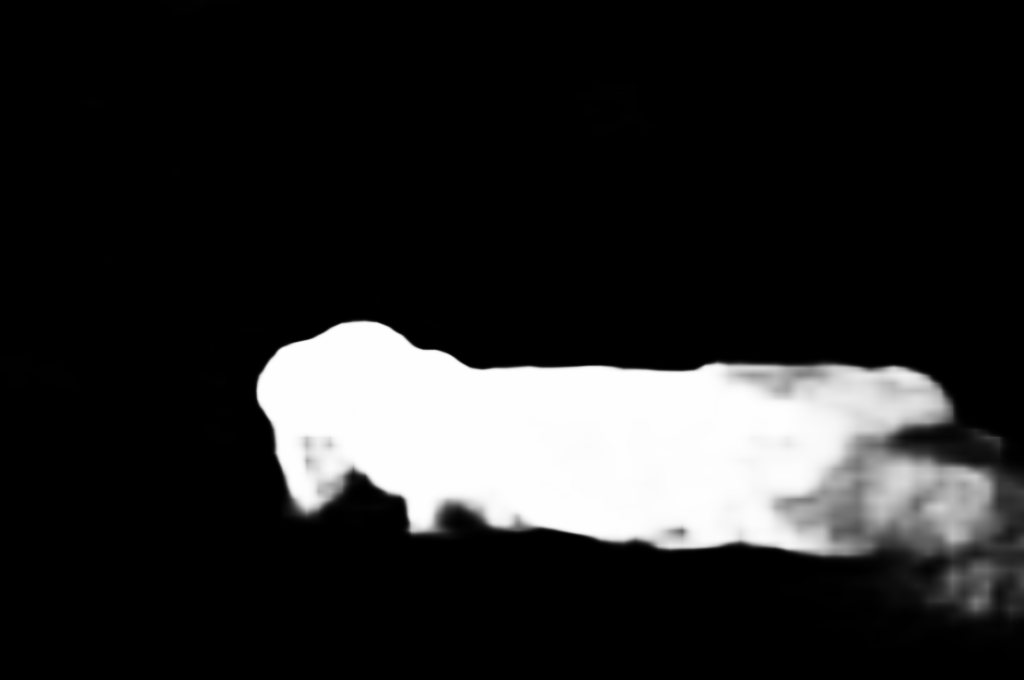}}&
     {\includegraphics[width=0.185\linewidth]{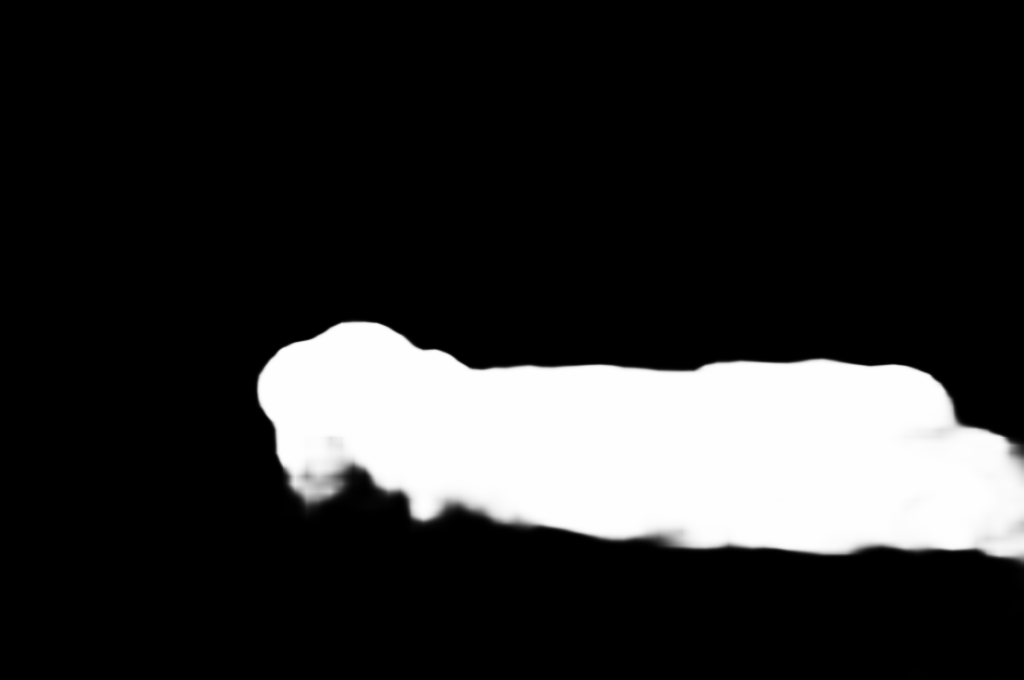}}&
     {\includegraphics[width=0.185\linewidth]{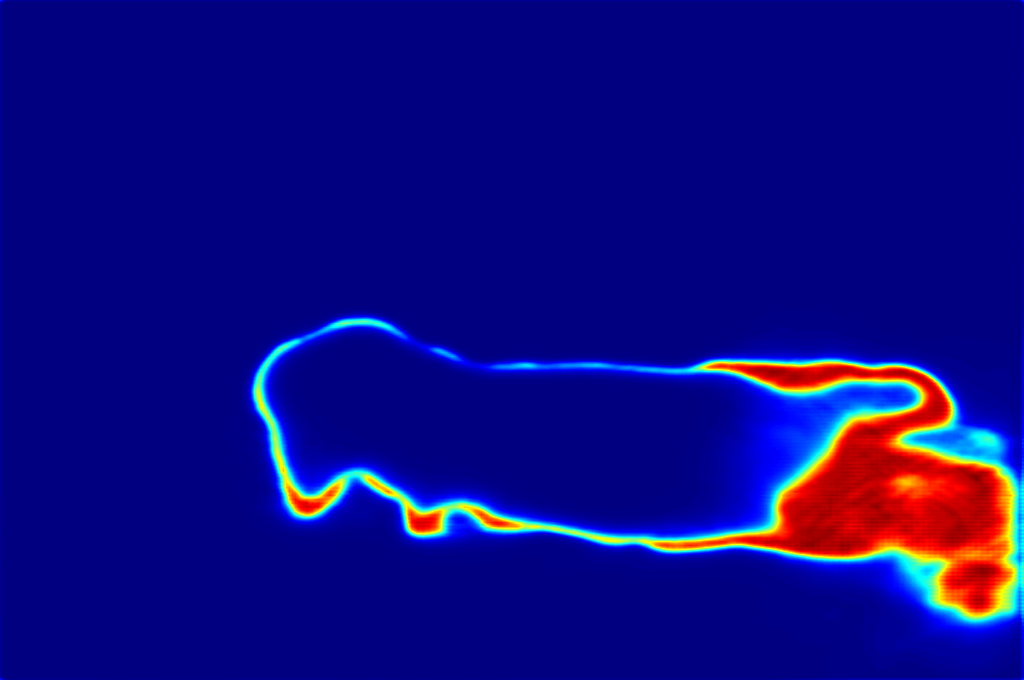}} \\
  {\includegraphics[width=0.185\linewidth]{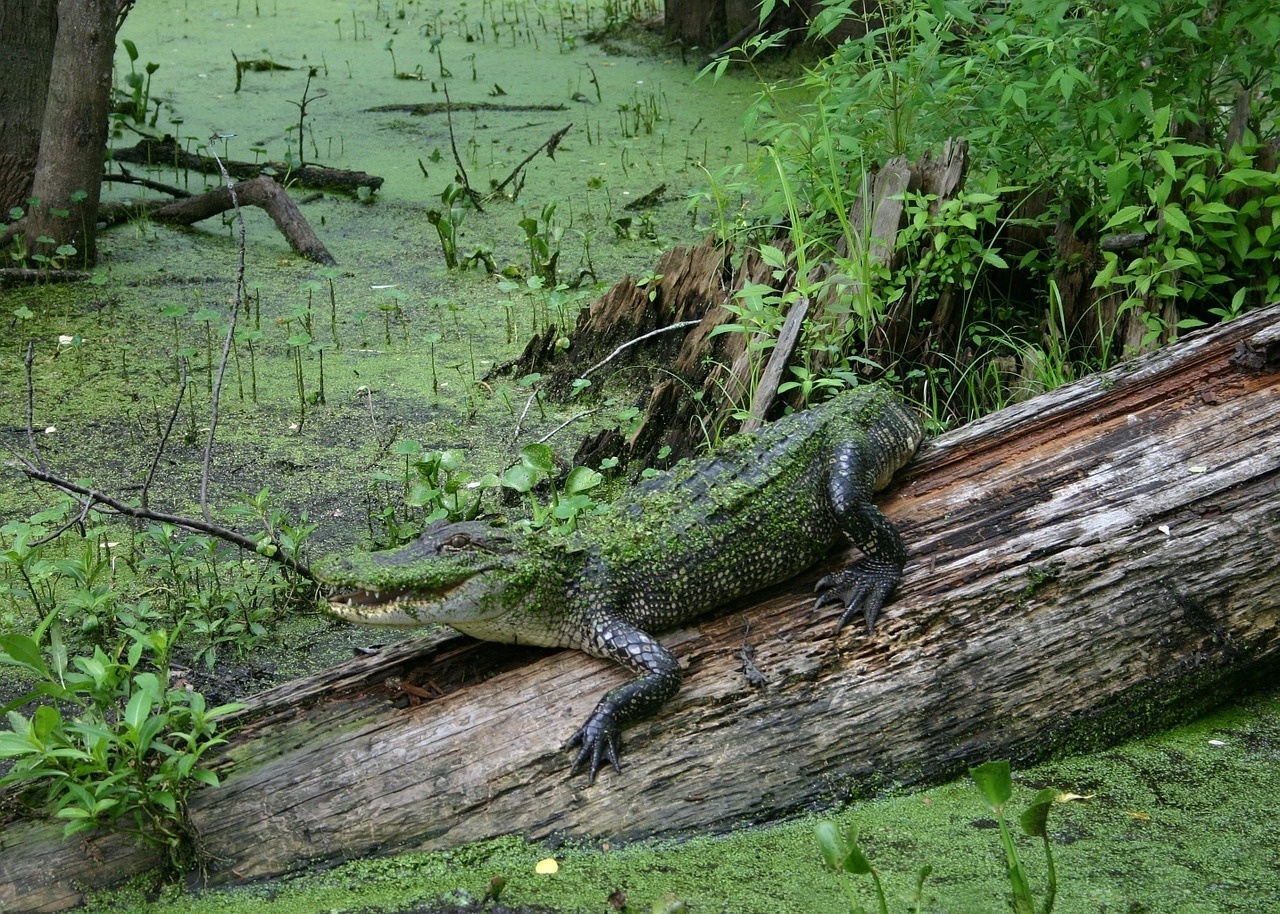}}&
     {\includegraphics[width=0.185\linewidth]{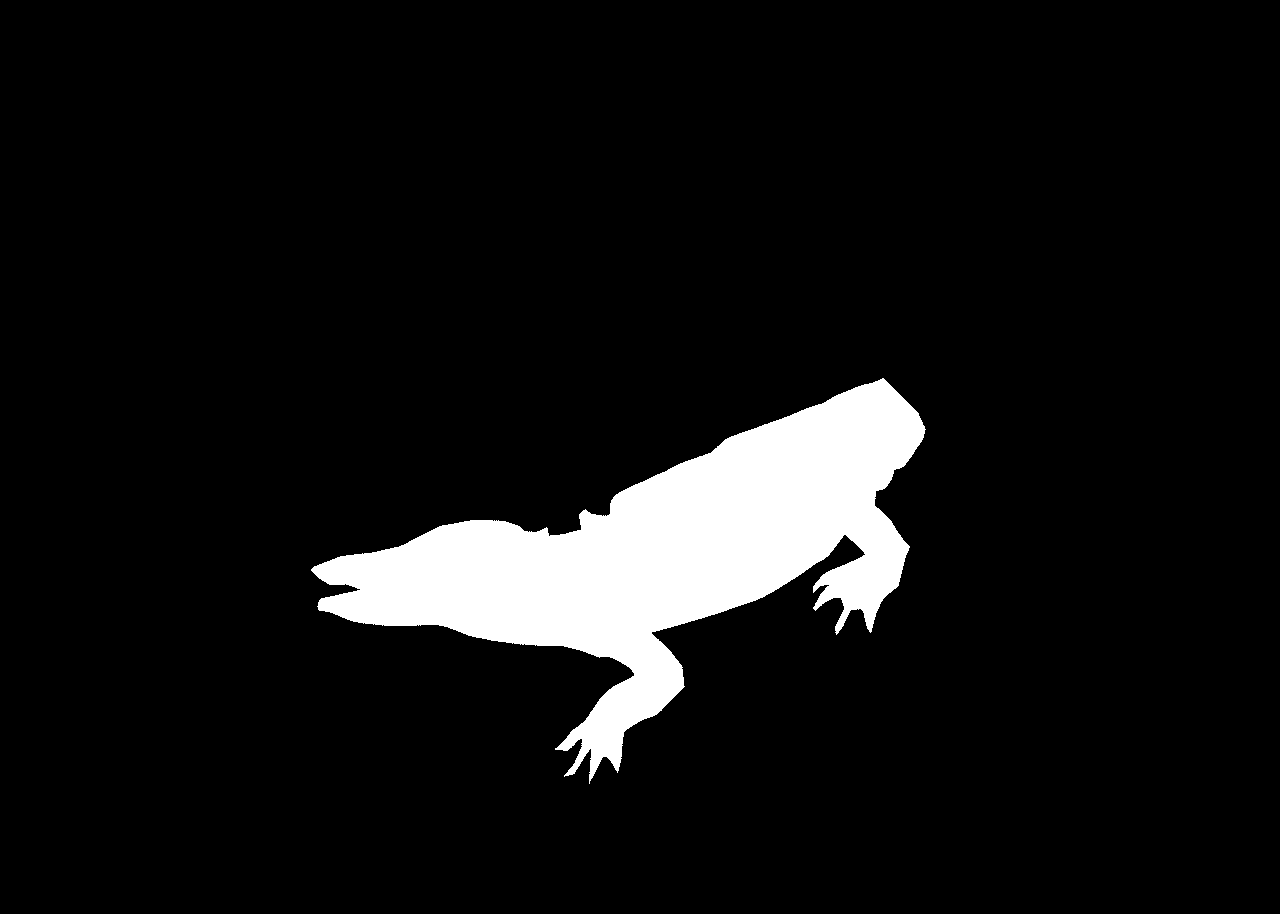}}&
     {\includegraphics[width=0.185\linewidth]{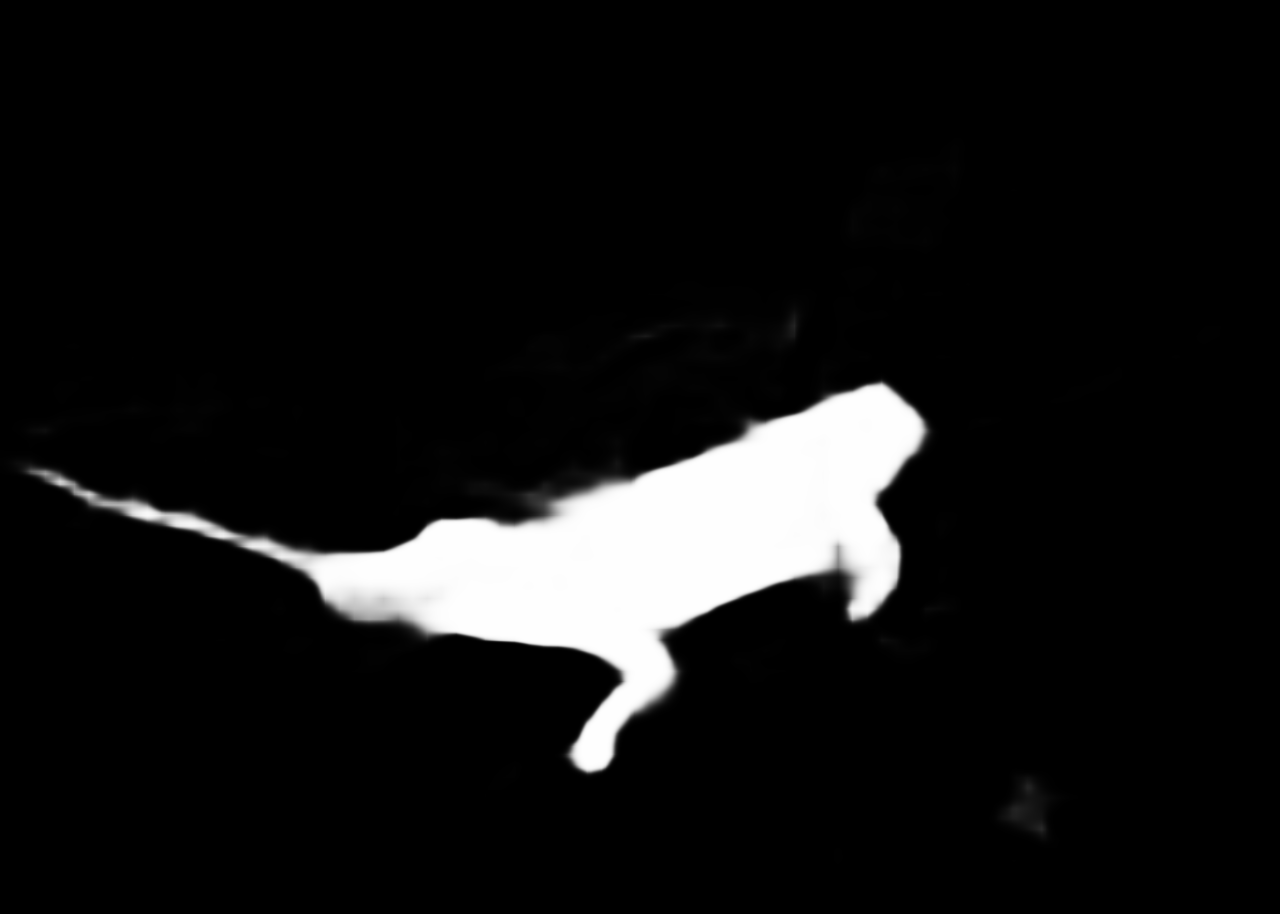}}&
     {\includegraphics[width=0.185\linewidth]{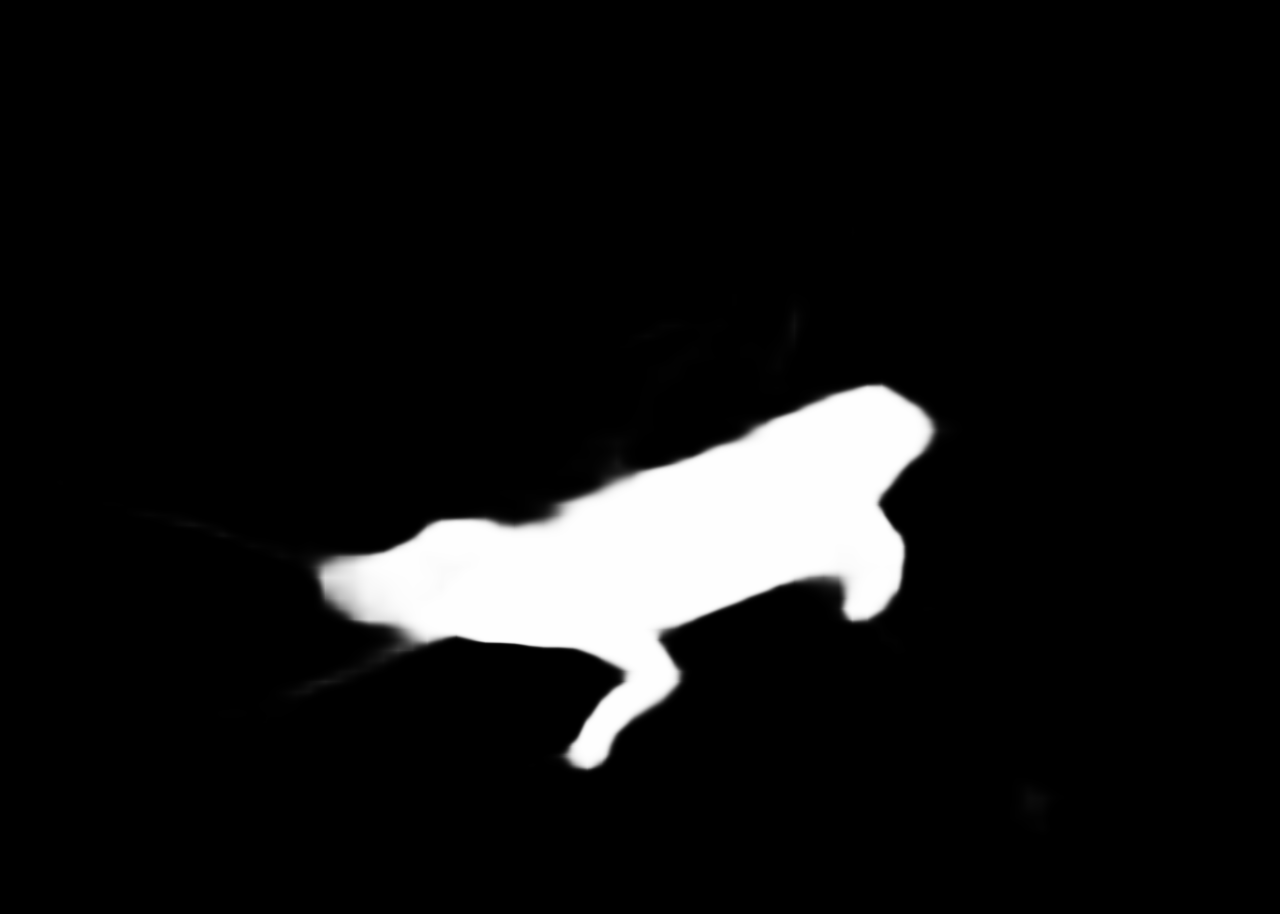}}&
     {\includegraphics[width=0.185\linewidth]{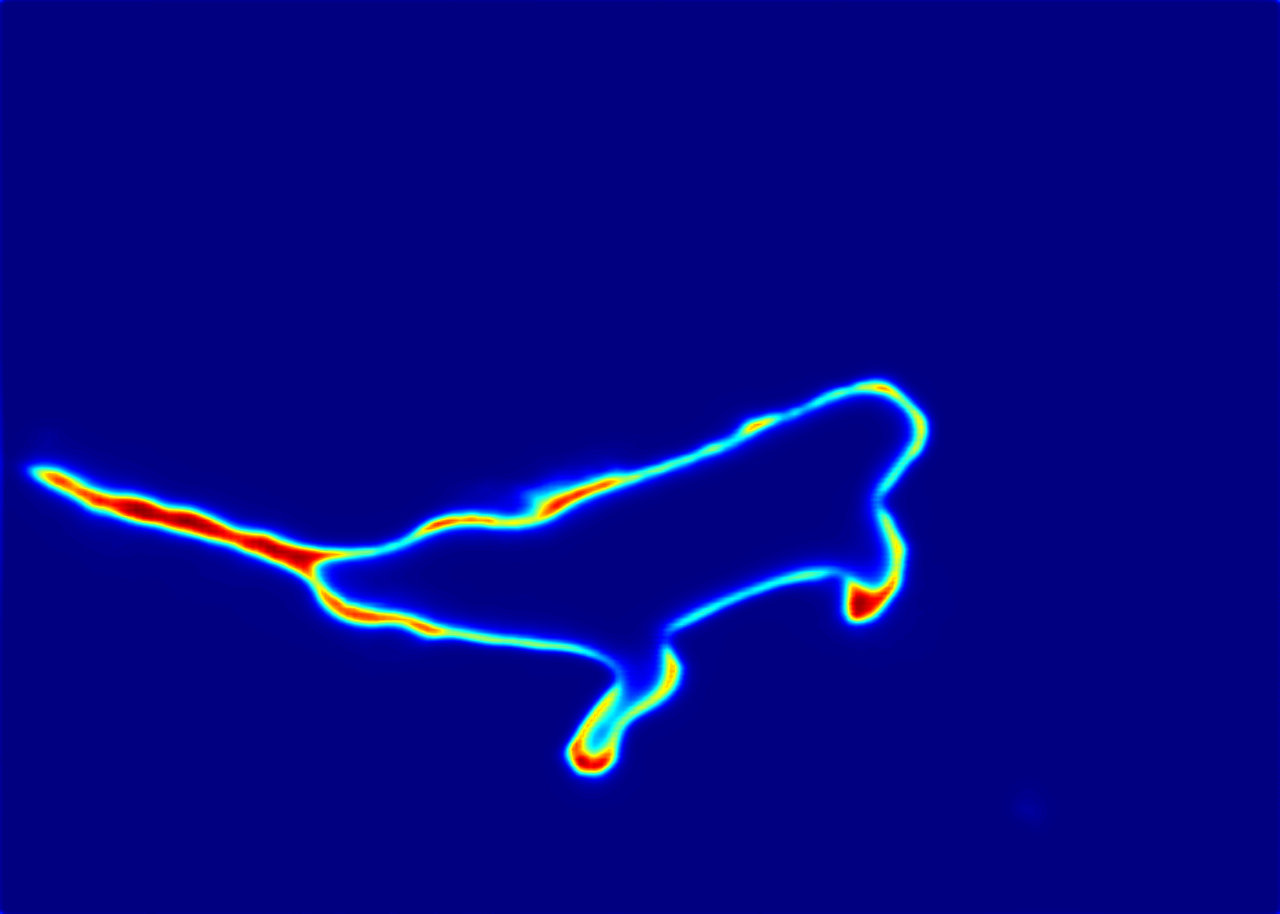}} \\
  {\includegraphics[width=0.185\linewidth]{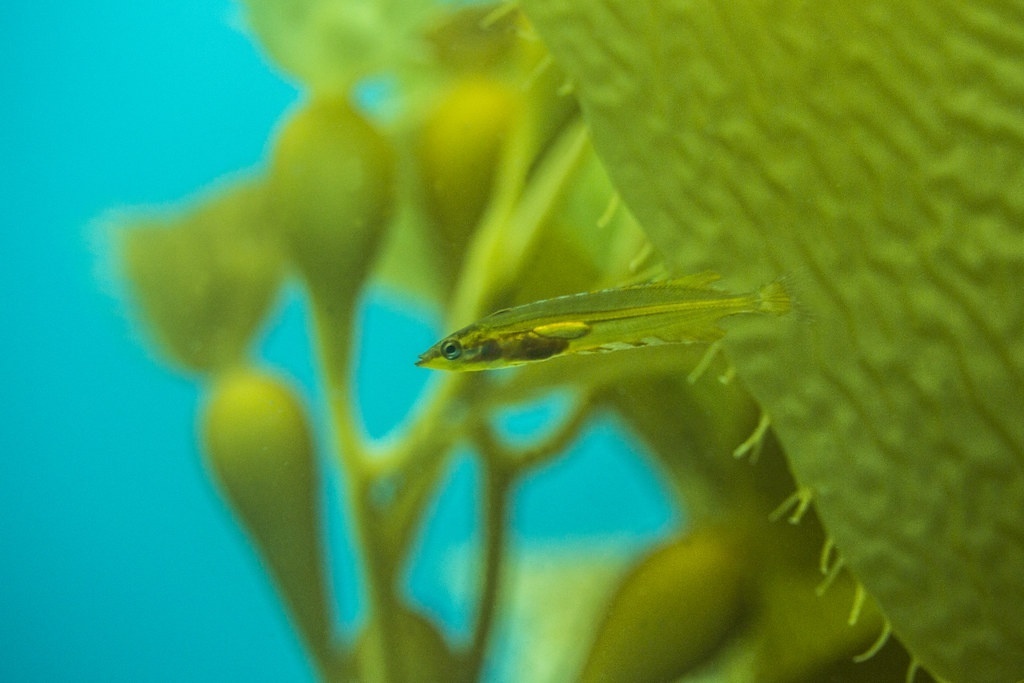}}&
     {\includegraphics[width=0.185\linewidth]{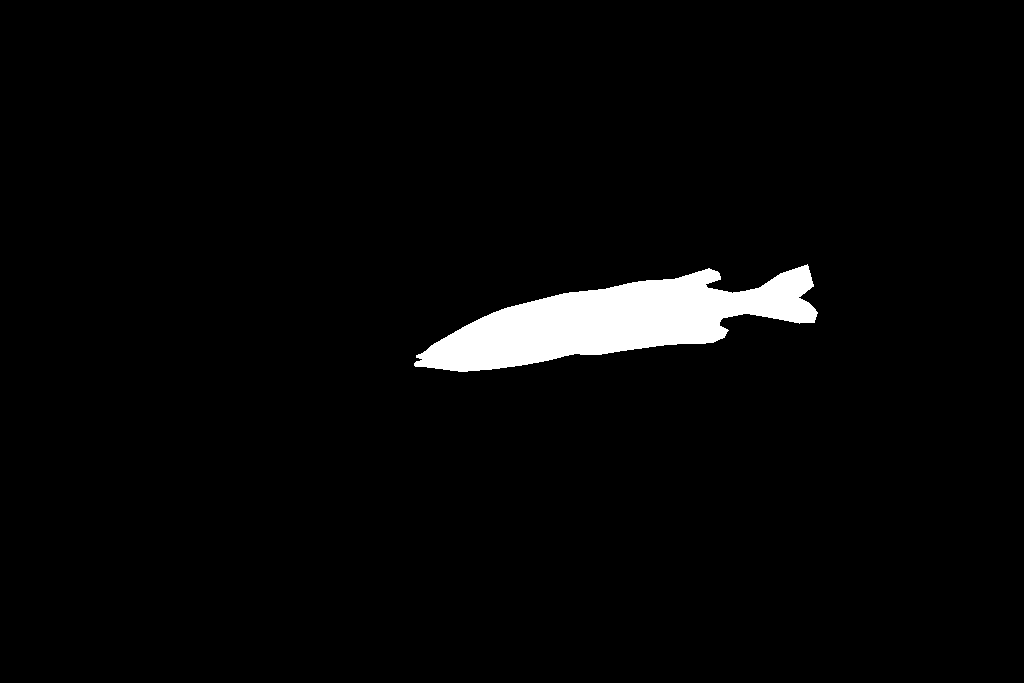}}&
     {\includegraphics[width=0.185\linewidth]{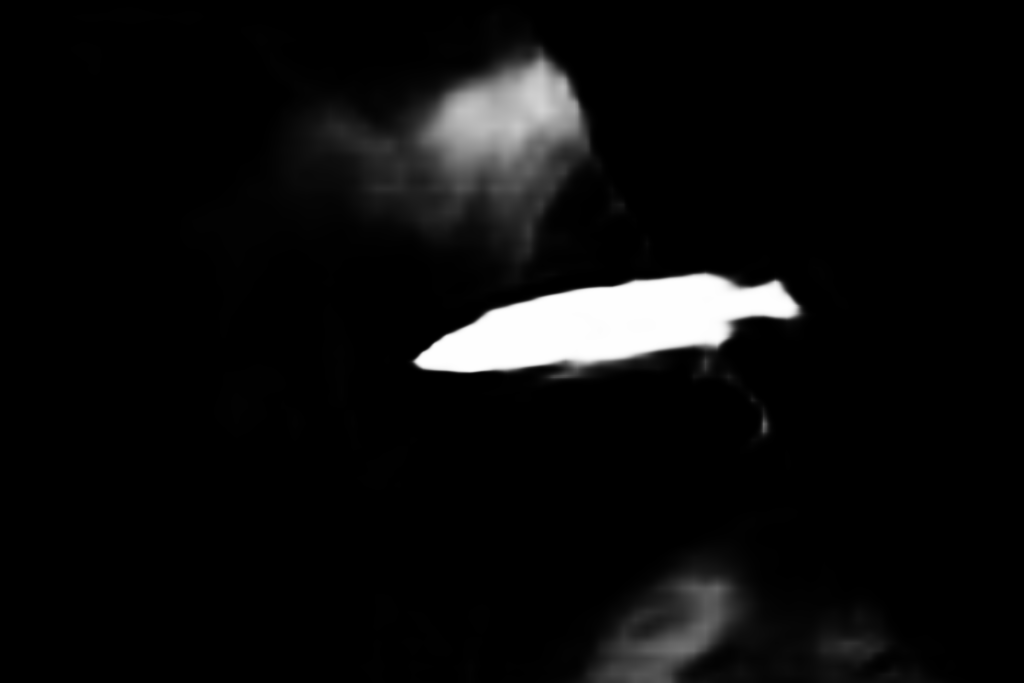}}&
     {\includegraphics[width=0.185\linewidth]{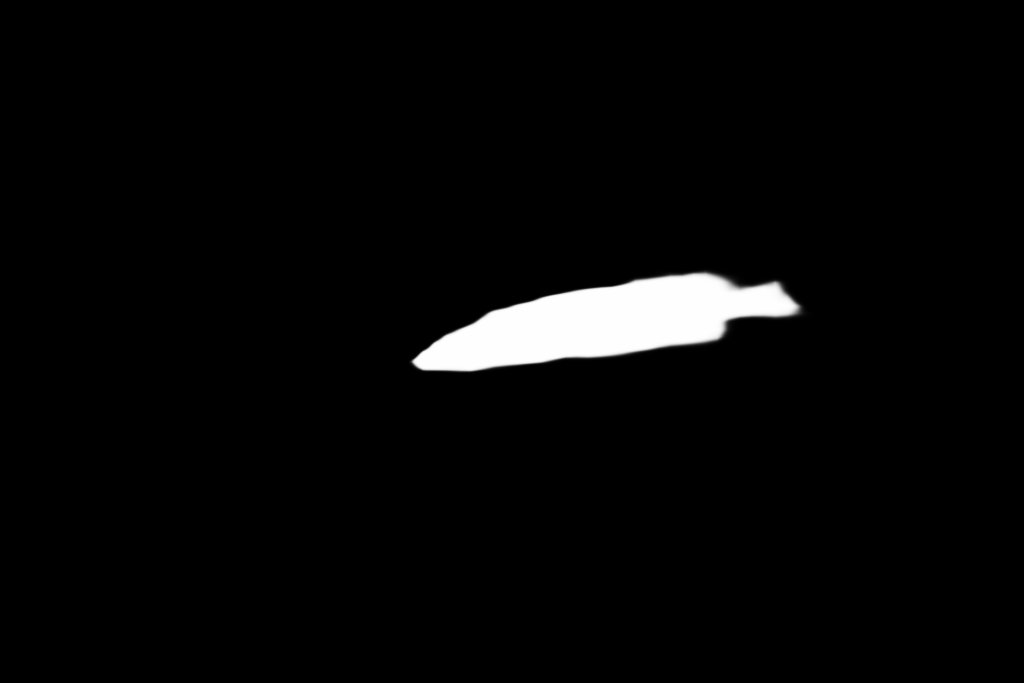}}&
     {\includegraphics[width=0.185\linewidth]{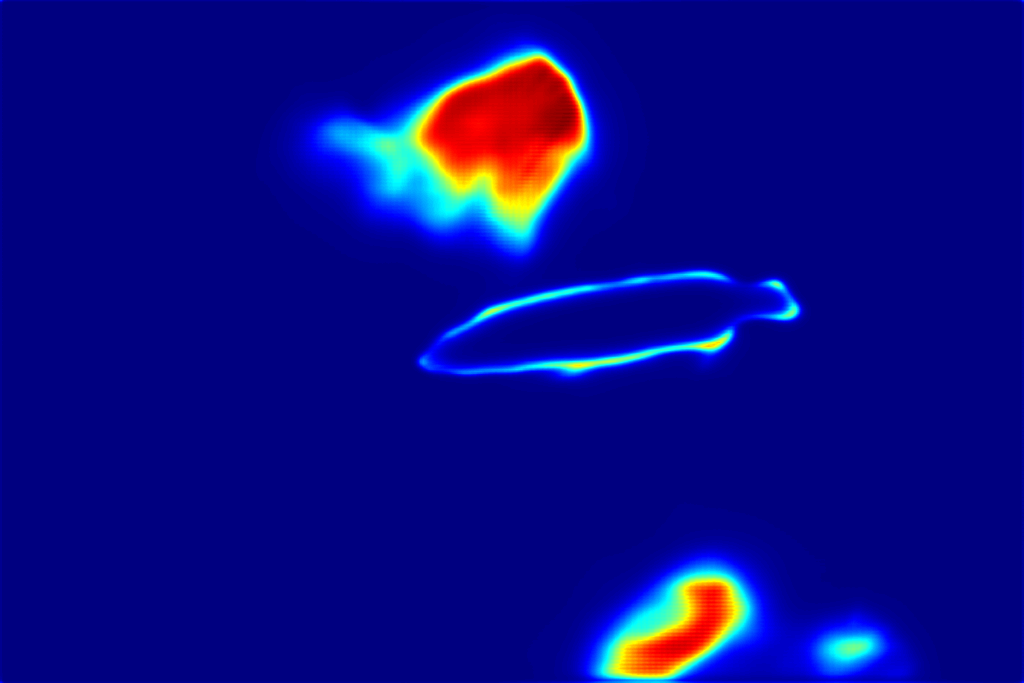}} \\
  {\includegraphics[width=0.185\linewidth]{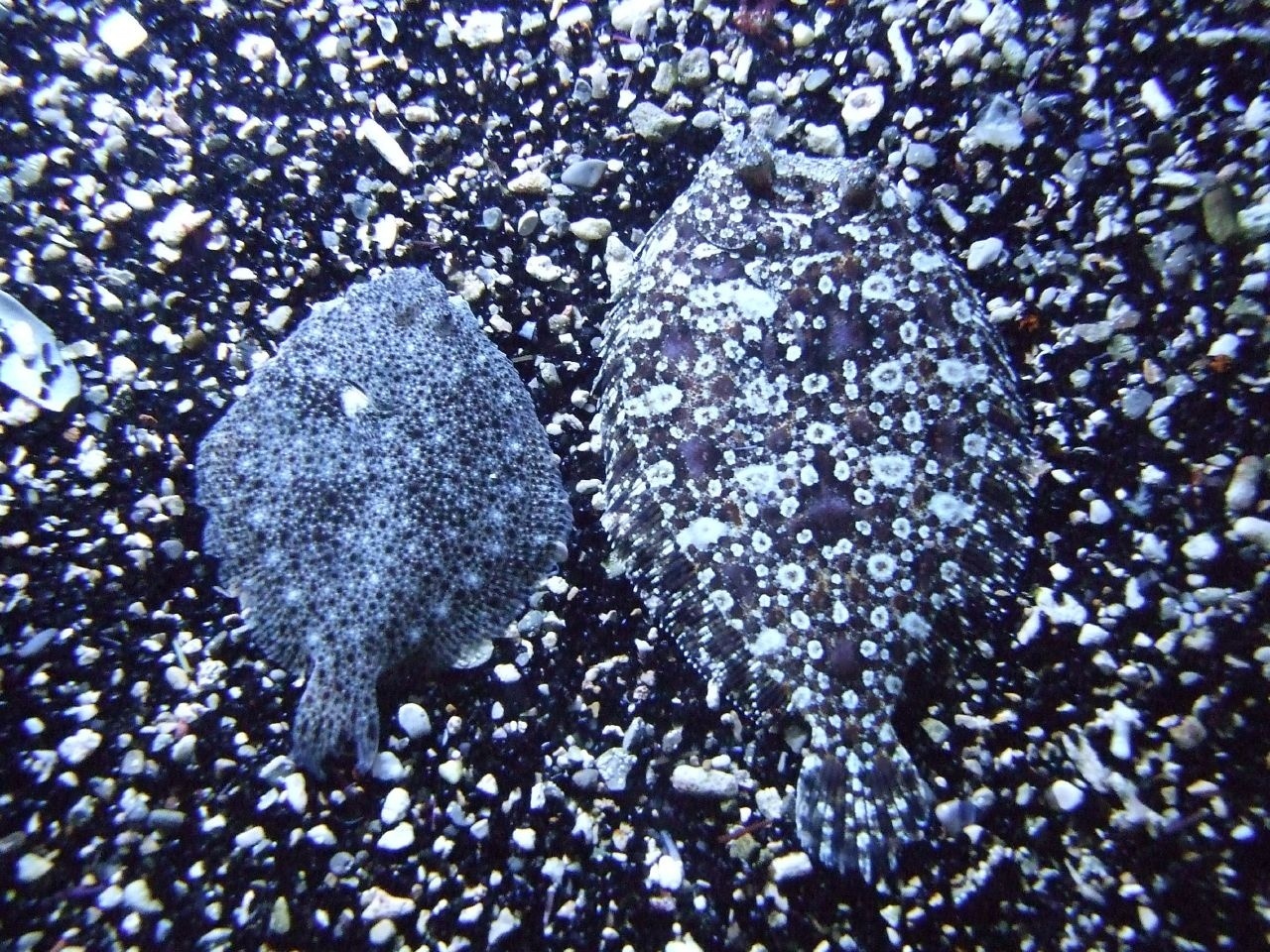}}&
     {\includegraphics[width=0.185\linewidth]{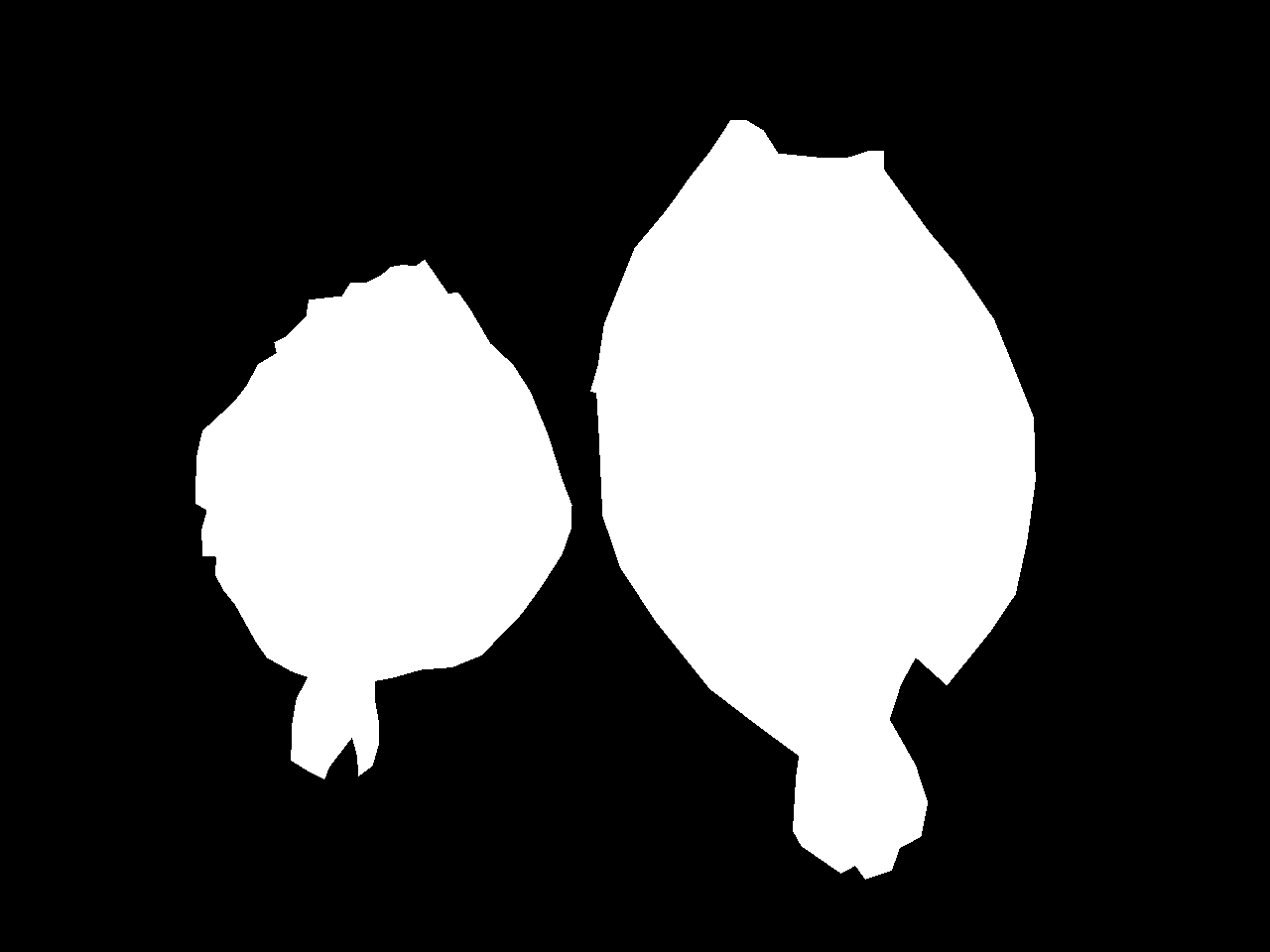}}&
     {\includegraphics[width=0.185\linewidth]{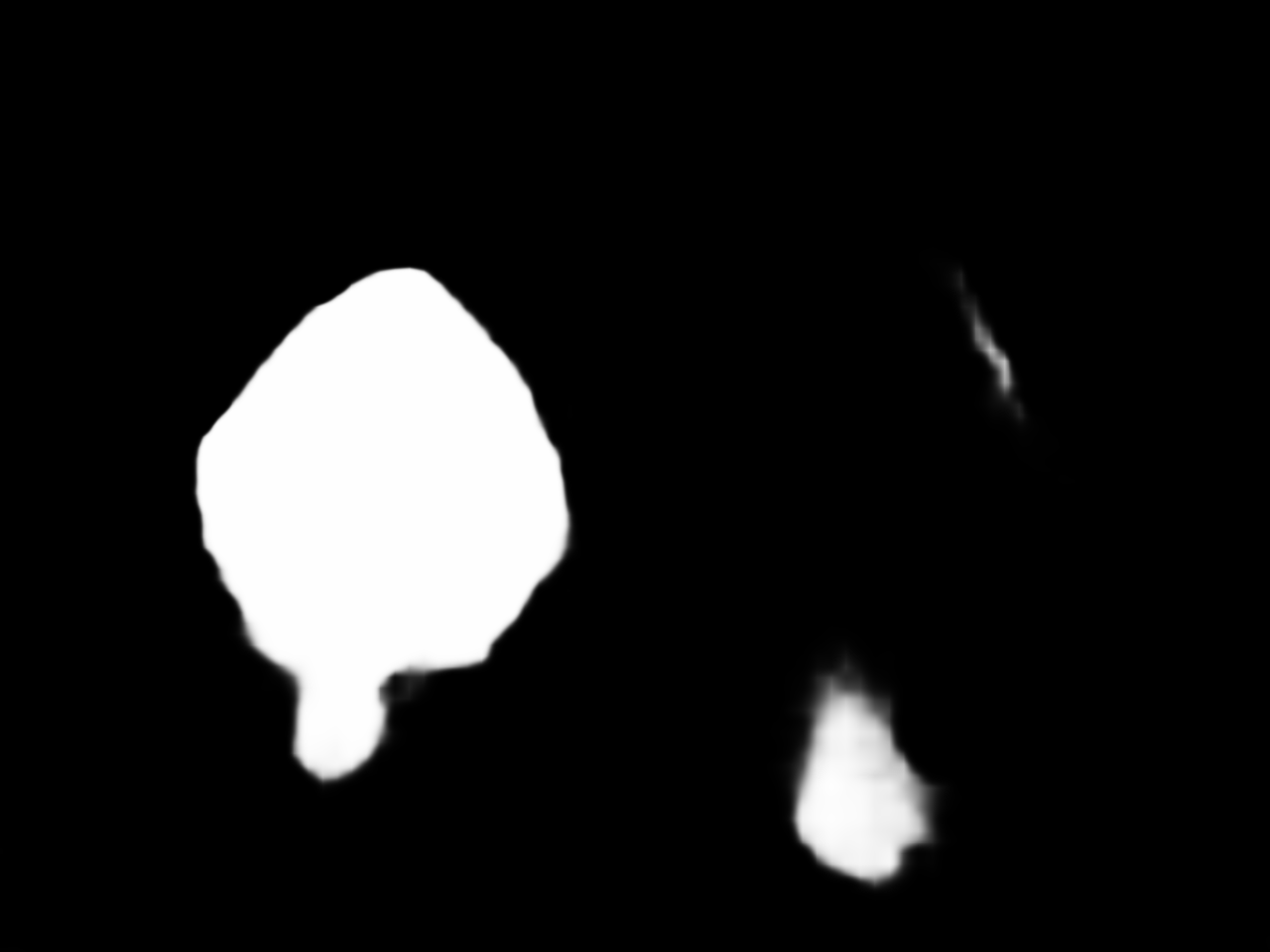}}&
     {\includegraphics[width=0.185\linewidth]{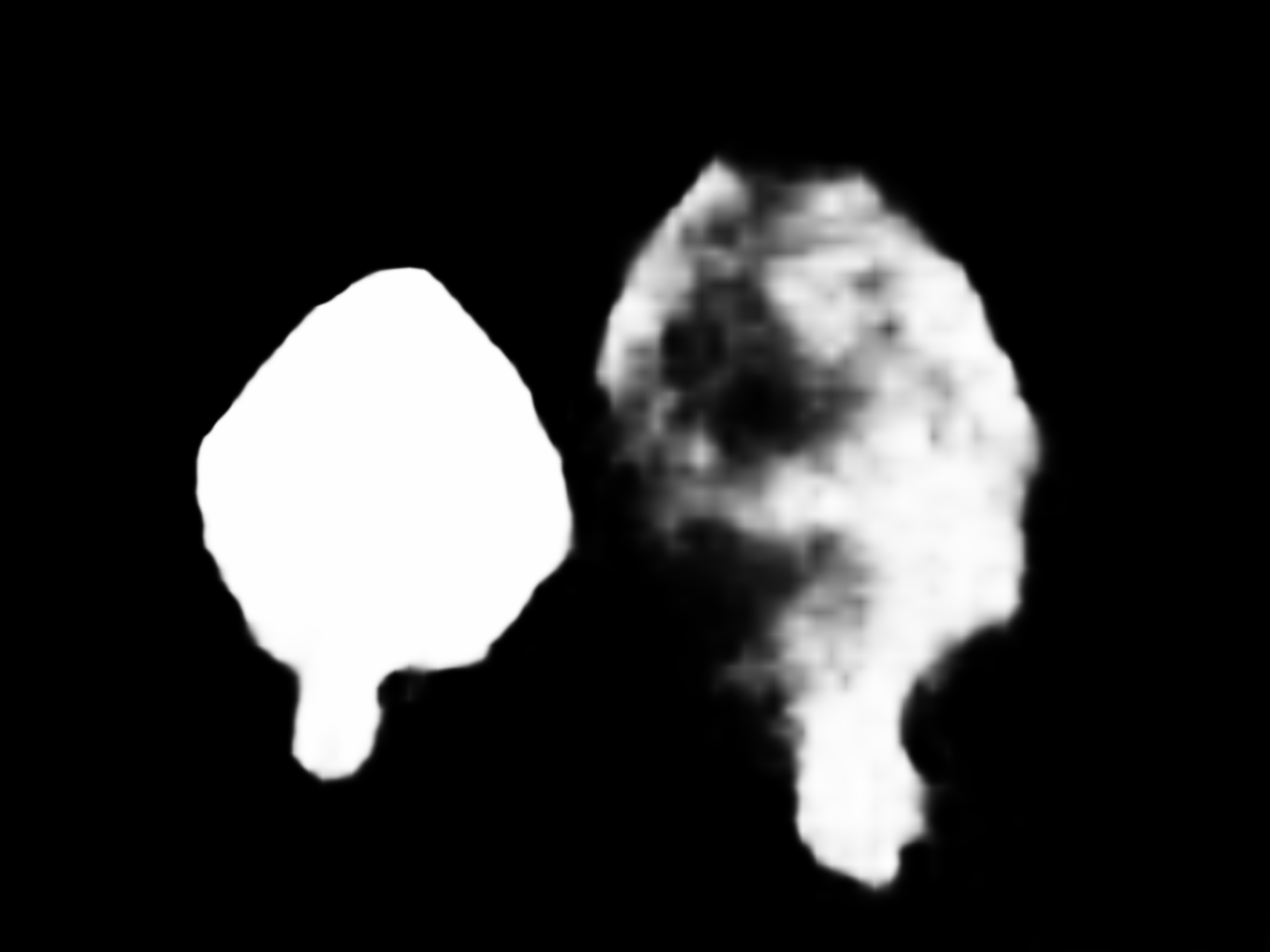}}&
     {\includegraphics[width=0.185\linewidth]{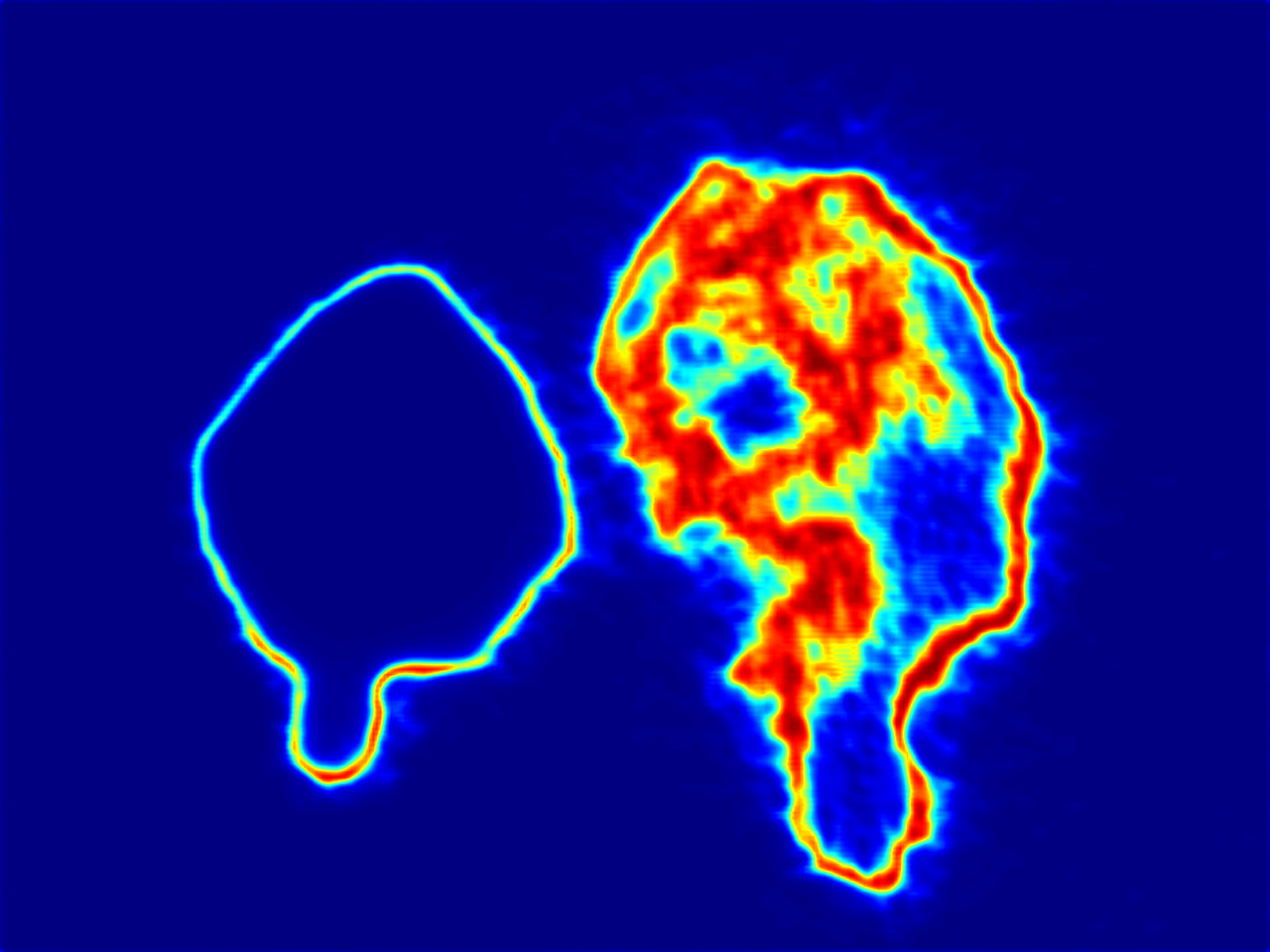}} \\
     \footnotesize{Image} &
     \footnotesize{GT} & \footnotesize{P w/o confi}& \footnotesize{P w/ confi} &\footnotesize{Confidence} \\
  \end{tabular}
  \end{center}
    \caption{Prediction of our confidence-aware COD network. Red indicates low confidence and blue indicates high confidence. From left to right: input image, ground truth, prediction without confidence as guidance, and with confidence as guidance, and confidence map. For easy samples in the first and second rows, the low-confidence regions mainly distribute along object boundaries.
    For hard samples in the third and fourth rows, our confidence map can effectively identify the false-positive regions (third row) leading to their removal and true-negative regions (fourth row) greatly improving coverage of the second object which is difficult to see.
    } 
    \label{fig: introduction figure}
\end{figure}


To conceal themselves, prey often
evolve to have a similar appearance to their surroundings, which makes camouflaged object detection a difficult task due to the similar foreground/background appearance or structure information. Conventionally, camouflaged object detection (COD) is defined as a binary segmentation task \cite{yan2020mirrornet,le2019anabranch,fan2020camouflaged}, where models learn to achieve mapping from the input image to a corresponding ground-truth camouflage map, showing the entire scope of the camouflaged objects.
We argue that the difficulty in detecting each part of a camouflaged object varies;
it is often much more difficult to locate the 
boundary between the camouflaged objects and their surroundings. 

A good deal of 
research has 
been done
to produce a difficulty-aware model.
They usually adopt Bayesian neural network approaches \cite{neal2012bayesian, welling2011bayesian, ma2015complete, gong2018meta, chen2014stochastic, wu2018deterministic, louizos2017multiplicative, blundell2015weight}, sampling based methods \cite{gal2016dropout, lakshminarayanan2016simple, zhang2020uncertainty, ayhan2018test} or deep neural network based approaches \cite{corbiere2019addressing, schonfeld2020u} to indicate model belief in its prediction.
Among them,
\cite{nie2019difficulty} learns a discriminator to distinguish between prediction and ground truth, where the output of the discriminator is defined as confidence map.
\cite{moon2020confidence, ding2020uncertainty} optimise the confidence estimation network with a ranking loss that assigns higher uncertainties to wrongly predicted samples or pixels. Specifically, \cite{moon2020confidence} develops a correctness ranking loss that enforces that samples with higher accuracy to have higher confidence, and \cite{ding2020uncertainty} proposes a loss function that maximises the difference between the estimated confidences of correct predictions and wrong predictions.
\cite{wannenwetsch2020probabilistic} presents a multi-task learning loss function derived by maximising the Gaussian likelihood with respect to the noise parameters representing Homoscedastic uncertainties.

Different from the existing methods, we propose an innovative confidence estimation approach to achieve uncertainty-aware learning\footnote{Confidence-, uncertainty- and difficulty-aware learning are interchangeable in this paper.}.
We directly model uncertainty as difference between prediction and ground truth. The supervision signal for the confidence estimation network is
dynamically derived from the prediction of camouflaged object detection network during training. With this setting, our confidence estimation network is able to identify wrongly classified areas as uncertain and assign high confidence values to correctly predicted areas.
As shown in Fig.~\ref{fig: introduction figure}, our estimated confidence map is able to assign high uncertainty to under-segmentation, over-segmentation areas, phantom segmentation areas where false foreground predictions are distant from the target object, and object boundaries. With the confidence map, we present a confidence-aware learning framework to pay more attention to the hard/uncertain pixels for effective model learning.

We summarise our main contributions as: 
1) we propose a confidence-aware camouflaged object detection framework with an interdependent camouflaged object detection network and confidence estimation network providing mutual guidance for difficulty-aware learning;
2) we propose a dynamic confidence supervision that uses the difference between the prediction of the camouflaged object detection network and ground-truth label to train a confidence estimation network, which then outputs pixel-wise confidence revealing both true-negative and false-positive predictions to prevent the network becoming overconfident;
3) Our confidence estimation network can provide an initial evaluation of the prediction without relying on the ground truth, and experimental results show that our method compares favourably against the state-of-the-art methods.




\section{Related Work}

\subsection{Camouflaged Object Detection} 
Camouflaged object detection models \cite{le2019anabranch,yan2020mirrornet,fan2020camouflaged, lamdouar2020betrayed, ltnghia-AAAI2021, zhu2021inferring, zhai2021mutual, ren2021deep, dong2021towards, le2021camouflaged, fan2021concealed} are designed to discover the entire scope of camouflaged object(s). Different from regular objects that usually have different levels of contrast with their surroundings, camouflaged objects show similar appearance to
the environment.
\cite{cuthill2005disruptive} observes
that an effective camouflage includes two mechanisms: 1) 
background pattern matching, where the colour is similar to the environment, and 2) disruptive coloration, which usually involves bright colours along edge, obscuring
the boundary between camouflaged object and the background.
To detect camouflaged objects, \cite{le2019anabranch} introduces a multi-task learning
network with a segmentation module to produce the camouflage map and a classification module to estimate the possibility of the input image containing camouflaged objects. \cite{ltnghia-AAAI2021} presents a camouflaged instance segmentation network to produce instance level camouflage predictions. \cite{fan2020camouflaged} contributes the largest camouflaged object training set with an SINet for camouflaged object detection. \cite{yunqiu_cod21} designs a triple task learning framework to simultaneously detect, localize and rank the camouflaged objects. Different from existing techniques, we introduce confidence-aware camouflaged object detection by modelling
the difficulty in detecting different parts of camouflaged objects.

\begin{figure*}[htb]
    \centering
    \includegraphics[width=0.95\linewidth]{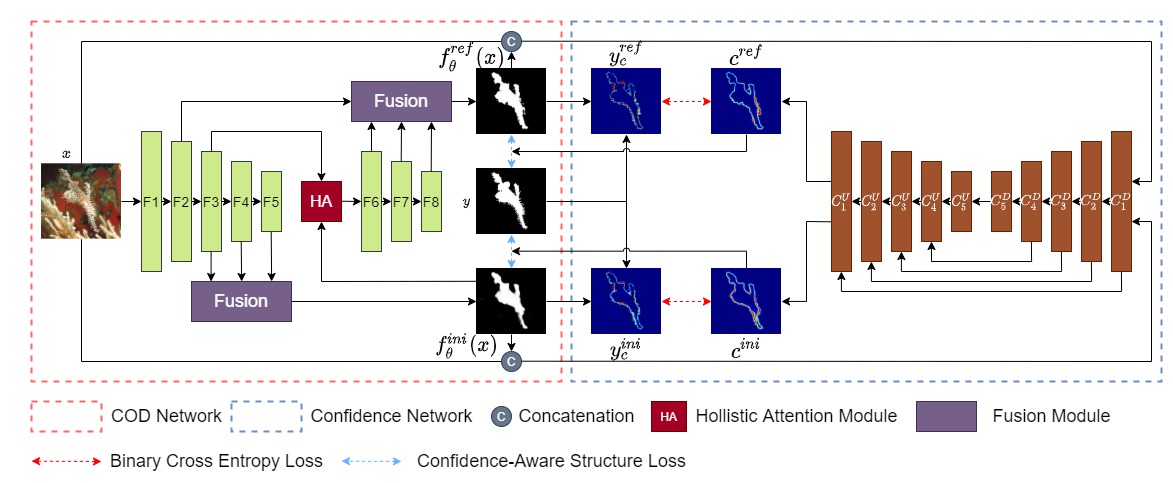}
    \vspace*{-0cm}
    \caption{The proposed confidence-aware camouflaged object detection network (CANet) is composed of two interdependent networks. The dynamic confidence supervision is derived from the predicted result of the COD network and the ground-truth camouflage map. The output of the confidence estimation network is used to guide the COD network to focus on learning image parts with low confidences through the uncertainty-guided structure loss. $F_{i}(i = 1, ..., 8)$ denotes the feature maps of the camouflaged object detection network. $C^{D}_{i}(i = 1, ..., 5)$ and $C^{U}_{i}(i = 1, ..., 5)$ denote the feature maps associated with the down convolution and up convolution operations of the confidence estimation network separately. $y$ and $f_{\theta}(x)$ denote the ground-truth and predicted camouflage maps. $y_{C}$ and $c$ denote the dynamic confidence supervision and the predicted confidence maps.}
    \label{fig: Network Overview}
\end{figure*}

\subsection{Confidence-aware Learning}
Confidence estimation has become an active research field in deep neural network based tasks, which is usually related to uncertainty estimation \cite{kendall2017uncertainties} that models the uncertainty of model predictions. Two main uncertainties have been widely studied, namely aleatoric uncertainty and epistemic uncertainty \cite{kendall2017uncertainties}. Aleatoric uncertainty captures the natural randomness in data arising from noise in the data collection, \eg sensor noise. Epistemic uncertainty captures the lack of representativeness of the model which can be reduced with increasing training data \cite{kendall2017uncertainties}.

\noindent\textbf{Aleatoric uncertainty modeling:}
\cite{kong2020sde} presents
a network which yields a probabilistic distribution as output in order to capture such uncertainty. \cite{shen2021real} employs a teacher-student paradigm to distill the aleatoric uncertainty. The teacher network generates multiple predicative samples by incorporating aleatoric uncertainty for the student network to learn.
\cite{lakshminarayanan2016simple} uses an adversarial perturbation technique to generate additional training data for the model to capture the aleatoric uncertainty.

\noindent\textbf{Epistemic uncertainty modeling:}
Bayesian neural network approaches predict epistemic uncertainty by learning the distribution of network parameters. 
Markov Chain Monte Carlo (MCMC) \cite{neal2012bayesian} has been proposed as an approximation solution. A few of its variants, such as stochastic MCMC \cite{welling2011bayesian, ma2015complete, gong2018meta, chen2014stochastic} are designed to improve its scalability to larger datasets.
An alternative approximation solution is through variational inference \cite{wu2018deterministic, louizos2017multiplicative, blundell2015weight}.
Another line of work adopts a sampling based approach \cite{ayhan2018test, kendall2017uncertainties}. The drop-out method \cite{gal2016dropout} derives the confidence from the multiple forward passes of samples.
Ensemble based solutions pass the input data through multiple replicated models \cite{lakshminarayanan2016simple} or a model with a multi-head decoder \cite{zhang2020uncertainty} to obtain multiple results in order to compute the inference mean and variance. 


\noindent\textbf{Confidence-aware learning:}
Confidence estimation has shown to be effective in improving model
performance.
\cite{neumann2018relaxed} utilises the estimated confidence to relax the softmax loss function to achieve better performance in pedestrian detection. \cite{nie2019difficulty} uses the learnt confidence to pick out hard pixels and directs the segmentation model to focus on them.
\cite{wannenwetsch2020probabilistic} employs the estimated confidence as an additional filter on the pixel-adaptive convolution to improve the performance of the upsampling operation.
Focal loss \cite{lin2017focal} emphasises learning hard samples in the classification task to deal with the imbalanced learning problem.

\noindent\textbf{Uniqueness of our solution:} We propose a fully convolutional neural network based confidence estimation model to directly output a pixel-wise confidence map.
Different from the existing techniques \cite{lin2017focal} that assign difficulty-aware weights to the entire dataset, we dynamically generate pixel-level
confidence map as weights throughout the training. These weights are tuned specifically for each sample to direct the camouflaged object detection network to put more emphasis on learning areas where predictions are regarded as uncertain. 


\section{Our Method}
\label{Methodology}
\subsection{Overview}
As a binary segmentation network, camouflaged object detection models usually follow the conventional practice of regressing the camouflage map given the input image \cite{le2019anabranch,fan2020camouflaged}.
We introduce confidence-aware camouflaged object detection to explicitly model the confidence of network towards the current prediction. Two main modules are included in our network, namely a camouflaged object detection network to produce the camouflage map, and a confidence estimation network to explicitly estimate the confidence in the current prediction. We show the pipeline of our framework in Fig.~\ref{fig: Network Overview}.

Our training dataset is $D=\{x_i,y_i\}_{i=1}^N$, where $x_i$ and $y_i$ are the image and its corresponding ground-truth camouflage map, $i$ indexes the training images, and $N$ is size of the training dataset. We define the camouflaged object detection module as $f_\theta$ (\enquote{COD Network} in Fig.~\ref{fig: Network Overview}), which generates our predicted camouflage map. Then the confidence estimation network $g_\beta$ (\enquote{Confidence Network} in Fig.~\ref{fig: Network Overview}) takes the concatenation of the predicted camouflage map and image as input to estimate the pixel-wise confidence map indicating the awareness of the model towards its prediction. 



\subsection{Camouflaged Object Detection Network}
\label{COD Network}
\begin{figure}[htb]
\small
    \centering
    \includegraphics[width=0.7\columnwidth]{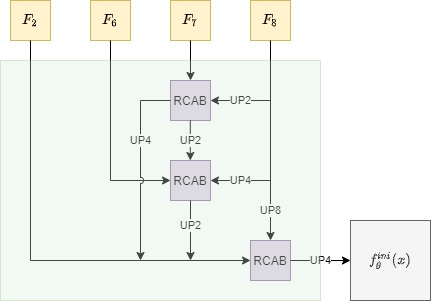}
    \vspace*{+0.2cm}
    \caption{The structure of the fusion module used to produce the final predicted camouflage map. RCAB is the residual channel attention block from \cite{zhang2018image}. UP denotes the upsampling through bilinear interpolation and its suffix indicates the scale factor.}
    \label{fig: Fusion Module}
\end{figure}

The proposed COD network employs a ResNet-50 \cite{he2016deep} encoder to produce the feature maps $F_{i}(i = 1, ..., 5)$. A Fusion Module (FM) is proposed to combine the feature maps of different levels. As illustrated in Fig.~\ref{fig: Fusion Module}, the FM progressively fuses the high-level features with the lower-level features. In each fusion operation, the highest-level feature is included to provide semantic guidance. Similar to \cite{cpd_sal}, the initial prediction $\hat{y}^{ini}=f^{ini}_\theta(x)$ utilising feature maps $F_{3-5}$ also serves as an attention mechanism on feature map $F_{3}$, leading to the computation of feature map $F_{6}$. $F_{7, 8}$ are obtained by passing $F_{6}$ through residual blocks. The final prediction $\hat{y}^{ref}=f^{ref}_\theta(x)$ is computed by fusing the feature maps $F_{2, 6-8}$. The relatively low-level feature map $F_{2}$ provides more spatial information which is important for segmentation tasks to recover a more crisp structure.

Given an input image $x$, our camouflaged object detection network produces two different predictions: $\hat{y}^{ini}$ and $\hat{y}^{ref}$ in the range of $(0, 1)$, where supervision is provided for both predictions.
This setting allows the initial prediction to recover a more complete 
camouflaged object, which subsequently serves as a better attention map to filter the feature map $F_{3}$. The final prediction $\hat{y}^{ref}$ is adopted as the camouflaged object detection result for evaluation.

\subsection{Confidence Estimation Network}
\label{Confidence Network}
The confidence estimation network employs a UNet \cite{ronneberger2015u} structure to obtain pixel-accurate confidence prediction. It consists of 5 down-convolution features denoted as $C^{D}_{i}(i = 1, ..., 5)$ and 5 up-convolution features denoted as $C^{U}_{i}(i = 1, ..., 5)$ with pairwise corresponding resolutions. The proposed down-convolution block has two $3 \times 3$ convolutional layers (\enquote{Conv3}), each followed by a batch normalisation \cite{ioffe2015batch} and a Leaky ReLU \cite{he2015delving} activation function with negative slope set to 0.2 and ends with a dropout layer ("$D(\cdot)$") of rate 0.5.
The down-convolution operation can be summarised as in Eq.~\ref{eqn: Down Convolution Block}: 
\begin{equation}
\label{eqn: Down Convolution Block}
    C^{D}_{n} = D(Conv3(Conv3(C^{D}_{n-1})))
\end{equation}
The up-convolution block consists of a $2 \times 2$ transposed convolutional layer (\enquote{TConv2}) and two $3 \times 3$ convolutional layers, each followed by a batch normalisation and a Leaky ReLU activation function with 0.2 negative slope. Down-convolution and up-convolution features are concatenated before the two convolutional layers. Dropout layers of rate 0.5 are used after the transposed convolutional layer and at the end of the up-convolution operation. The up-convolution operation can be summarised as in Eq.~\ref{eqn: up convolution block}:
\begin{equation}
    C^{U}_{n} = D(Conv3(Conv3(\amalg(C^{D}_{n}, D(TConv2(C^{U}_{n + 1}))))))
    \label{eqn: up convolution block}
\end{equation}
where $\amalg(\cdot)$ denotes a concatenation operation. 

The confidence estimation network takes the concatenation of model prediction ($\hat{y}^{ini}$ and $\hat{y}^{ref}$) and image $x$ as input to produce an one-channel confidence map, which is defined as $c^{ini} = g_\beta(\amalg(x,\hat{y}^{ini}))$ for the initial prediction, and $c^{ref} = g_\beta(\amalg(x,\hat{y}^{ref}))$ for the final prediction. The estimated confidence maps are supervised with dynamic confidence supervisions derived from the predictions of the camouflaged object detection network $f_{\theta}(x)$ and the ground-truth camouflage map $y$.

\subsection{Dynamic Confidence Supervision}
\label{Online Confidence Supervision}
We design the confidence estimation network to explicitly produce an uncertainty or difficulty map to guide the learning of the camouflaged object detection network during training, and for reasonable awareness of accuracy during testing. To achieve this, we introduce a dynamic supervision $y_c$ for the confidence estimation network, which is defined as in Eq.~\ref{eqn: Online Confidence Supervision}:
\begin{equation}
    \label{eqn: Online Confidence Supervision}
    y_c = y\times(1-\hat{y})+(1-y)\times \hat{y}.
\end{equation}
The dynamic supervision $y_c$ is defined as a pixel-wise L1 distance between the prediction results $f_{\theta}(x)$ and its corresponding ground-truth labels $y$. It has high uncertainty to target pixels where the camouflaged object detection network makes confident but false predictions. For example, if the camouflage prediction for pixel $u,v$ is $\hat{y}^{u, v} =  0.01$, indicating a background pixel,
whereas its ground-truth label is $y^{u, v} = 1$, suggesting it is a foreground pixel, our dynamic supervision is $y_{c}^{u,v} = 0.99$ representing a difficult or uncertain pixel.


The confidence estimation network is trained with a binary cross-entropy loss which is defined as in Eq.~\ref{eqn: Confidence Estimation Network Objective Function}: 
\begin{equation}
\label{eqn: Confidence Estimation Network Objective Function}
\begin{aligned}
    \mathcal{L}_{c} = 0.5\times(\mathcal{L}_{ce}(c^{ini},y_c^{ini})+\mathcal{L}_{ce}(c^{ref},y_c^{ref})),
\end{aligned}
\end{equation}
where $\mathcal{L}_{ce}$ is the binary cross-entropy loss, $y_c^{ini}$ and $y_c^{ref}$ are dynamic supervisions for the initial prediction and our final prediction respectively.

\subsection{Confidence-Aware Learning}
\label{Uncertainty-Aware Learning}
Camouflaged object detection has different learning difficulties across the image. The pixels along the object boundary are harder to differentiate than the background pixels that are further away from the camouflaged objects. Further, the camouflage foreground contains parts with different level of camouflage, where some parts are easy to recognise, \eg eyes, mouths and \etc and some others are hard to distinguish, \eg the body region has similar appearance to the background. We intend to model such varying learning difficulty across the image by importing the estimated confidence map to our camouflaged object detection network.
Specifically, inspired by \cite{wei2020f3net}, we propose to train the camouflaged object detection network with a confidence-aware structure loss, which is defined in Eq.~\ref{eqn: Uncertainty-Aware Structure Loss}:
\begin{equation}
\label{eqn: Uncertainty-Aware Structure Loss}
\begin{aligned}
    \mathcal{L}_{s} = &\sum_{u, v} w^{u,v} \mathcal{L}_{ce} + \sum_{u, v} w^{u,v} \mathcal{L}_{dice},
\end{aligned}
\end{equation}
where the weight term is defined as: $w^{ini} = 1 + \lambda c^{ini}$ for the initial prediction $f_\theta^{ini}(x)$ and $w^{ref} = 1 + \lambda c^{ref}$ for our final prediction $f_\theta^{ref}(x)$, and $\lambda$ is a parameter controlling the scale of attention given to uncertain pixels.
The first term is a weighted binary cross-entropy loss and the second term is a weighted Dice Loss.
The weight term $w$ provides sample specific pixel-wise weights, letting the camouflaged object detection model focus on learning less confident or uncertain pixels, especially where confident wrong predictions are made.
Our whole algorithm is shown in Algorithm \ref{alg:whole_algorithm}. The comparisons between predictions with and without confidence as guidance in Fig.~\ref{fig: introduction figure} show the effectiveness of our confidence-aware learning.

\begin{algorithm}[H]
\small
\caption{Confidence-aware Camouflaged Object Detection}
\textbf{Input}: \\
(1) Training dataset $D=\{x_i,y_i\}_{i=1}^N$, where $N$ is size of the training dataset;\\
(2) maximal number of learning epochs $E$. 

\textbf{Output}: 
Parameters $\theta$ for the camouflaged object detection module (CODM) and parameters $\beta$ for the confidence estimation module (CEM).
\begin{algorithmic}[1]
\State Initialise $\theta$ and $\beta$ 
\For{$t \leftarrow  1$ to $E$}
\State Generate camouflage predictions $\hat{y}^{ini}=f_\theta^{ini}(x)$ and $\hat{y}^{ref}=f_\theta^{ref}(x)$ from the CODM.
\State Produce dynamic supervisions $y_c^{ini}$ and $y_c^{ref}$ for the CEM with Eq.~\ref{eqn: Online Confidence Supervision}.
\State Obtain the confidence maps $c^{ini}=g_{\beta}(\amalg(\hat{y}^{ini}, x))$ and $c^{ref}=g_{\beta}(\amalg(\hat{y}^{ref}, x))$ from the CEM. 
\State Update CEM with loss function in Eq.~\ref{eqn: Confidence Estimation Network Objective Function}.
\State Generate confidence-aware weight $\omega^{ini}=1+\lambda c^{ini}$ for $\hat{y}^{ini}$ and confidence-aware weight $\omega^{ref}=1+\lambda c^{ref}$ for $\hat{y}^{ref}$.
\State Update CODM with loss function in Eq.~\ref{eqn: Uncertainty-Aware Structure Loss}.
\EndFor
\end{algorithmic} 
\label{alg:whole_algorithm}
\end{algorithm}


\section{Experimental Results}
\subsection{Setting:}
\noindent\textbf{Dataset:} We train our model using the COD10K training set \cite{fan2020camouflaged}, and test on four camouflaged object detection testing sets, including the CAMO \cite{le2019anabranch}, CHAMELEON \cite{Chameleon2018}, COD10K testing dataset \cite{fan2020camouflaged} and the newly introduced NC4K dataset \cite{yunqiu_cod21}.

\noindent\textbf{Evaluation Metrics:} We use four evaluation metrics to evaluate the performance of the camouflaged object detection models, including Mean Absolute Error ($\mathcal{M}$), Mean F-measure ($F_\beta^{\mathrm{mean}}$), Mean E-measure \cite{fan2018enhanced} ($E_\xi^{\mathrm{mean}}$) and S-measure \cite{fan2017structure} ($S_{\alpha}$). A detailed introduction to those metrics appears in the supplementary materials.

\noindent\textbf{Training details:}
We train our model in Pytorch with ResNet-50 \cite{he2016deep} as backbone, where the encoder part is initialized with 
weights trained on ImageNet, and other newly added layers are randomly initialized. We resize all the images and ground truth to $352\times352$. The maximum epoch is 20. The initial learning rates are $2.5 \times 10^{-5}$ and $1.5 \times 10^{-5}$ for the camouflaged object detection network and confidence estimation network respectively. The whole training takes 8.5 hours with batch size 10 on two NVIDIA GTX 2080Ti GPUs.

\begin{table*}[t!]
  \centering
  \scriptsize
  \renewcommand{\arraystretch}{1.3}
  \renewcommand{\tabcolsep}{1.0mm}
  \caption{Performance comparison with baseline models on benchmark testing dataset.}
  \begin{tabular}{l|cccc|cccc|cccc|cccc}
  \hline
  &\multicolumn{4}{c|}{CAMO~\cite{le2019anabranch}}&\multicolumn{4}{c|}{CHAMELEON~\cite{Chameleon2018}}&\multicolumn{4}{c|}{COD10K~\cite{fan2020camouflaged}}&\multicolumn{4}{c}{NC4K~\cite{yunqiu_cod21}} \\
    Method & $S_{\alpha}\uparrow$&$F_{\beta}^{\mathrm{mean}}\uparrow$&$E_{\xi}^{\mathrm{mean}}\uparrow$&$\mathcal{M}\downarrow$& $S_{\alpha}\uparrow$&$F_{\beta}^{\mathrm{mean}}\uparrow$&$E_{\xi}^{\mathrm{mean}}\uparrow$&$\mathcal{M}\downarrow$ &  $S_{\alpha}\uparrow$ & $F_{\beta}^{\mathrm{mean}}\uparrow$ & $E_{\xi}^{\mathrm{mean}}\uparrow$ & $\mathcal{M}\downarrow$ & $S_{\alpha}\uparrow$
    & $F_{\beta}^{\mathbf{\mathrm{mean}}}\uparrow$ & $E_{\xi}^{\mathbf{\mathrm{mean}}}\uparrow$ & $\mathcal{M}\downarrow$  \\
  \hline
  CPD \cite{cpd_sal} & 0.716 & 0.618 & 0.723 & 0.113 & 0.857 & 0.771 & 0.874 & 0.048 & 0.750 & 0.595 & 0.776 & 0.053 & 0.790 & 0.708 & 0.810 & 0.071  \\
  SCRN \cite{scrn_sal}& 0.779 & 0.705 & 0.796 & 0.090 & 0.876 & 0.787 & 0.889 & 0.042 & 0.789 & 0.651  & 0.817 & 0.047 & 0.832 & 0.759 & 0.855 & 0.059  \\
  PoolNet \cite{Liu19PoolNet}  & 0.730 & 0.643 & 0.746 & 0.105 & 0.845 & 0.749 & 0.864 & 0.054 & 0.740 & 0.576 & 0.776 & 0.056  & 0.785 & 0.699 & 0.814 & 0.073  \\
  F3Net \cite{wei2020f3net} & 0.711 & 0.616 & 0.741 & 0.109 & 0.848 & 0.770 & 0.894 & 0.047 & 0.739 & 0.593 & 0.795 & 0.051 & 0.782 & 0.706 & 0.825 & 0.069 \\
  ITSD\cite{zhou2020interactive} & 0.750 & 0.663 & 0.779 & 0.102 & 0.814 & 0.705 & 0.844 & 0.057 & 0.767 & 0.615 & 0.808 & 0.051  & 0.811 & 0.729 & 0.845 & 0.064  \\ 
  BASNet \cite{basnet_sal} & 0.615 & 0.503 & 0.671 & 0.124 & 0.847 & 0.795 & 0.883 & 0.044 & 0.661 & 0.486 & 0.729 & 0.071  & 0.698 & 0.613 & 0.761 & 0.094  \\
  EGNet \cite{zhao2019EGNet} & 0.737 & 0.655 & 0.758 & 0.102 & 0.856 & 0.766 & 0.883 & 0.049 & 0.751 & 0.595 & 0.793 & 0.053  & 0.796 & 0.718 & 0.830 & 0.067  \\
  SINet \cite{fan2020camouflaged} & 0.745 & 0.702 & 0.804 & 0.092 & 0.872 & 0.827 & 0.936 & 0.034 & 0.776 & 0.679 & 0.864 & 0.043 & 0.810 & 0.772 & 0.873 & 0.057 \\ 
  R2Net~\cite{feng2020residual}  & 0.772 & 0.685 & 0.777 & 0.098 & 0.861 & 0.766 & 0.869 & 0.047 & 0.787 & 0.636 & 0.801 & 0.048  & 0.823 & 0.739 & 0.835 & 0.064  \\ 
  RASNet~\cite{chen2020reverse}  & 0.763 & 0.716 & 0.824 & 0.090 & 0.857 & 0.804 & 0.923 & 0.040 & 0.778 & 0.673 & 0.865 & 0.044  & 0.817 & 0.772 & 0.880 & 0.057  \\ 
  PraNet~\cite{fan2020pranet} & 0.769 & 0.711 & 0.825 & 0.094 & 0.860 & 0.790 & 0.908 & 0.044 & 0.790 & 0.672 & 0.861 & 0.045 & 0.822 & 0.763 & 0.877 & 0.059 \\
  TINet~\cite{zhu2021inferring} & 0.781 & 0.678 & 0.847 & 0.087 & 0.874 & 0.783 & 0.916 & 0.038 & 0.793 & 0.635 & 0.848 & 0.043 &-&-&-&-\\
  MirrorNet~\cite{yan2020mirrornet} & 0.785 & 0.719 & 0.849 & 0.077 &-&-&-&-&-&-&-&-\\
  LSR~\cite{yunqiu_cod21} & 0.793 & 0.725 & 0.826 & 0.085 & \textbf{0.893} & \textbf{0.839} & 0.938 & 0.033 & 0.793 & 0.685 & 0.868 & 0.041 & 0.839 & 0.779 & 0.883 & 0.053  \\ 
    \hline
 Ours & \textbf{0.799} & \textbf{0.770} & \textbf{0.865} & \textbf{0.075} & 0.885 & 0.831 & \textbf{0.940} & \textbf{0.029} & \textbf{0.809} & \textbf{0.703} & \textbf{0.885} & \textbf{0.035}  & \textbf{0.842} & \textbf{0.803} & \textbf{0.904} & \textbf{0.047}   \\ 
   \hline
  \end{tabular}
  \label{tab: Benchmark model comparison}
\end{table*}

\begin{figure*}[tp]
  \begin{center}
  \begin{tabular}{{c@{ } c@{ } c@{ } c@{ } c@{ } c@{ } c@{ } c@{ } c@{ }}}
  {\includegraphics[width=0.105\linewidth]{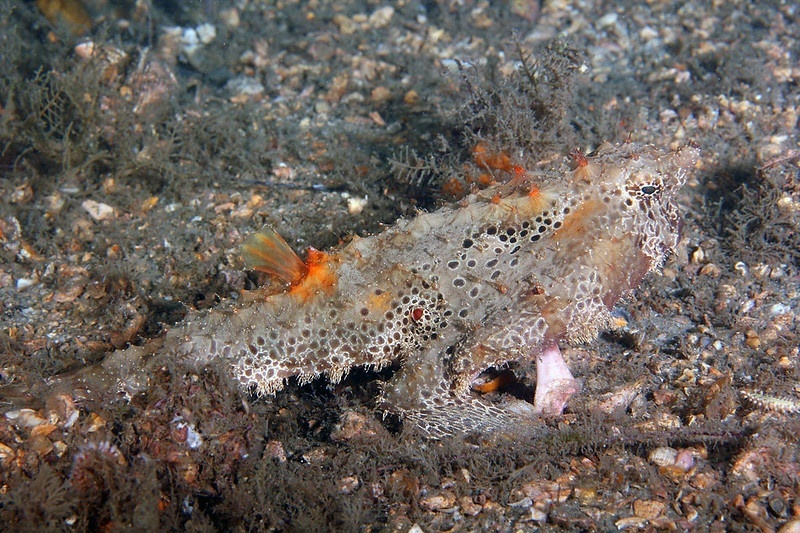}} &
  {\includegraphics[width=0.105\linewidth]{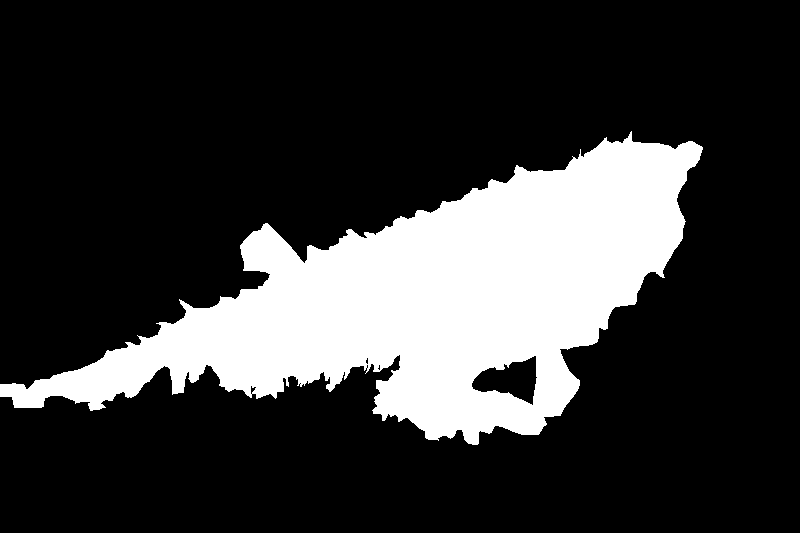}} &
  {\includegraphics[width=0.105\linewidth]{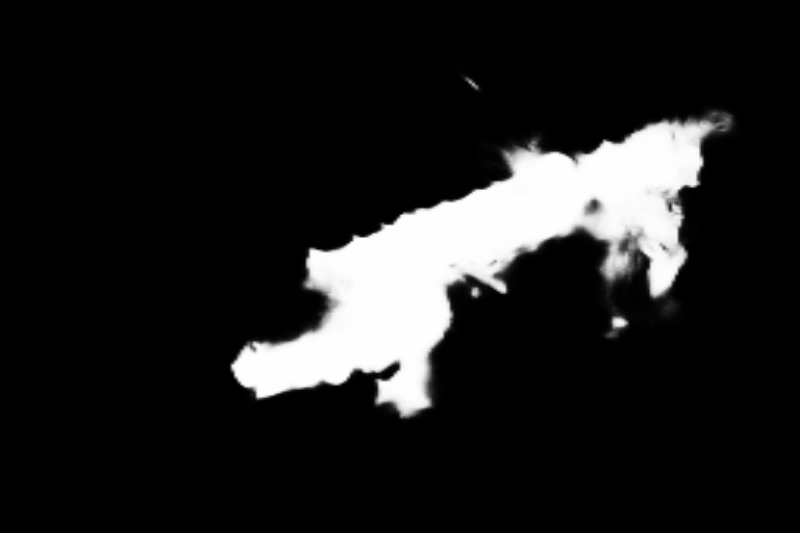}} &
  {\includegraphics[width=0.105\linewidth]{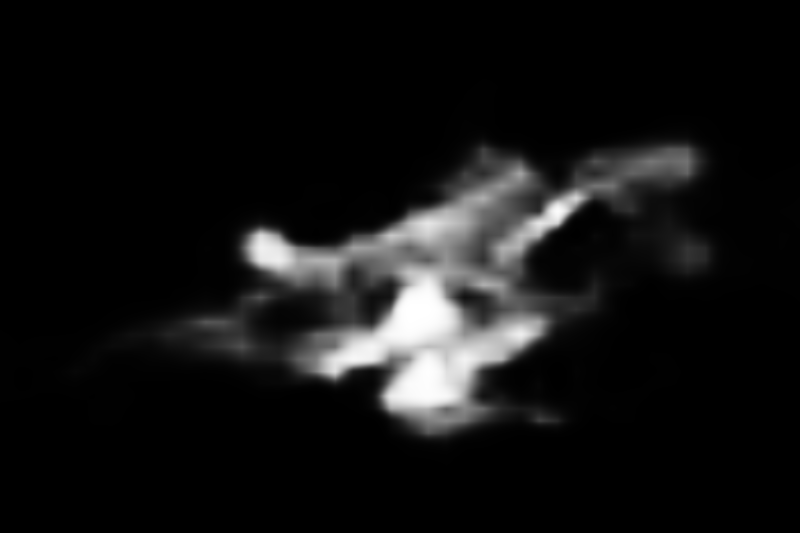}} &
  {\includegraphics[width=0.105\linewidth]{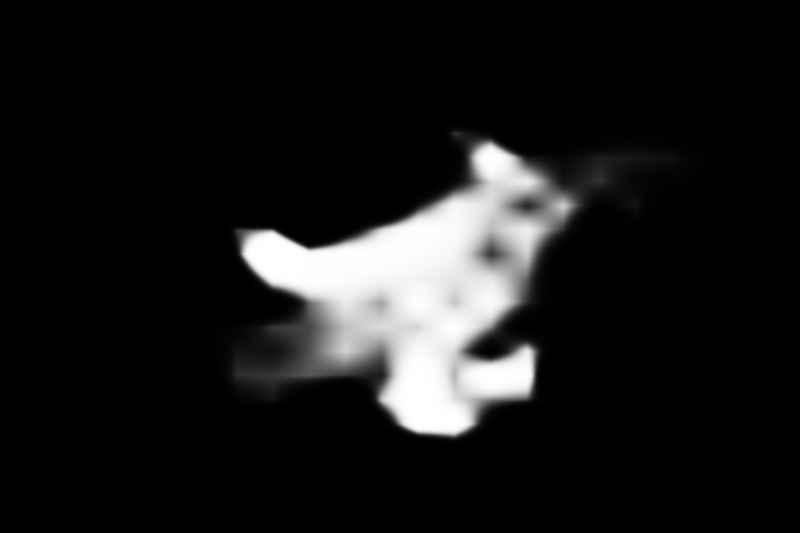}} &
  {\includegraphics[width=0.105\linewidth]{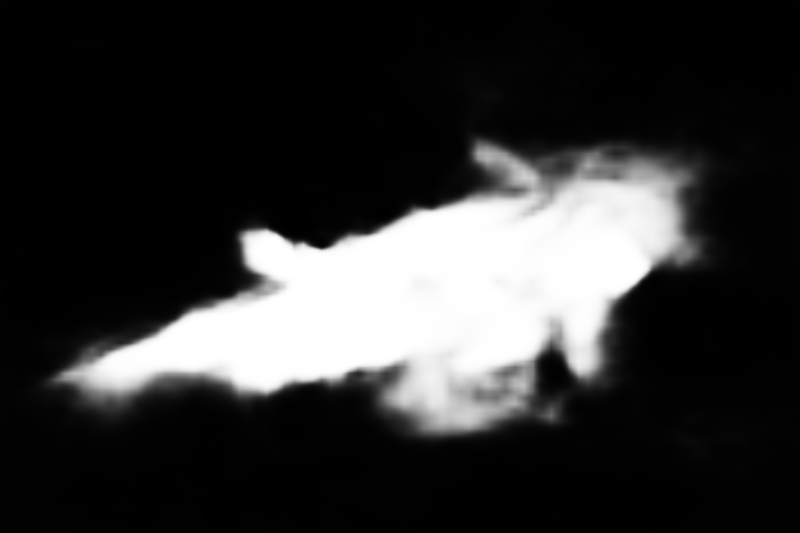}} &
  {\includegraphics[width=0.105\linewidth]{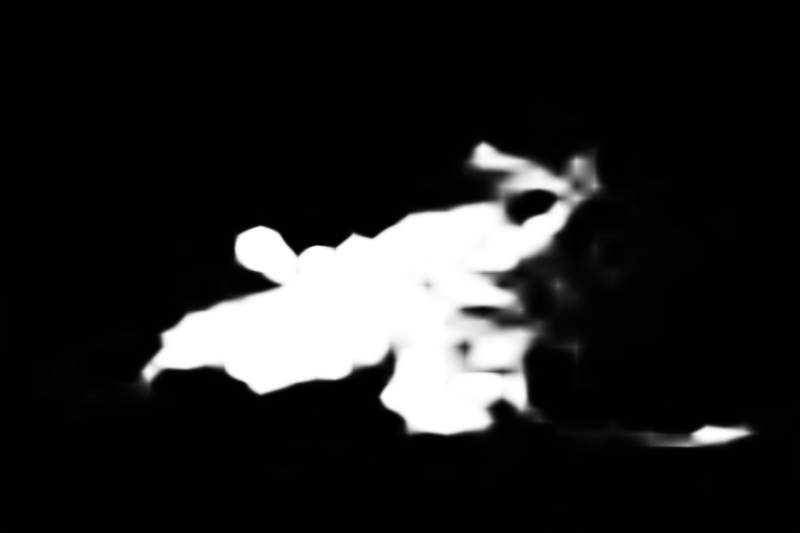}} &
  {\includegraphics[width=0.105\linewidth]{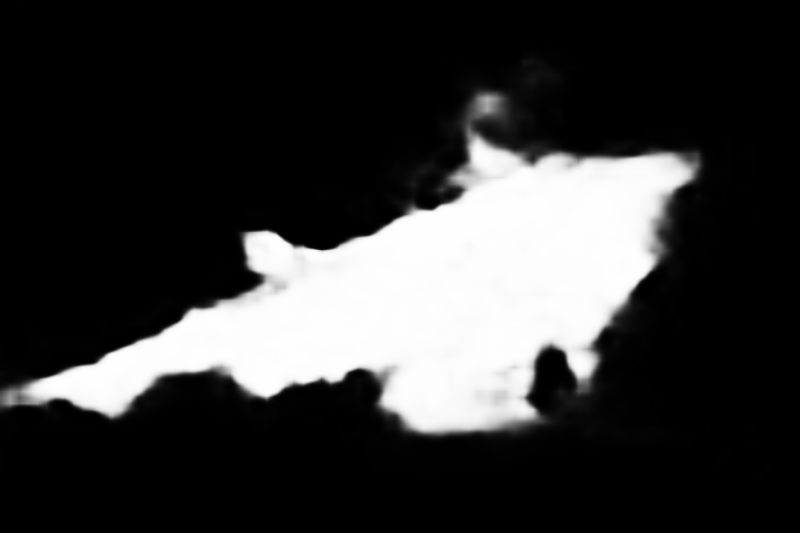}} &
  {\includegraphics[width=0.105\linewidth]{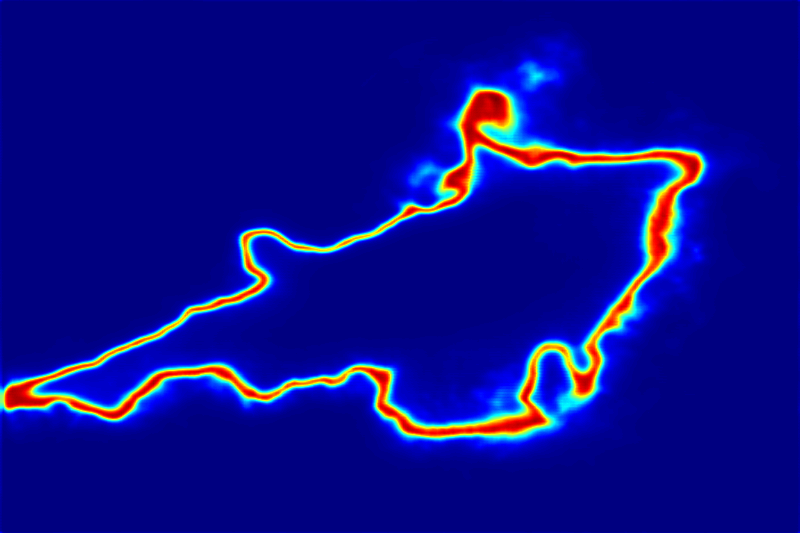}}\\
  {\includegraphics[width=0.105\linewidth]{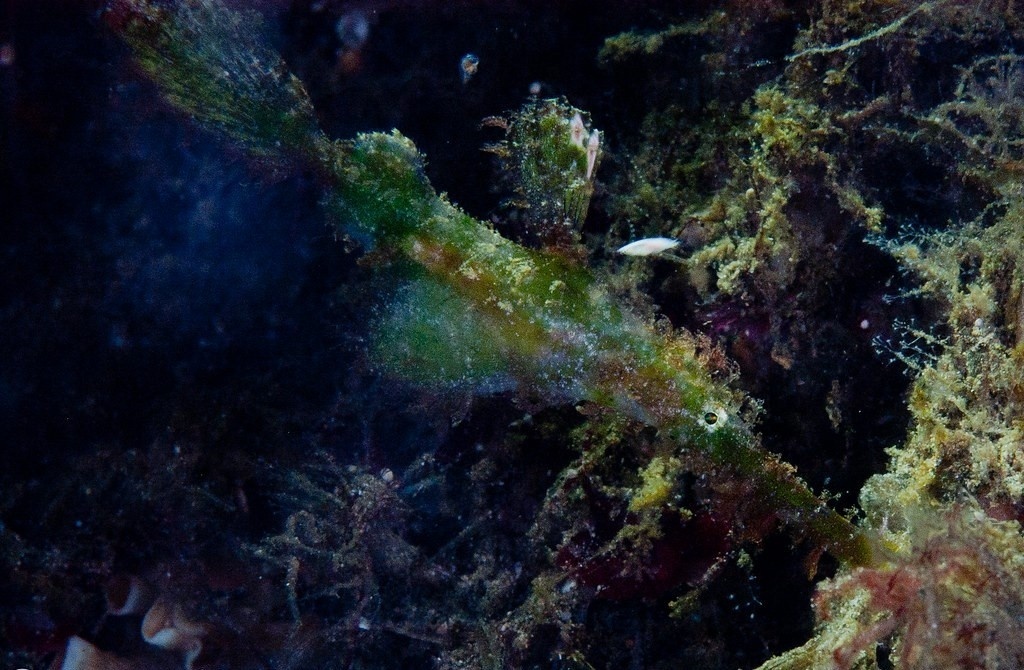}} &
  {\includegraphics[width=0.105\linewidth]{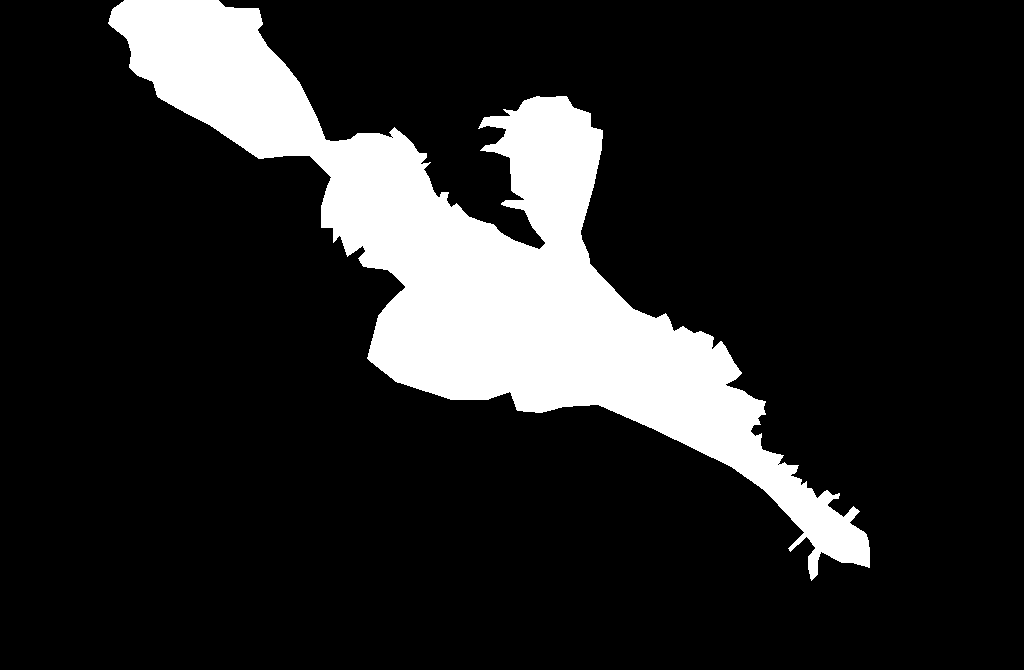}} &
  {\includegraphics[width=0.105\linewidth]{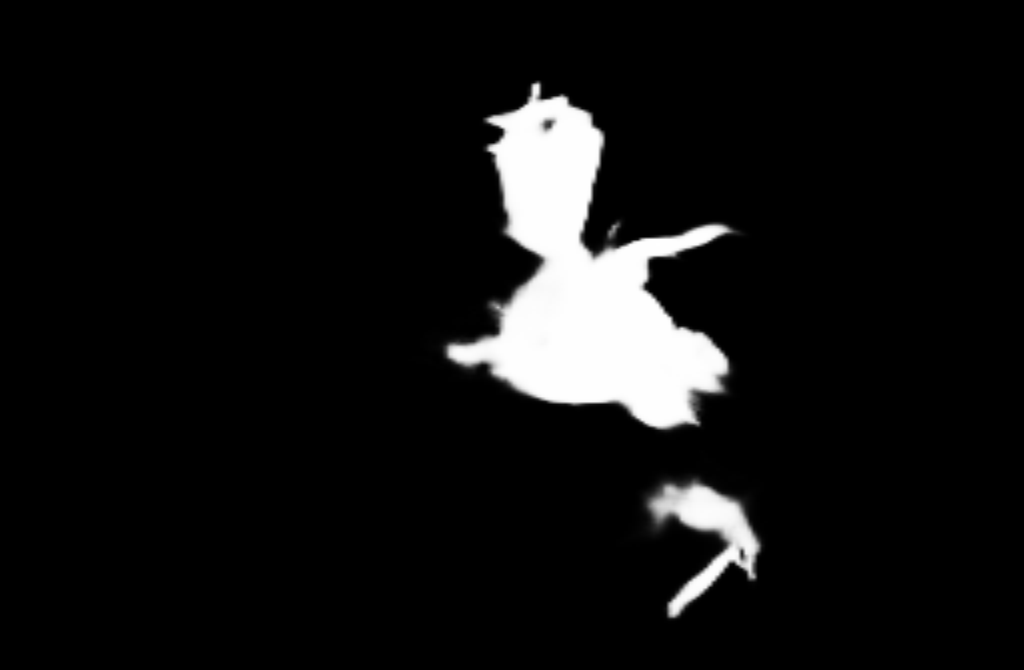}} &
  {\includegraphics[width=0.105\linewidth]{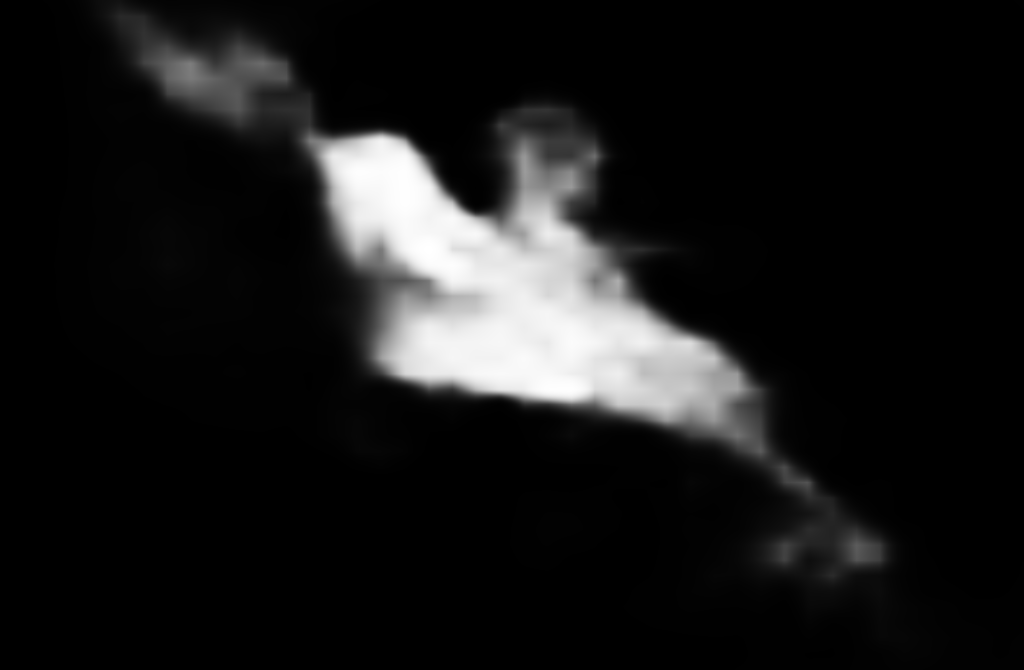}} &
  {\includegraphics[width=0.105\linewidth]{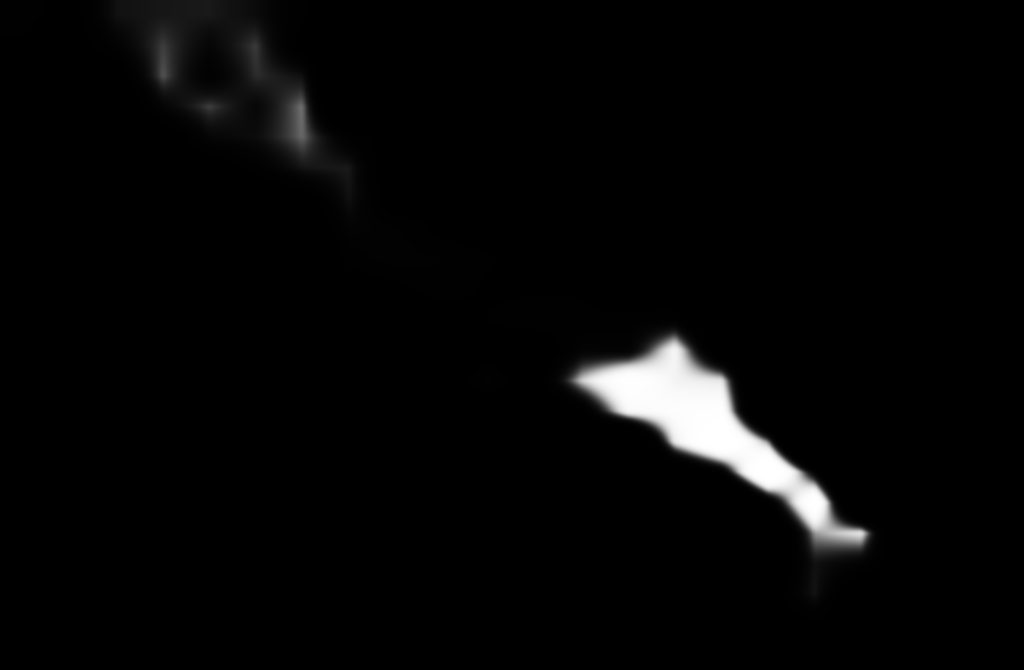}} &
  {\includegraphics[width=0.105\linewidth]{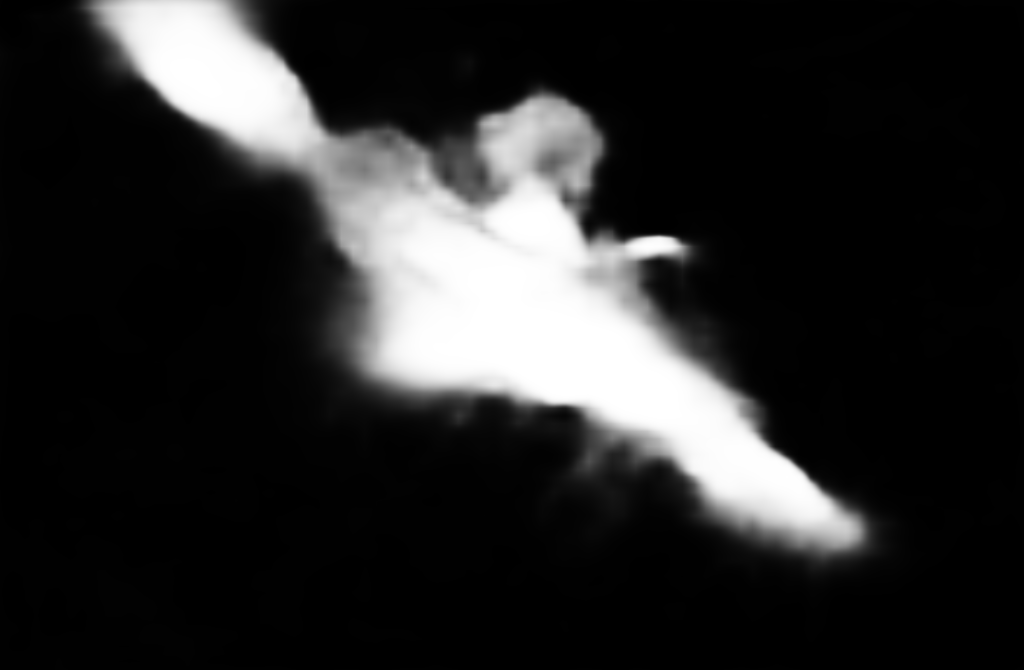}} &
  {\includegraphics[width=0.105\linewidth]{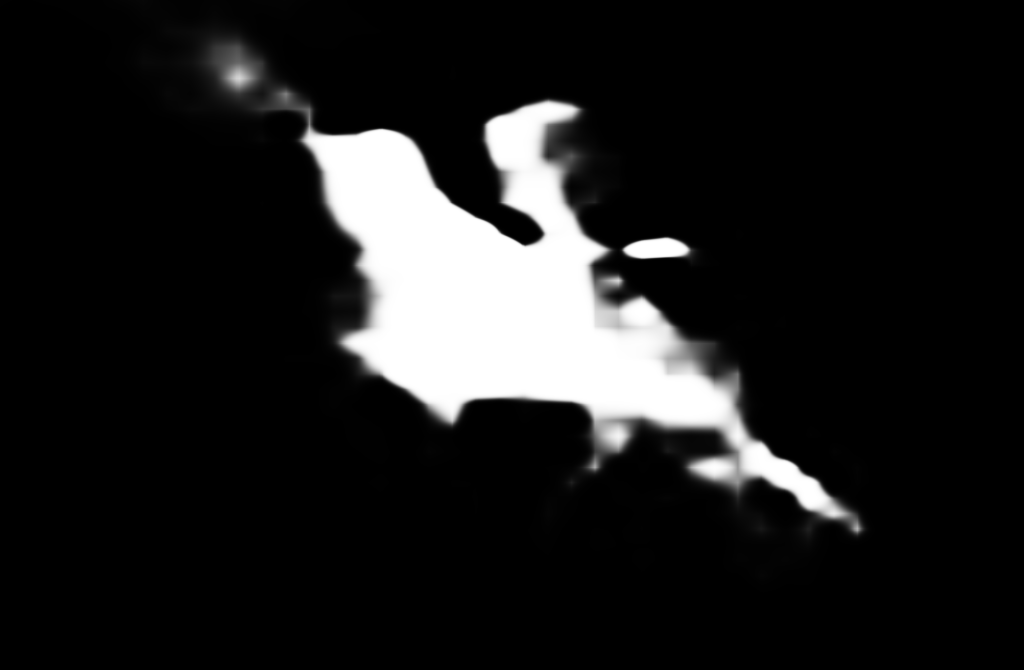}} &
  {\includegraphics[width=0.105\linewidth]{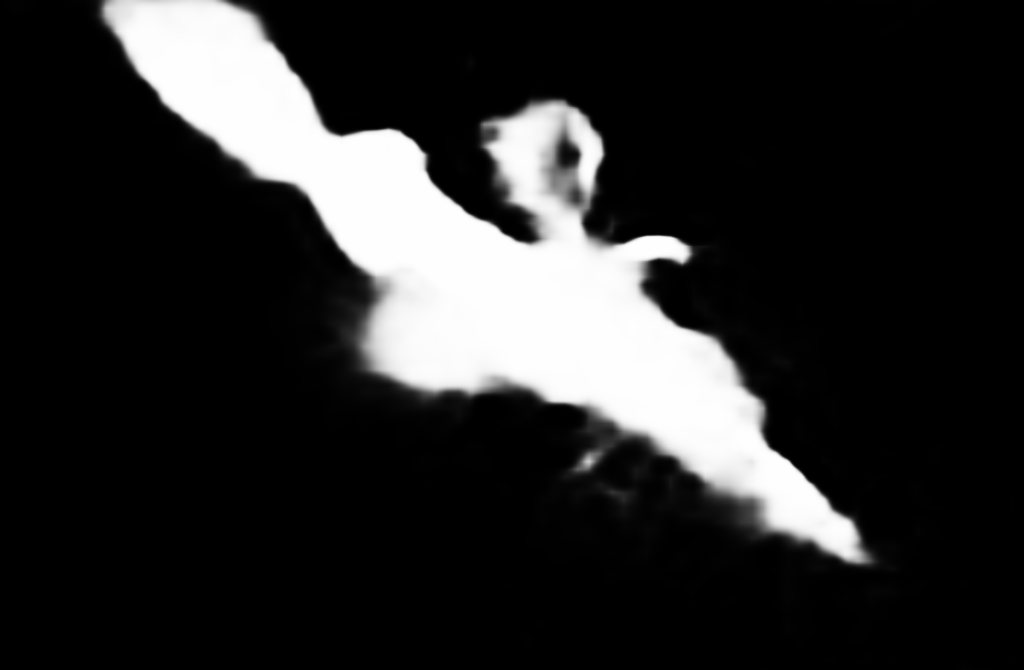}} &
  {\includegraphics[width=0.105\linewidth]{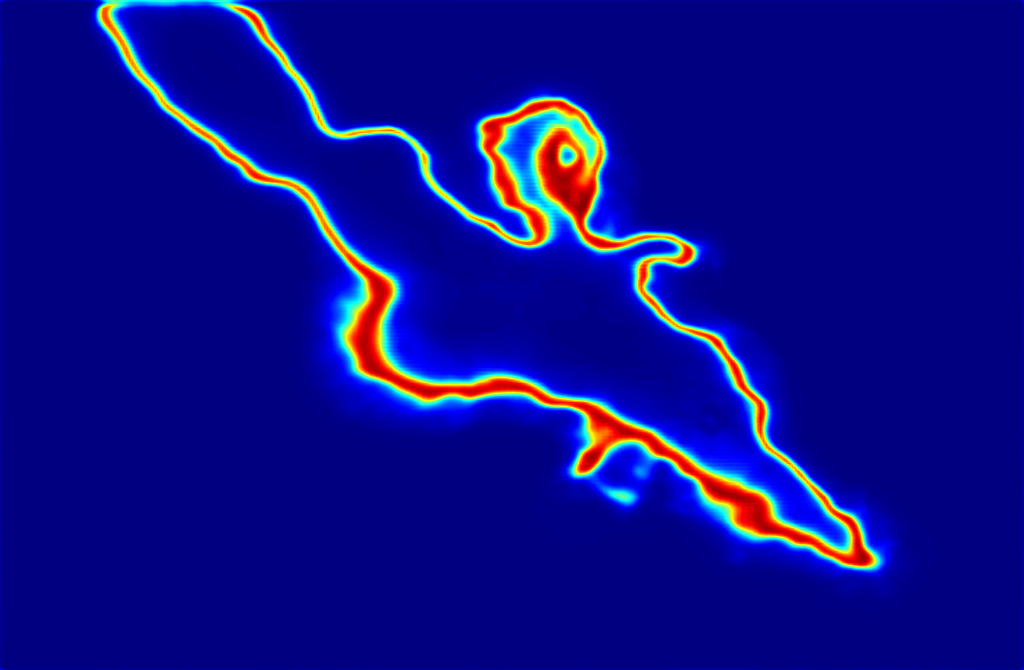}}\\
  {\includegraphics[width=0.105\linewidth]{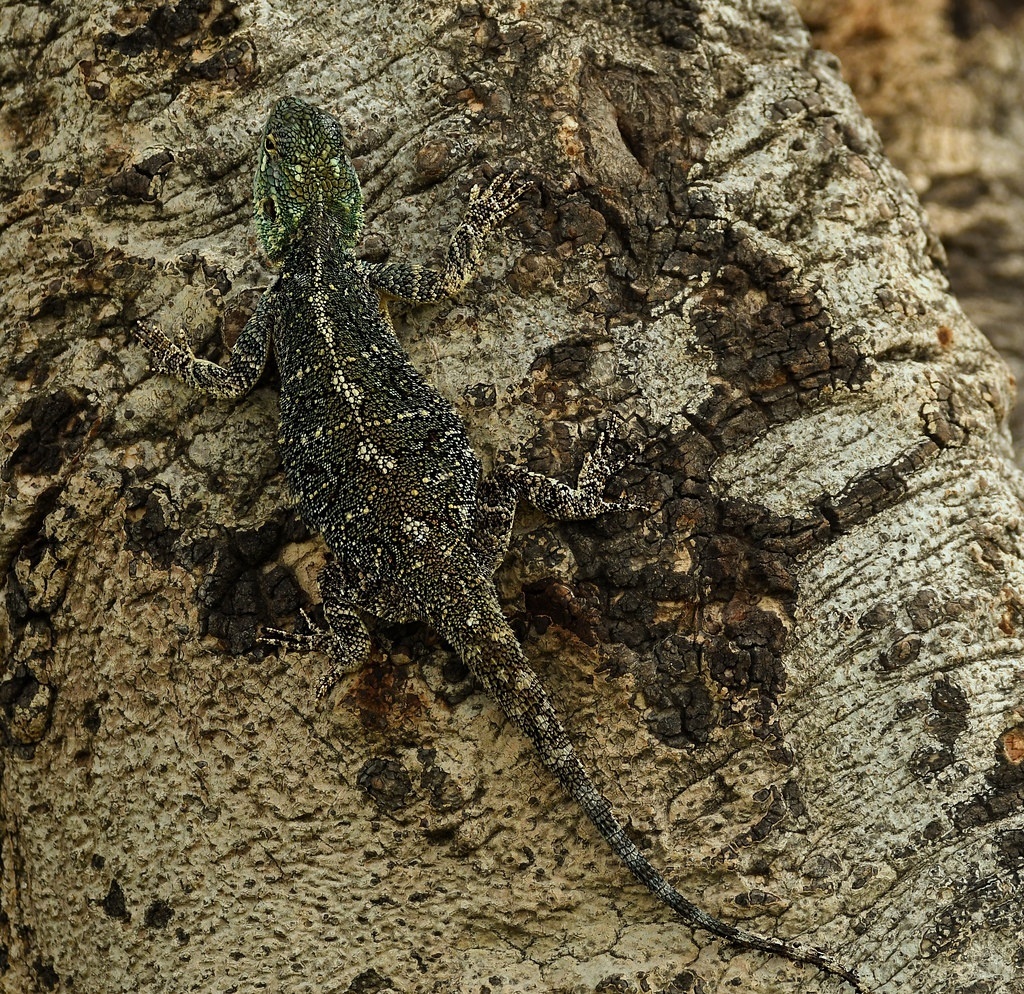}} &
  {\includegraphics[width=0.105\linewidth]{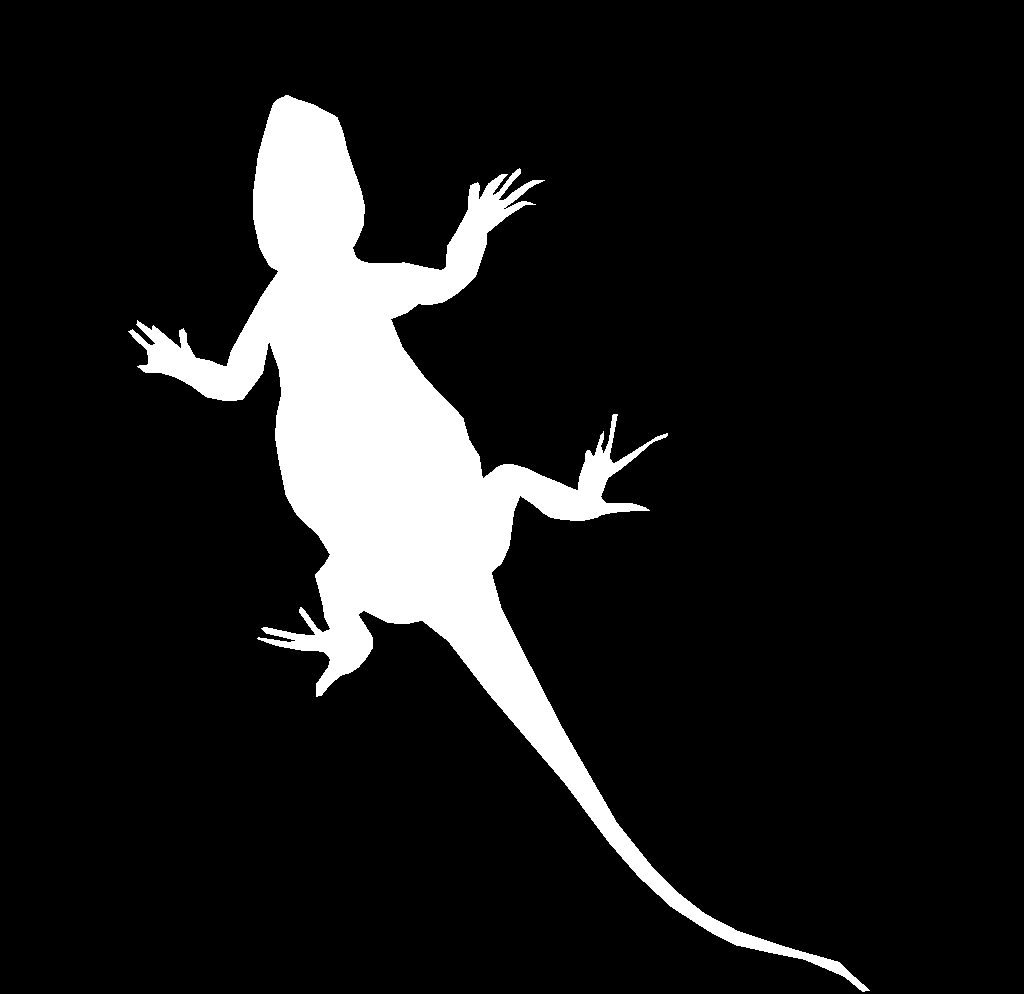}} &
  {\includegraphics[width=0.105\linewidth]{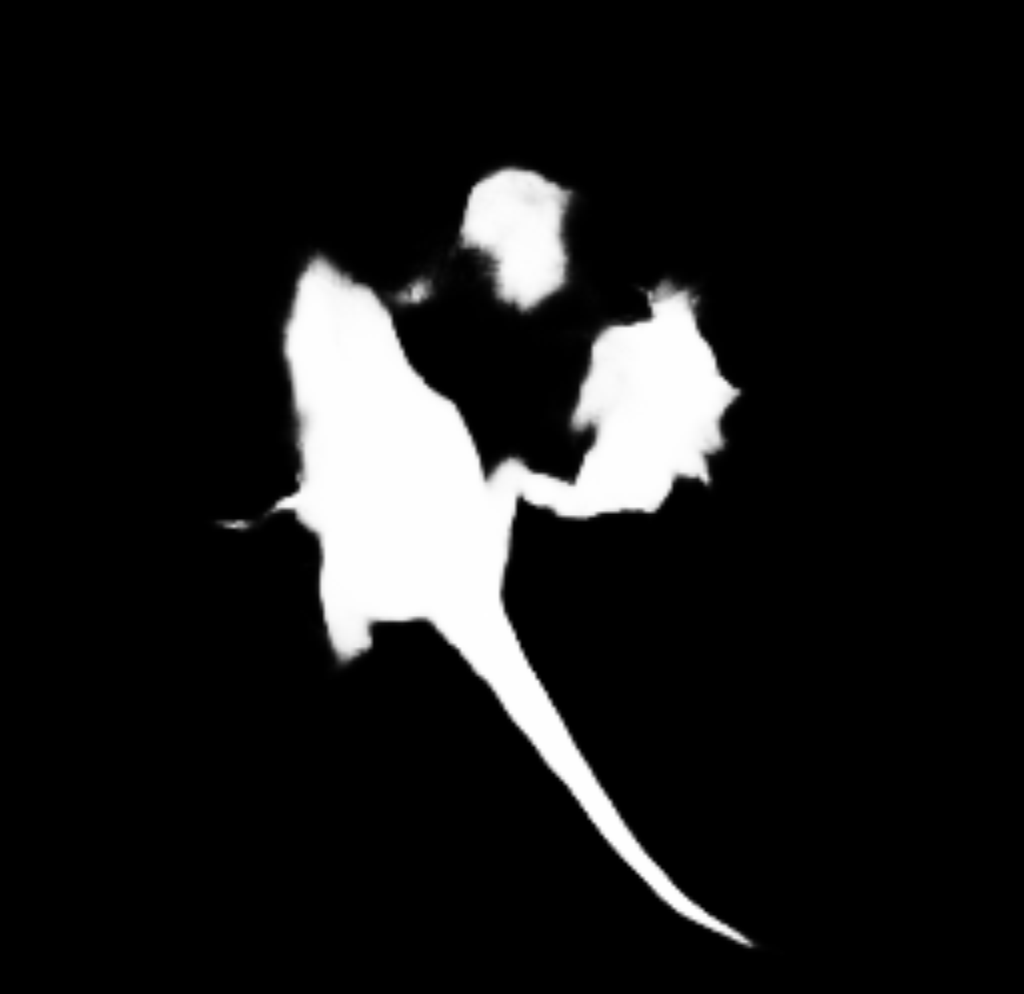}} &
  {\includegraphics[width=0.105\linewidth]{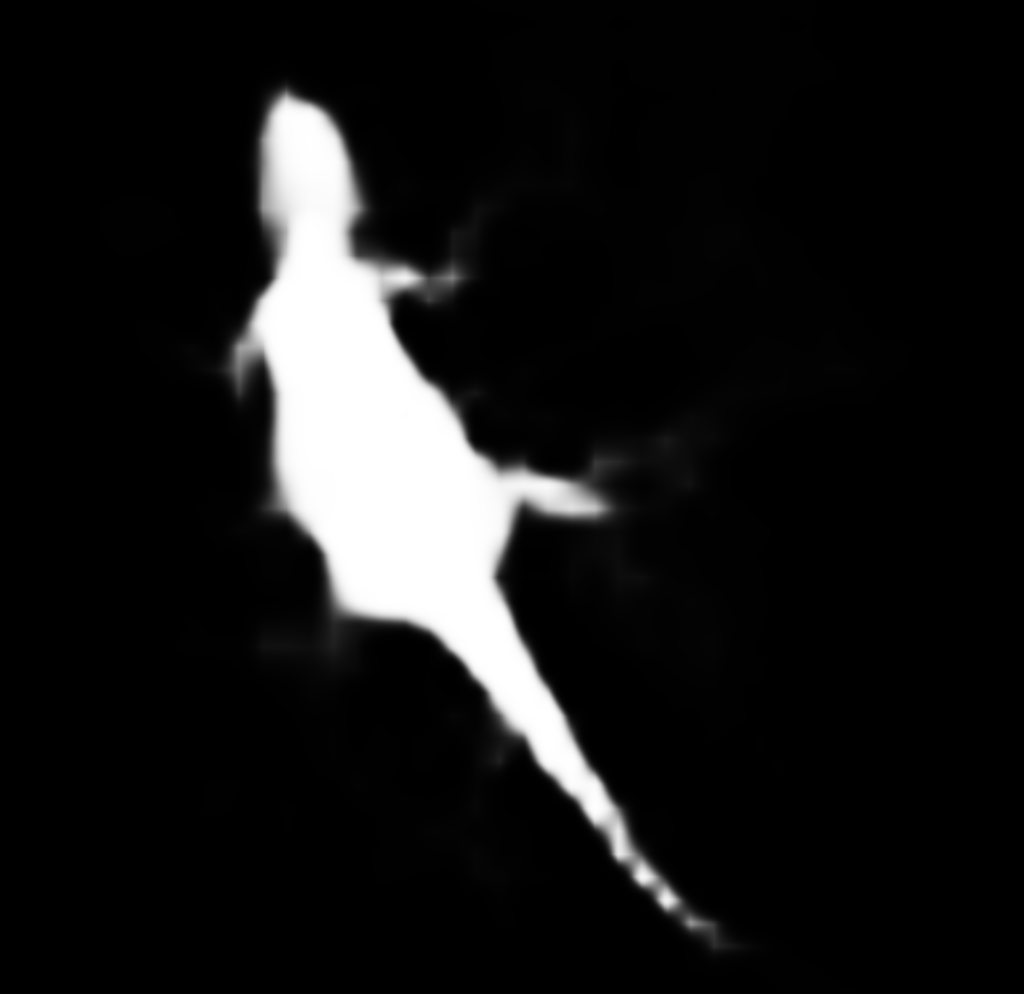}} &
  {\includegraphics[width=0.105\linewidth]{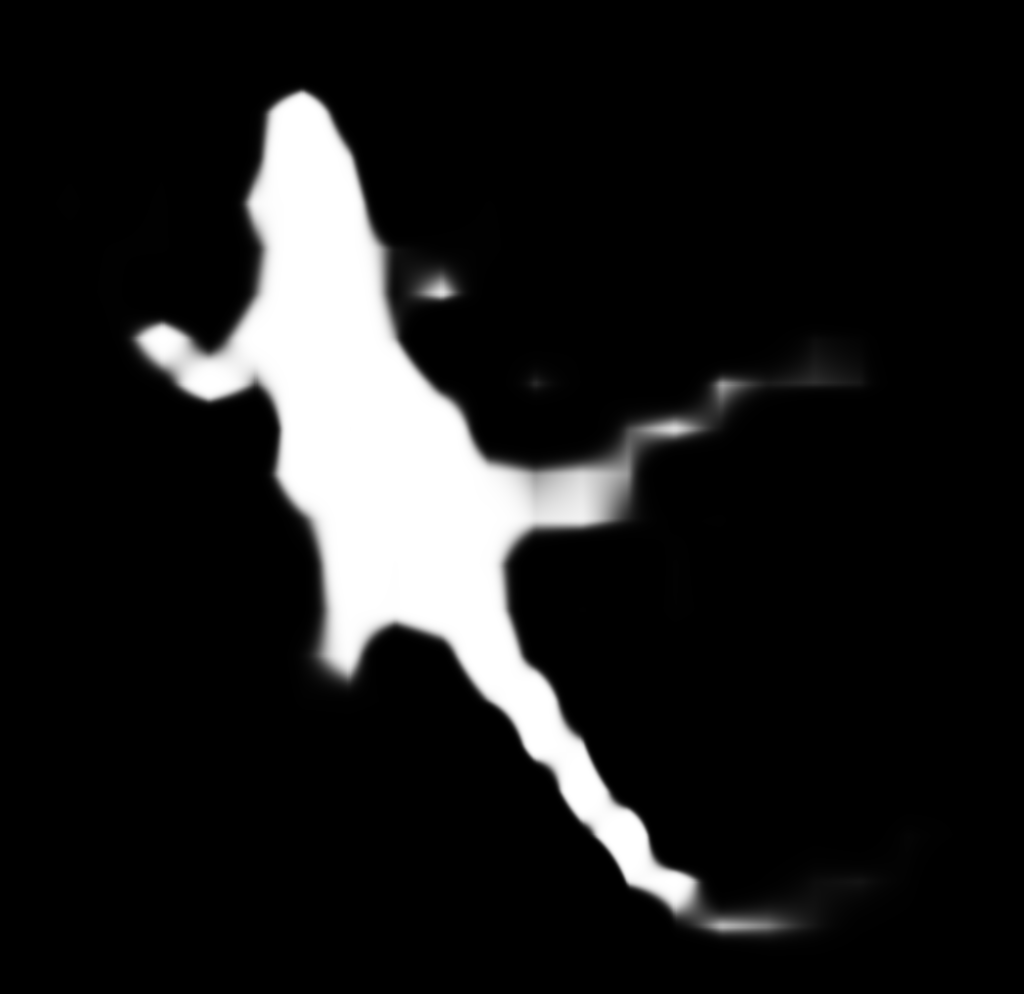}} &
  {\includegraphics[width=0.105\linewidth]{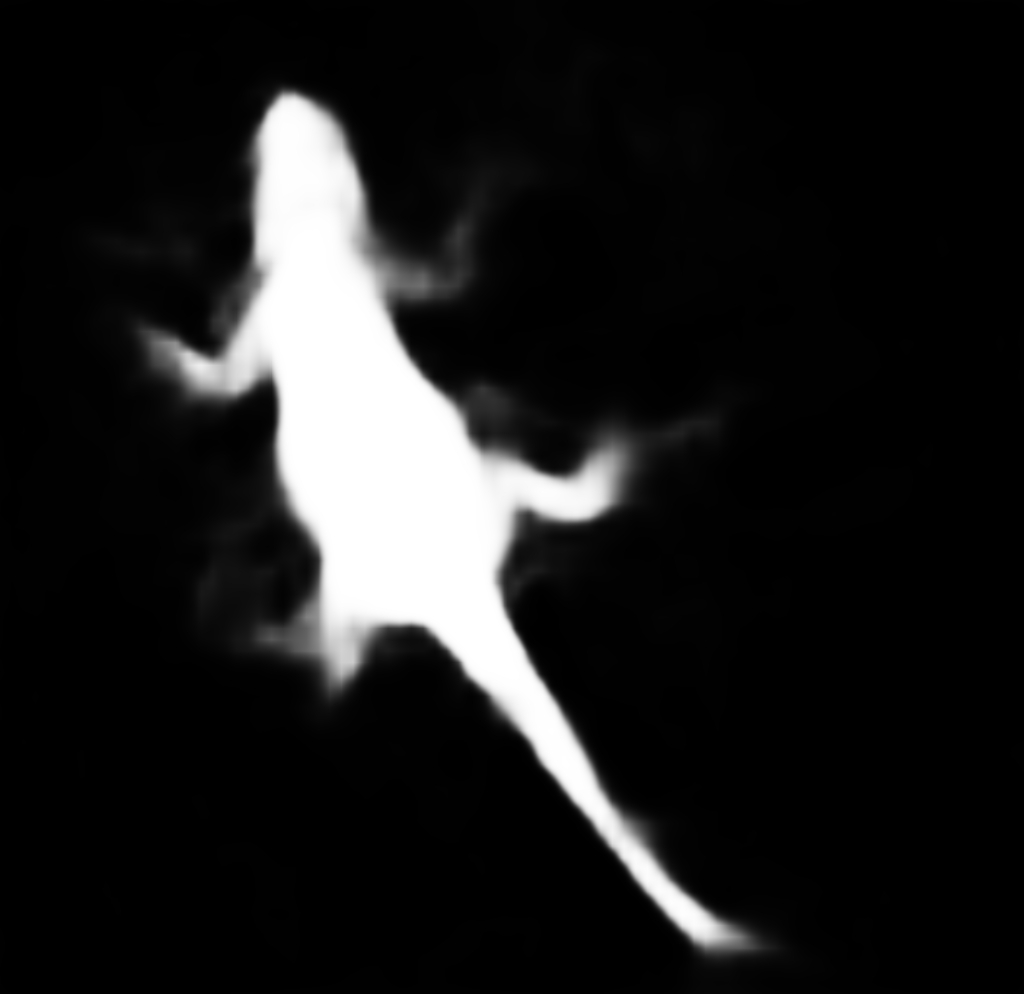}} &
  {\includegraphics[width=0.105\linewidth]{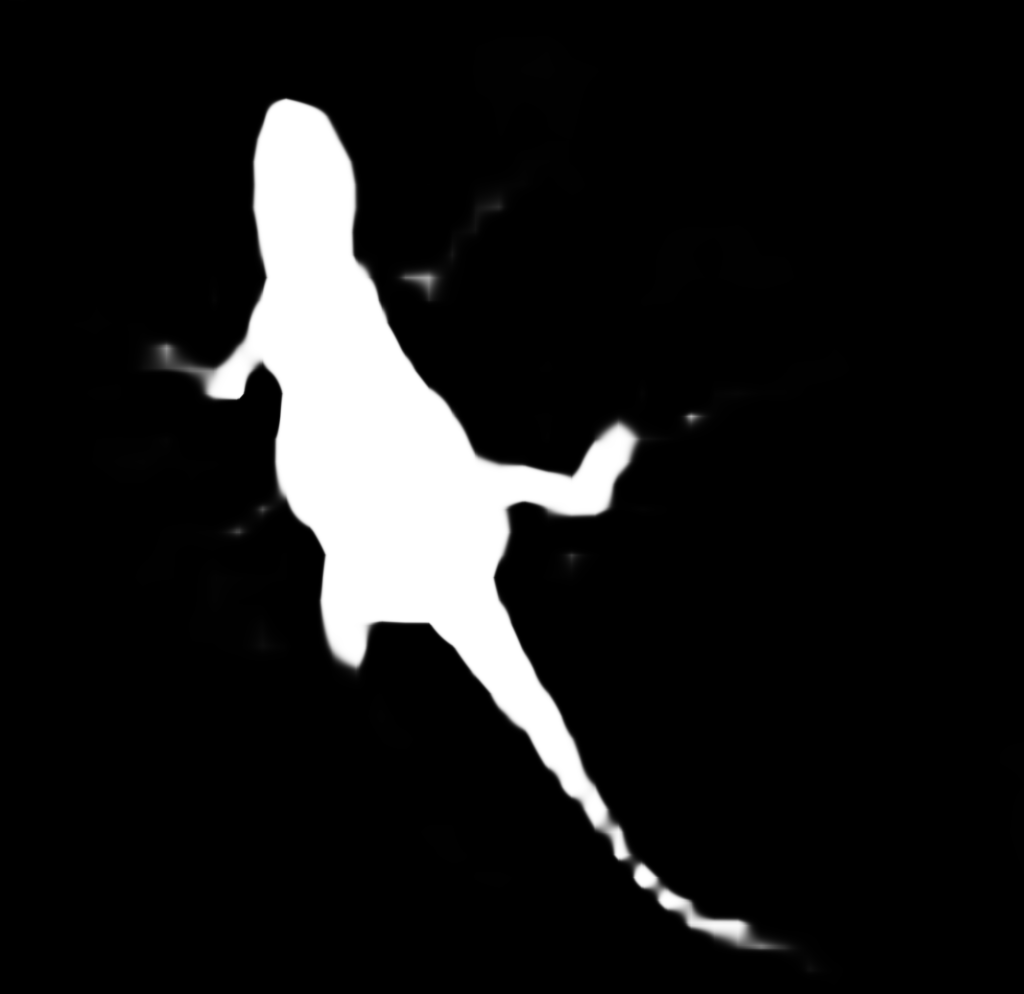}} &
  {\includegraphics[width=0.105\linewidth]{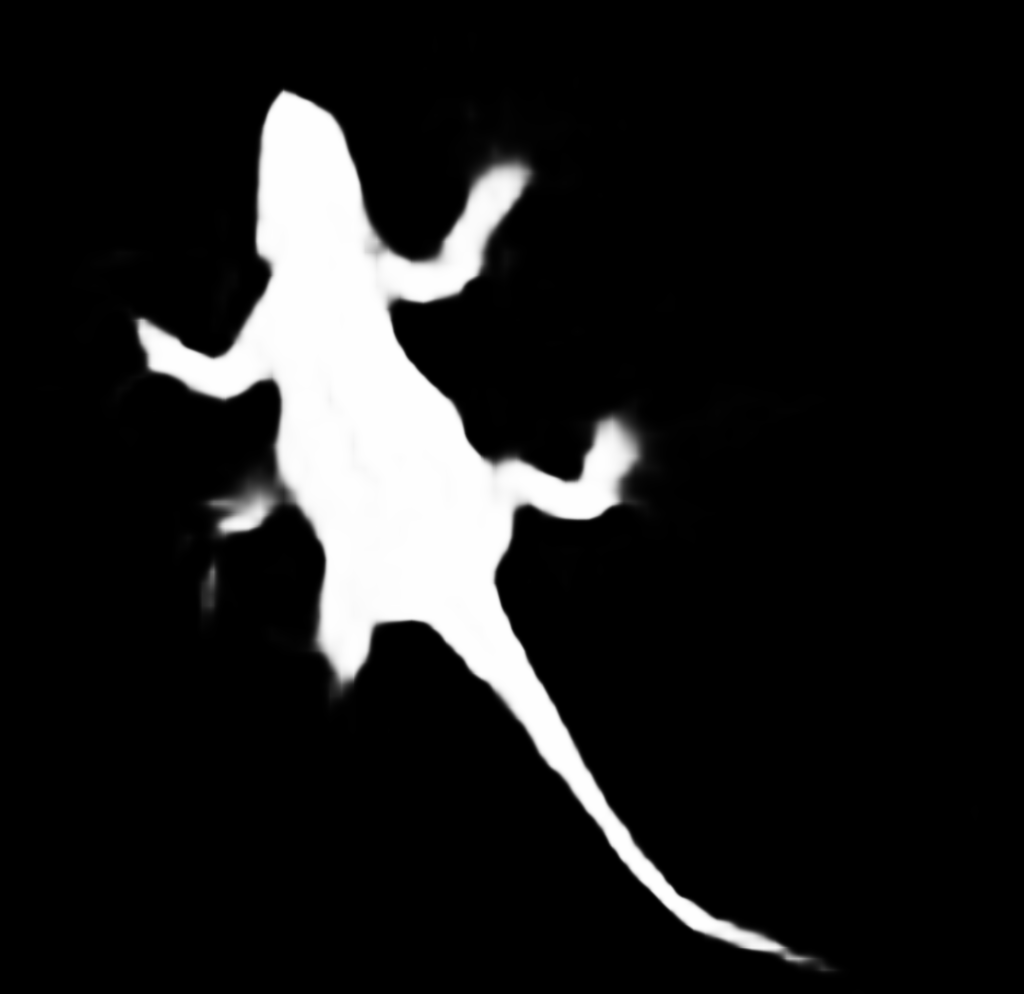}} &
  {\includegraphics[width=0.105\linewidth]{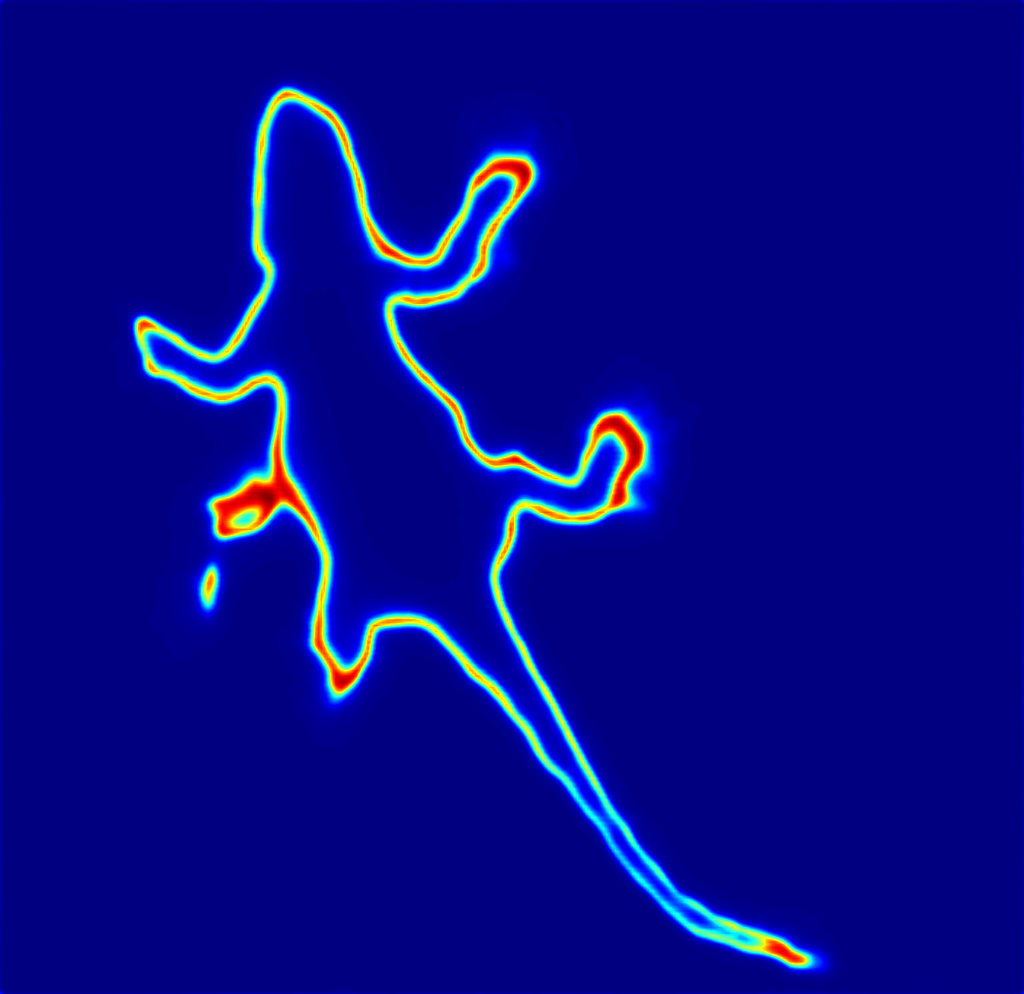}}\\
     \footnotesize{Image} & \footnotesize{Ground Truth} & \footnotesize{BASNet \cite{basnet_sal}} & \footnotesize{CPD \cite{cpd_sal}} & \footnotesize{F3Net \cite{wei2020f3net}} & \footnotesize{SCRN \cite{scrn_sal}} & \footnotesize{SINet \cite{fan2020camouflaged}} & \footnotesize{Ours}& \footnotesize{Confidence} \\
  \end{tabular}
  \end{center}
    \caption{Predictions of our method and those compared methods.}
    \label{fig:visual_comparison}
\end{figure*}

\begin{table*}[t!]
  \centering
  \scriptsize
  \renewcommand{\arraystretch}{1.3}
  \renewcommand{\tabcolsep}{1.1mm}
  \caption{Performance comparison of ablation study models.}
  \begin{tabular}{l|cccc|cccc|cccc|cccc}
  \hline
  &\multicolumn{4}{c|}{CAMO~\cite{le2019anabranch}}&\multicolumn{4}{c|}{CHAMELEON~\cite{Chameleon2018}}&\multicolumn{4}{c|}{COD10K~\cite{fan2020camouflaged}}&\multicolumn{4}{c}{NC4K~\cite{yunqiu_cod21}} \\
    Method & $S_{\alpha}\uparrow$&$F_{\beta}^{\mathrm{mean}}\uparrow$&$E_{\xi}^{\mathrm{mean}}\uparrow$&$\mathcal{M}\downarrow$& $S_{\alpha}\uparrow$&$F_{\beta}^{\mathrm{mean}}\uparrow$&$E_{\xi}^{\mathrm{mean}}\uparrow$&$\mathcal{M}\downarrow$ &  $S_{\alpha}\uparrow$ & $F_{\beta}^{\mathrm{mean}}\uparrow$ & $E_{\xi}^{\mathrm{mean}}\uparrow$ & $\mathcal{M}\downarrow$ & $S_{\alpha}\uparrow$
    & $F_{\beta}^{\mathbf{\mathrm{mean}}}\uparrow$ & $E_{\xi}^{\mathbf{\mathrm{mean}}}\uparrow$ & $\mathcal{M}\downarrow$  \\
  \hline
  M1 & 0.780 & 0.751 & 0.858 & 0.080 & 0.862 & 0.794 & 0.918 & 0.031 & 0.791 & 0.667 & 0.864 & 0.037  & 0.828 & 0.778 & 0.893 & 0.048  \\ 
  M2 & 0.794 & 0.767 & 0.859 & 0.076 & 0.881 & 0.819 & 0.926 & 0.031 & 0.808 & 0.700 & 0.881 & 0.036  & 0.839 & 0.802 & 0.900 & 0.047   \\ 
  M3 & 0.799 & 0.770 & 0.865 & 0.075 & 0.885 & 0.831 & 0.940 & 0.029 & 0.809 & 0.703 & 0.885 & 0.035  & 0.842 & 0.803 & 0.904 & 0.047   \\ 
    \hline
  \end{tabular}
  \label{tab:ablation_study_model_comparison}
\end{table*}

\subsection{Performance comparison}
There are
limited camouflaged object detection methods \cite{le2019anabranch,yan2020mirrornet} due to the unavailability of a large training dataset prior to the release of COD10K \cite{fan2020camouflaged}.
Considering the similarity in the settings of camouflaged object detection and salient object detection, we consider state-of-the-art salient object detection methods \cite{cpd_sal, scrn_sal, Liu19PoolNet, wei2020f3net, zhou2020interactive, basnet_sal, zhao2019EGNet, feng2020residual, chen2020reverse} and retrain those models on the COD10K training dataset for performance comparison. Note that, SINet \cite{fan2020camouflaged} and LSR \cite{yunqiu_cod21} are camouflaged object detection models, we report their performance directly without retraining.

\noindent\textbf{Quantitative comparison:}
We show performance of the
compared methods Tab.~\ref{tab: Benchmark model comparison}. It can be seen that our proposed confidence-aware camouflaged object detection network 
compares favourably against the previous state-of-the-art methods on all four datasets. The improvements over SINet \cite{fan2020camouflaged} are most significant on Mean Absolute Error evaluation, which ranges between $14.7-18.5\%$ on the four datasets. Among the salient object detection methods, RASNet \cite{chen2020reverse} obtains comparative performance with SINet \cite{fan2020camouflaged} although it is not designed specifically for the camouflaged object detection task. However, it is still outperformed by our proposed method on all evaluation metrics. We notice the relatively similar S-measure and mean F-measure of our solution compared with LSR \cite{yunqiu_cod21} on the CHAMELEON \cite{Chameleon2018} dataset. This mainly due to the small size of the CHAMELEON \cite{Chameleon2018} dataset, with 76 samples in total. The performance gap on MAE further indicates effectiveness of our solution.  

\noindent\textbf{Qualitative comparison:} We show predictions of our method and compared methods in Fig.~\ref{fig:visual_comparison}. In the first and second rows, \cite{basnet_sal, cpd_sal, wei2020f3net, fan2020camouflaged} fail to recover the main structure of the \textit{Batfish} and \textit{Ghost Pipefish}. \cite{scrn_sal} can only discover the main body while predictions around the object boundary are ambiguous and incorrect. On the contrary, our method is able to segment more complete camouflaged objects whose boundaries are closer to those of ground truths. On the third row, \cite{basnet_sal, cpd_sal, wei2020f3net, fan2020camouflaged, scrn_sal} recover the main body of the \textit{Lizard}, but they fail to find the limbs. In comparison, our method successfully segments both the main body and the four feet, both of which are close to the ground truth. Complementing to our camouflaged object detection, our estimated confidence map picks up inaccurate predictions associated with both over-segmentation and under-segmentation issues at the object boundary. 

\noindent\textbf{Inference time comparison:}
Different from the compared methods, which produce a single camouflage map in the end, we introduce a confidence estimation module to evaluate pixel-wise awareness of model of the predictions. Although the extra module is included in our framework, our inference time is 0.0211s per image for the camouflaged object detection network and 0.0094s per image for the confidence estimation network, which is comparable with existing techniques, such as 0.0296s of SINet \cite{fan2020camouflaged}.

\subsection{Ablation study}
We have two main modules in our confidence-aware camouflaged object detection network, namely a camouflaged object detection network and a confidence estimation network. We perform the following ablation study to examine the contribution of the main components of our framework. 
We show performance of these models
in Tab.~\ref{tab:ablation_study_model_comparison}.

\noindent\textbf{The structure of the camouflaged object detection network}: We adopt the holistic attention module in \cite{cpd_sal} to refine the module prediction with the initial prediction as attention. To test how our model performs without the holistic attention module, we train the camouflaged object detection network with only the initial prediction as output and denote it as \enquote{M1}. Further, we add the holistic attention module to \enquote{M1} and obtain \enquote{M2}. Tab.~\ref{tab:ablation_study_model_comparison} shows that \enquote{M2} consistently improves over \enquote{M1} on all evaluation metrics, demonstrating that the holistic attention is able to help the model extract more useful features.

\noindent\textbf{Joint training of the camouflaged object detection network and the confidence estimation network for confidence-aware learning:
} We add confidence estimation network with confidence-aware learning to \enquote{M2}, and use the confidence-aware structure loss in Eq.~\ref{eqn: Uncertainty-Aware Structure Loss} to train it, which is then our final model in Tab.~\ref{tab: Benchmark model comparison} \enquote{Ours}.
The improved performance compared with \enquote{M2} demonstrates that our confidence estimation is able to direct the focus of the camouflaged object detection network to correct
false predictions on hard pixels..


\begin{figure}[tp]
  \begin{center}
  \begin{tabular}{{c@{ } c@{ } c@{ } c@{ } c@{ }}}
  {\includegraphics[width=0.19\linewidth]{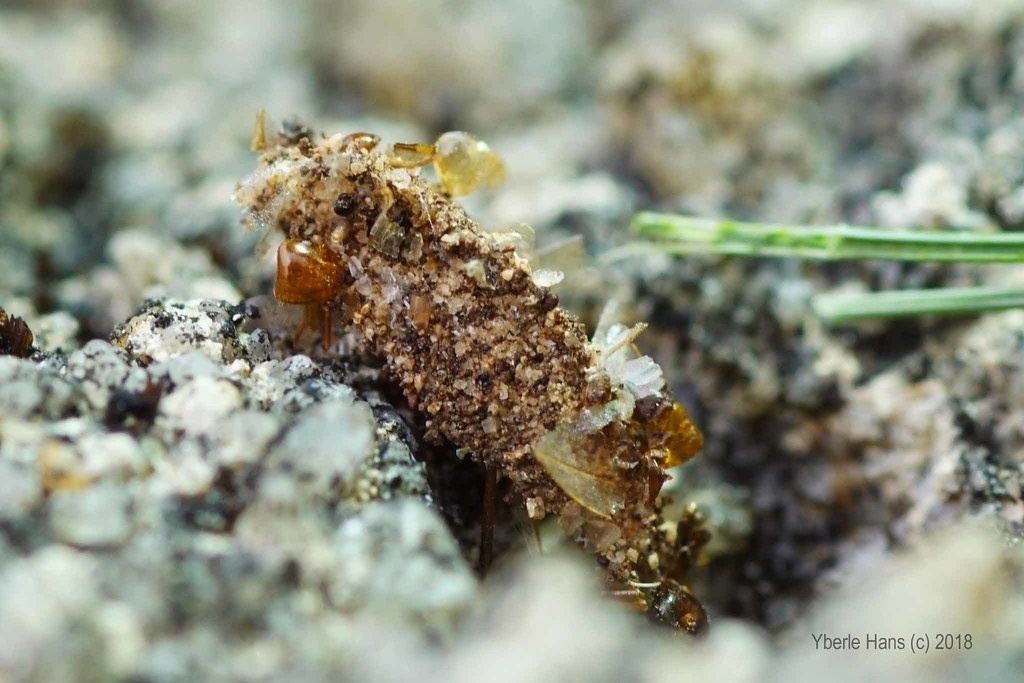}}&
     {\includegraphics[width=0.19\linewidth]{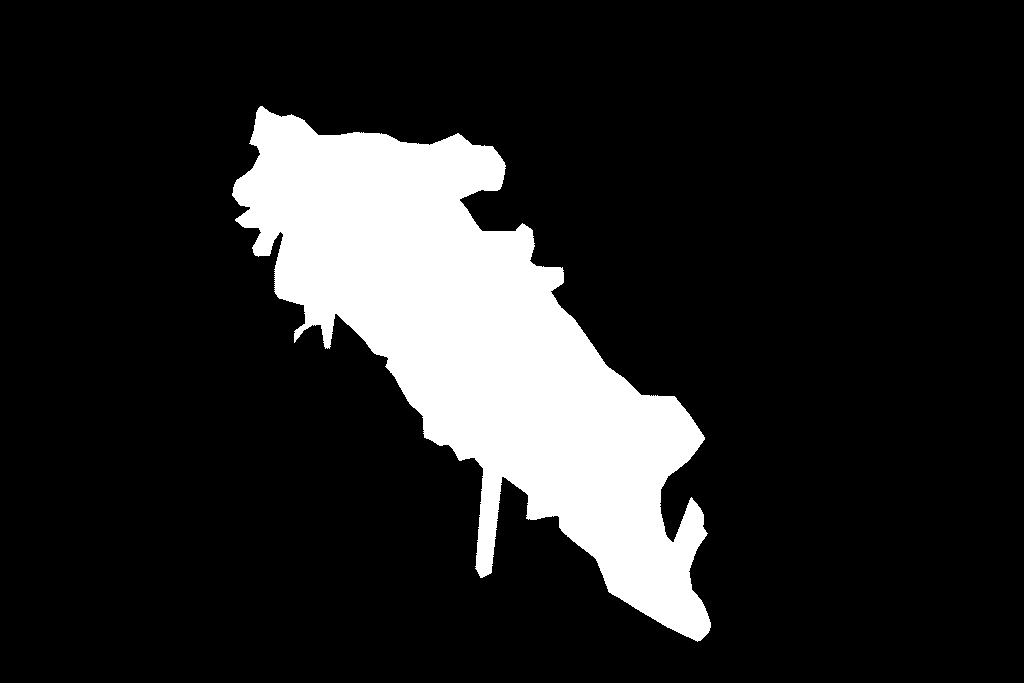}}&
     {\includegraphics[width=0.19\linewidth]{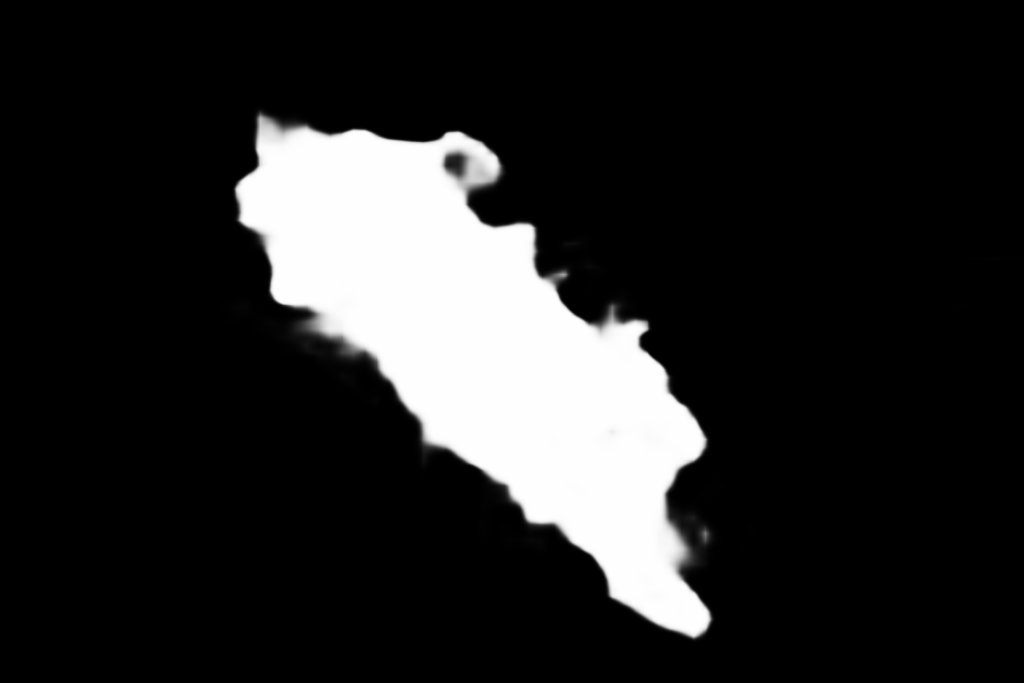}}&
     {\includegraphics[width=0.19\linewidth]{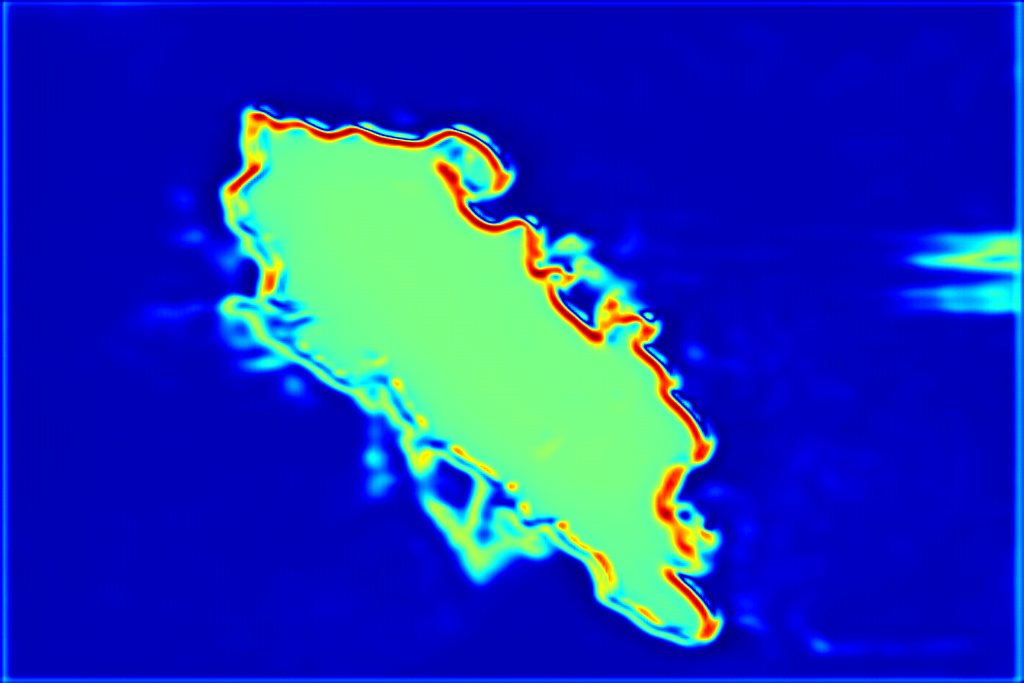}}&
     {\includegraphics[width=0.19\linewidth]{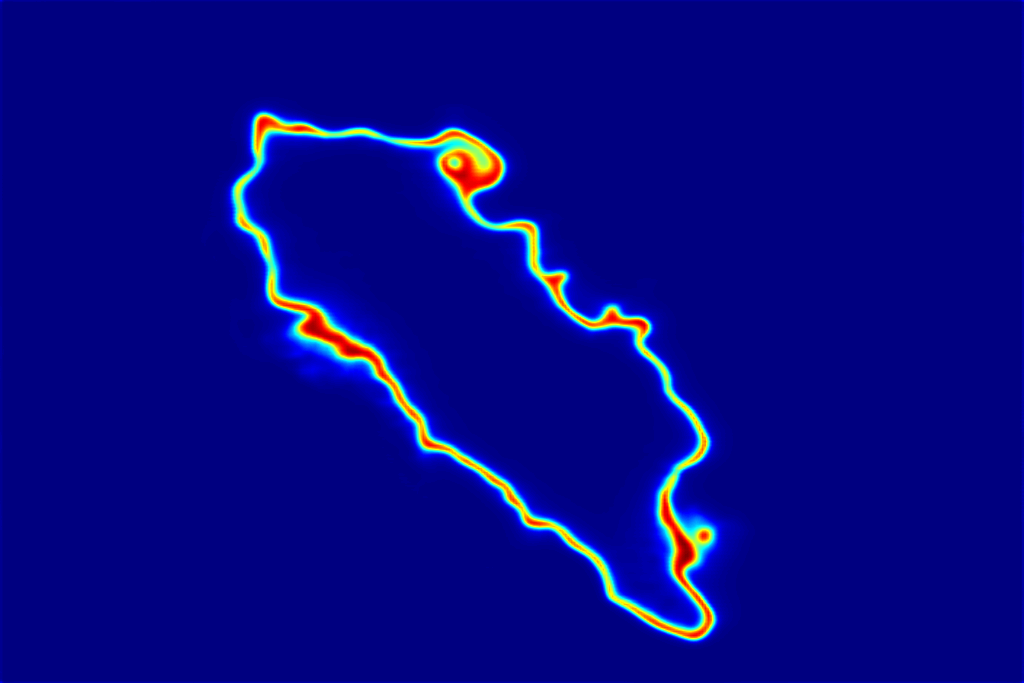}} \\
  {\includegraphics[width=0.19\linewidth]{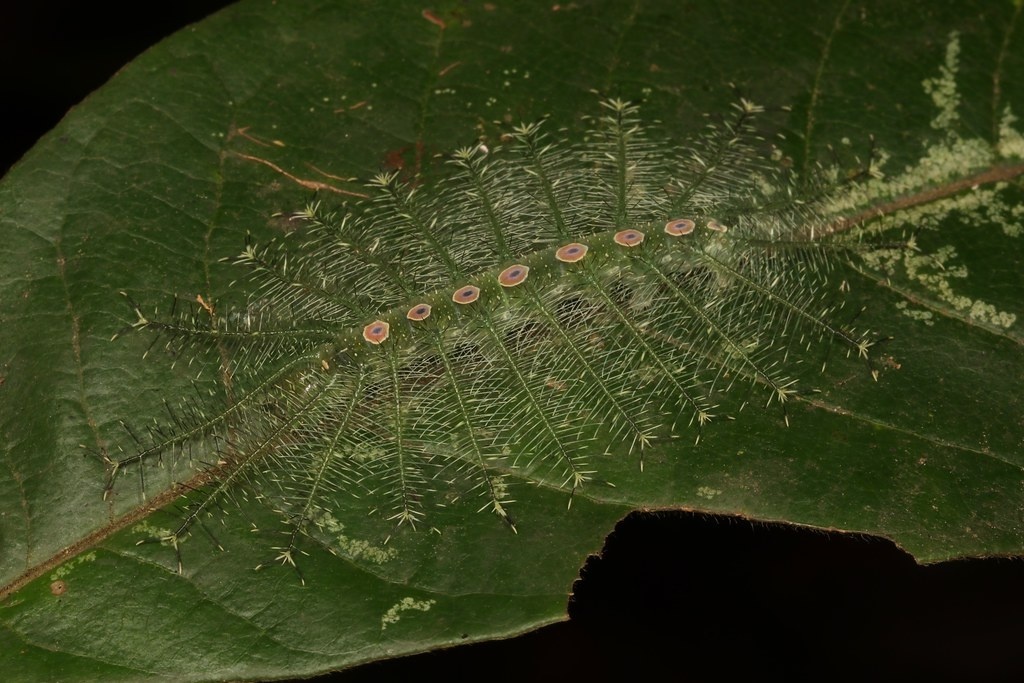}}&
     {\includegraphics[width=0.19\linewidth]{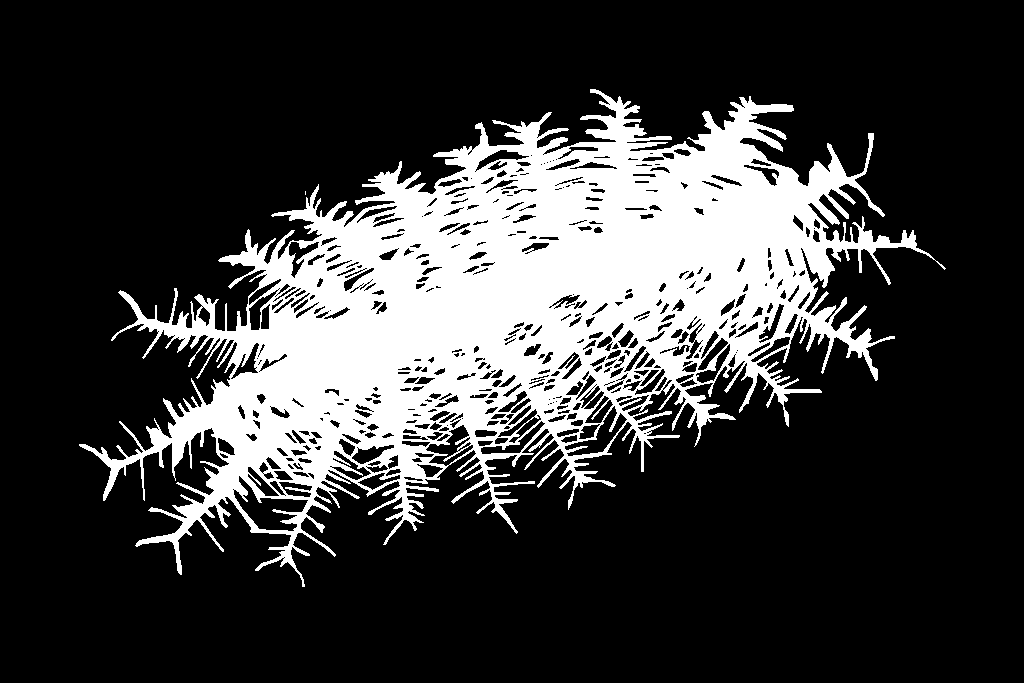}}&
     {\includegraphics[width=0.19\linewidth]{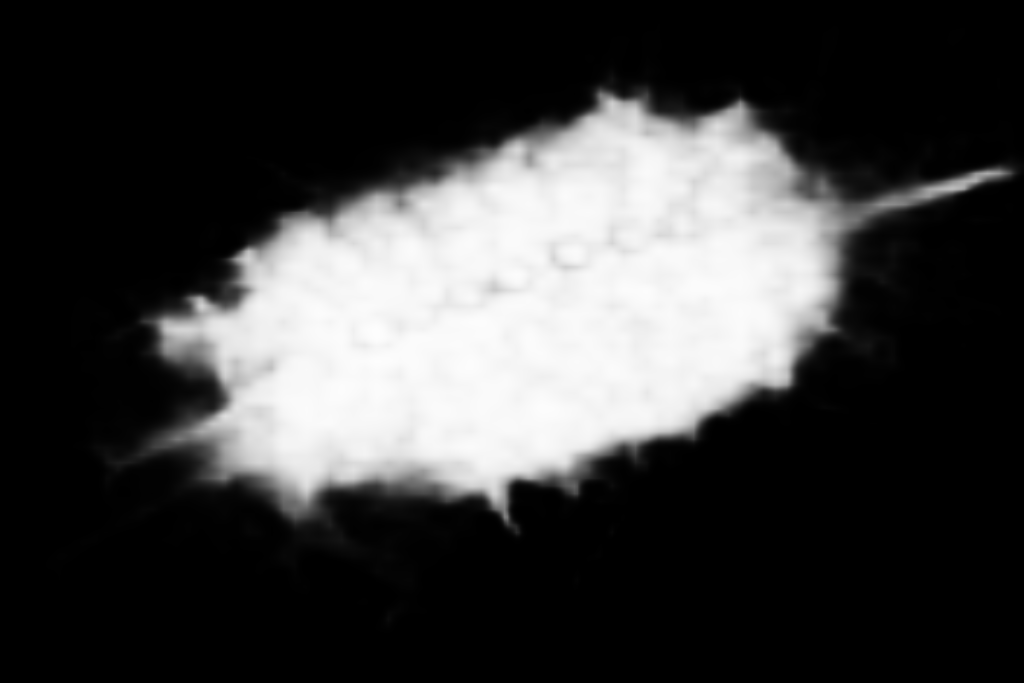}}&
     {\includegraphics[width=0.19\linewidth]{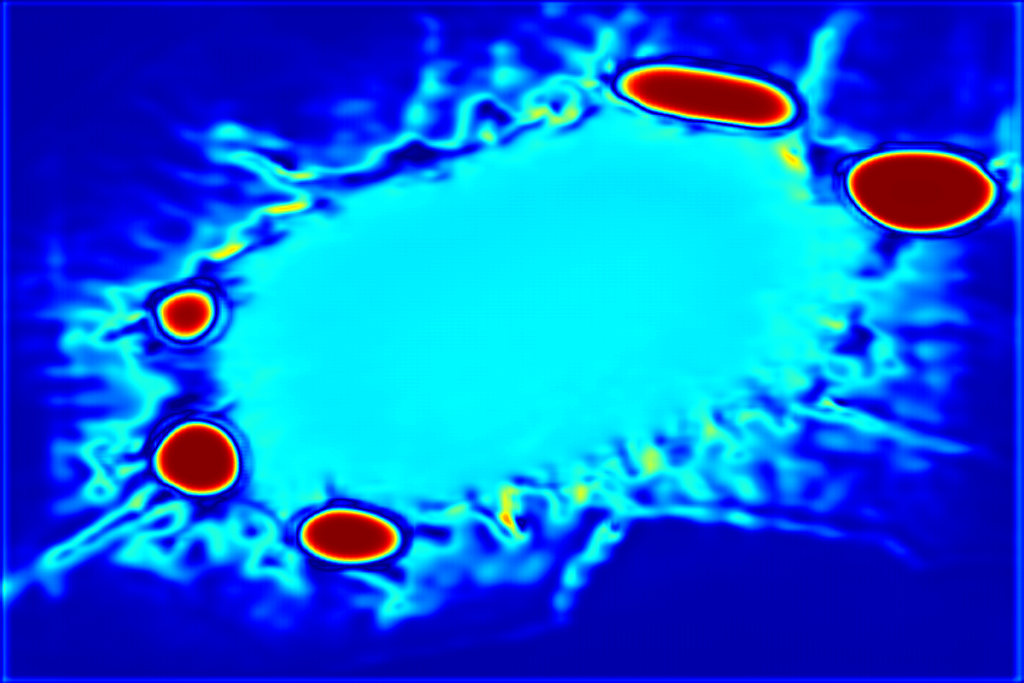}}&
     {\includegraphics[width=0.19\linewidth]{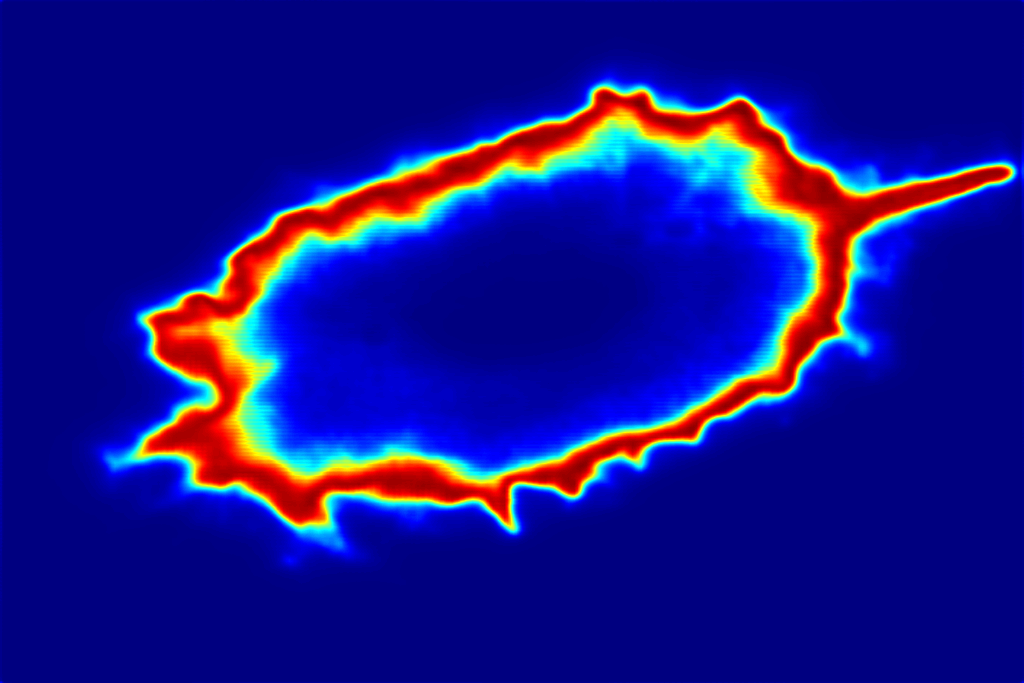}} \\
  {\includegraphics[width=0.19\linewidth]{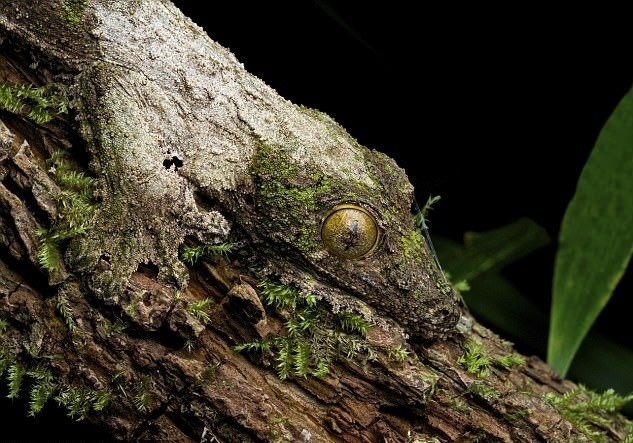}}&
     {\includegraphics[width=0.19\linewidth]{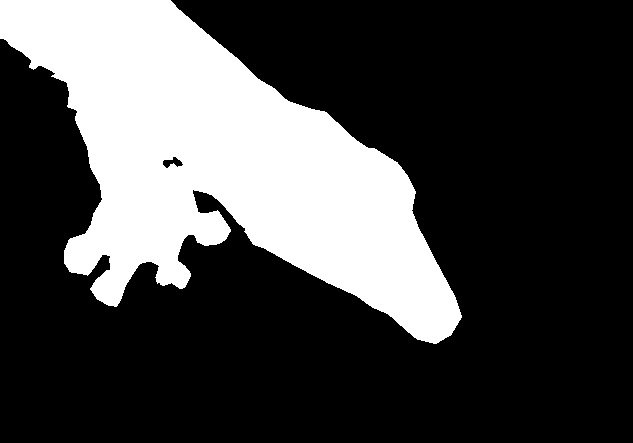}}&
     {\includegraphics[width=0.19\linewidth]{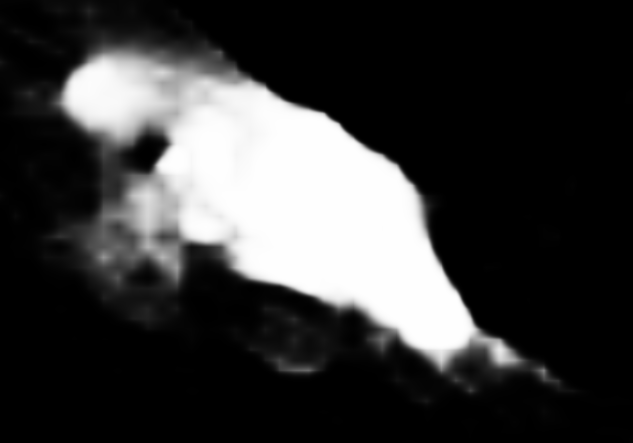}}&
     {\includegraphics[width=0.19\linewidth]{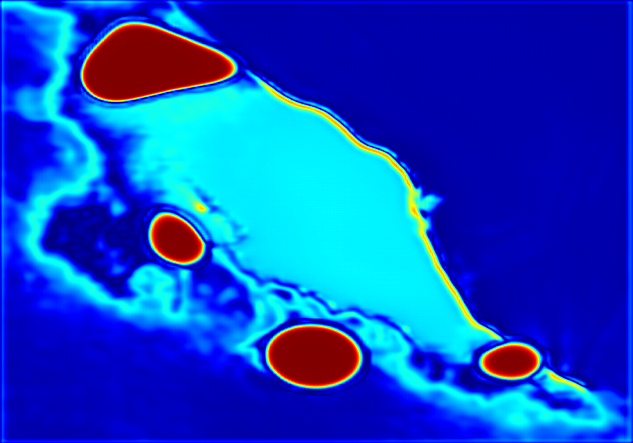}}&
     {\includegraphics[width=0.19\linewidth]{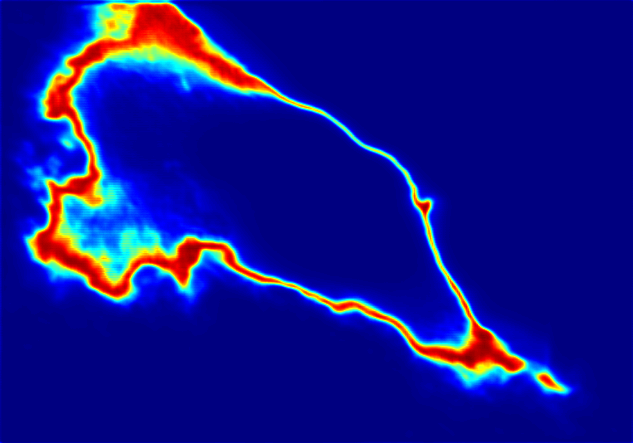}} \\
  \end{tabular}
  \end{center}
    \caption{Comparison of confidence maps produced with dynamic supervision and adversarial learning setting.
    From left to right are image, ground truth map, model prediction, confidence with adversarial learning and confidence with our dynamic supervision.}
    \label{fig: confidence map comparisons}
\end{figure}

\noindent\textbf{The supervision of the confidence estimation network:} Similar to \cite{Hung_semiseg_2018, nie2019difficulty}, another option to generate supervision for the confidence estimation module is to assign $0$ for the prediction and $1$ for the ground truth map following the adversarial learning pipeline. We perform this experiment and change the loss of the confidence estimation module to:
\begin{equation}
\begin{aligned}
    \mathcal{L}_{c}' = & 0.5*(\mathcal{L}_{ce} (g_\beta(\amalg(x, f^{ini}_\theta(x))), 0) + \\ &\mathcal{L}_{ce} (g_\beta(\amalg(x, f^{ref}_\theta(x))), 0)) + \\ &\mathcal{L}_{ce}(g_\beta(\amalg(x, y)), 1),
\end{aligned}
\end{equation}
where $\mathcal{L}_{ce}$ is a binary cross-entropy loss and $\amalg(\cdot)$ is a concatenation operation\footnote{Please refer to the supplementary material for the network overview adopting adversarial learning settings to train the confidence estimation network.}.
In this setting, we regard our confidence estimation network as a discriminator and our camouflaged object detection network as a generator. A well-trained discriminator should output 0.5 for output of a well-trained generator that is infinitely close to the ground truth, indicating that it cannot distinguish the prediction from the ground truth. Therefore, we define the estimated confidence in this adversarial learning setting as shown in Eq.~\ref{eqn: confidence estimation in adversarial learning setting}:
\begin{equation}
\begin{aligned}
    \hat{y}_{c} = \frac{|g_{\beta} (\amalg (x, f_{\theta}(x)))-0.5|}{0.5},
\end{aligned}
\label{eqn: confidence estimation in adversarial learning setting}
\end{equation}
where the uncertainty is measured as the distance between 0.5 and the output of the discriminator. The closer the output is to 0.5, the lower the uncertainty is. To prevent the discriminator from finding a trivial solution, which simply projects the ground-truth values $y \in \{0, 1\}$ to 1 and the predicted values $f_{\theta}(x) \in (0, 1)$ to 0, we introduce a label perturbation technique which relaxes the ground-truth labels from $\{0, 1\}$ to  $\{v \; | \; 0 < v < 0.01 \; \text{or } 0.99 < v < 1\}$ corresponding to the background label and the foreground label respectively. Confident correct predictions on pixels of the camouflaged object detection network are associated with values between these ranges, resulting in estimating high confidences in these pixels, while pixels with moderate scores, \eg 0.4 for weak background prediction or 0.6 for foreground prediction, are assigned high uncertainties.

\begin{table*}[t!]
  \centering
  \scriptsize
  \renewcommand{\arraystretch}{1.3}
  \renewcommand{\tabcolsep}{1.0mm}
  \caption{Performance comparison of choosing different $\lambda$ values.}
  \begin{tabular}{l|cccc|cccc|cccc|cccc}
  \hline
  &\multicolumn{4}{c|}{CAMO~\cite{le2019anabranch}}&\multicolumn{4}{c|}{CHAMELEON~\cite{Chameleon2018}}&\multicolumn{4}{c|}{COD10K~\cite{fan2020camouflaged}}&\multicolumn{4}{c}{NC4K~\cite{yunqiu_cod21}} \\
    Method & $S_{\alpha}\uparrow$&$F_{\beta}^{\mathrm{mean}}\uparrow$&$E_{\xi}^{\mathrm{mean}}\uparrow$&$\mathcal{M}\downarrow$& $S_{\alpha}\uparrow$&$F_{\beta}^{\mathrm{mean}}\uparrow$&$E_{\xi}^{\mathrm{mean}}\uparrow$&$\mathcal{M}\downarrow$ &  $S_{\alpha}\uparrow$ & $F_{\beta}^{\mathrm{mean}}\uparrow$ & $E_{\xi}^{\mathrm{mean}}\uparrow$ & $\mathcal{M}\downarrow$ & $S_{\alpha}\uparrow$
    & $F_{\beta}^{\mathbf{\mathrm{mean}}}\uparrow$ & $E_{\xi}^{\mathbf{\mathrm{mean}}}\uparrow$ & $\mathcal{M}\downarrow$  \\
  \hline
  $\lambda = 3$ & 0.758 & 0.728 & 0.823 & 0.087 & 0.861 & 0.811 & 0.941 & 0.037 & 0.798 & 0.703 & 0.889 & 0.037 & 0.826 & 0.793 & 0.894 & 0.052  \\ 
  $\lambda = 5$ & 0.776 & 0.749 & 0.848 & 0.080 & 0.874 & 0.805 & 0.939 & 0.035 & 0.798 & 0.692 & 0.879 & 0.038 & 0.831 & 0.792 & 0.897 & 0.050  \\ 
  $\lambda = 10$ & 0.799 & 0.770 & 0.865 & 0.075 & 0.885 & 0.831 & 0.940 & 0.029 & 0.809 & 0.703 & 0.885 & 0.035  & 0.842 & 0.803 & 0.904 & 0.047   \\ 
  $\lambda_{D}$ & 0.804 & 0.764 & 0.862 & 0.074 & 0.875 & 0.810 & 0.923 & 0.034 & 0.807 & 0.697 & 0.877 & 0.037  & 0.843 & 0.798 & 0.900 & 0.047   \\ 
    \hline
  \end{tabular}
  \label{tab: hyperparameter analysis}
\end{table*}

Experiment \enquote{M3} denotes the framework adopting an adversarial learning setting for the confidence estimation. Experimental results in Tab.~\ref{tab:ablation_study_model_comparison} and Tab.~\ref{tab: Benchmark model comparison} show that \enquote{Ours} in general outperforms \enquote{M3}, indicating that confidence estimation network trained with dynamic supervision produces more accurate confidence maps, which in turn guide the camouflaged object detection network to achieve improved performance. Sec.~\ref{Discussion section} compares the confidence estimation with these two methods in detail. 

\begin{figure}[tp]
  \begin{center}
  \begin{tabular}{{c@{ } c@{ } c@{ } c@{ } c@{ }}}
  {\includegraphics[width=0.19\linewidth]{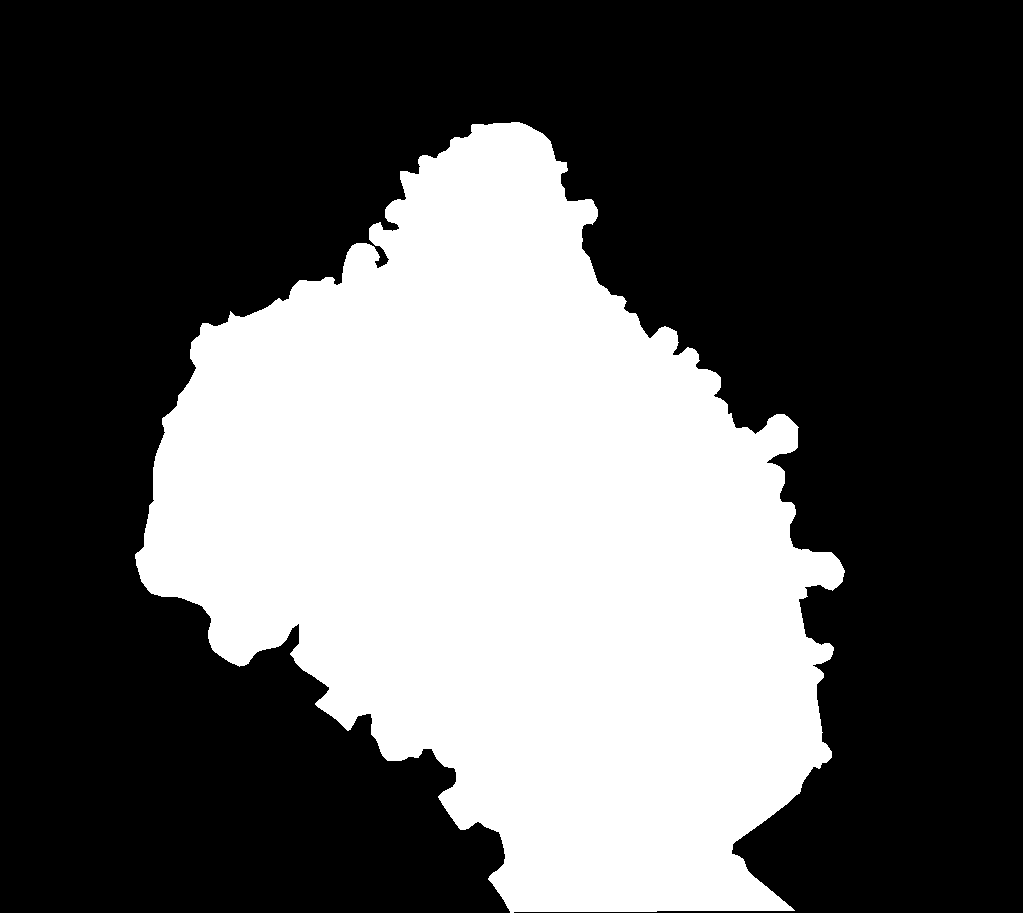}}&
     {\includegraphics[width=0.19\linewidth]{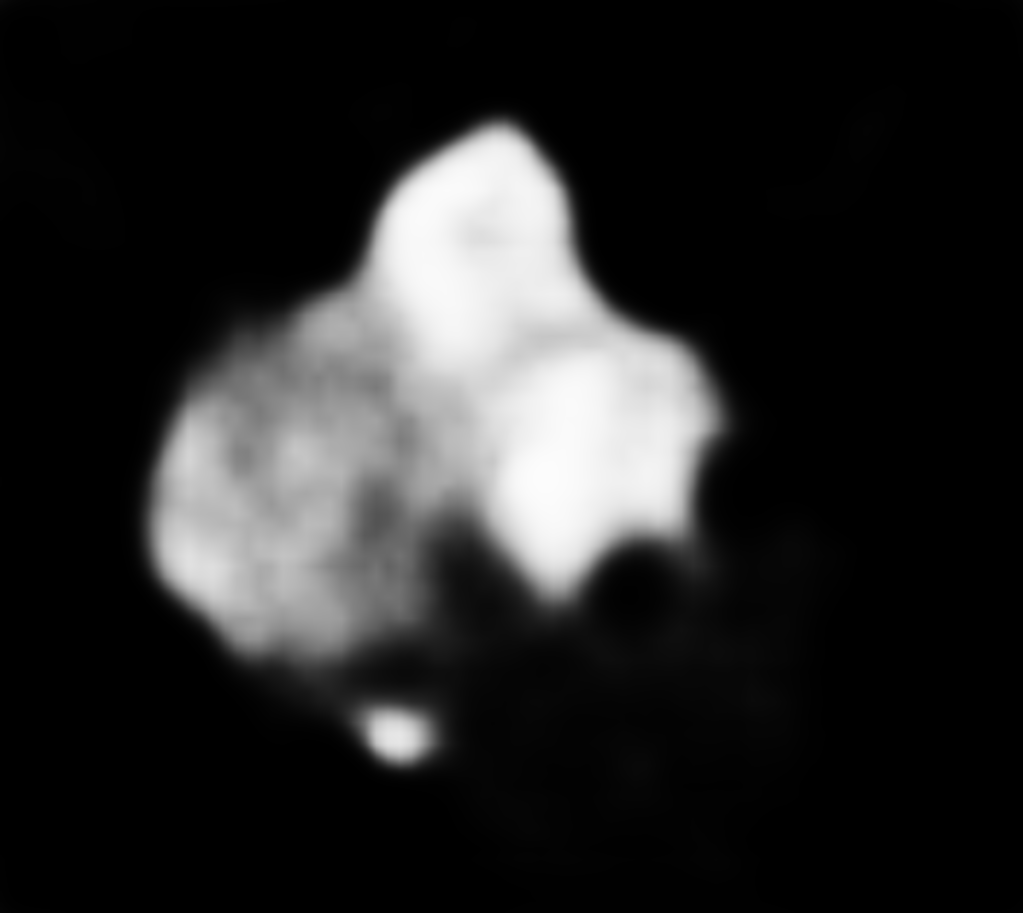}}&
     {\includegraphics[width=0.19\linewidth]{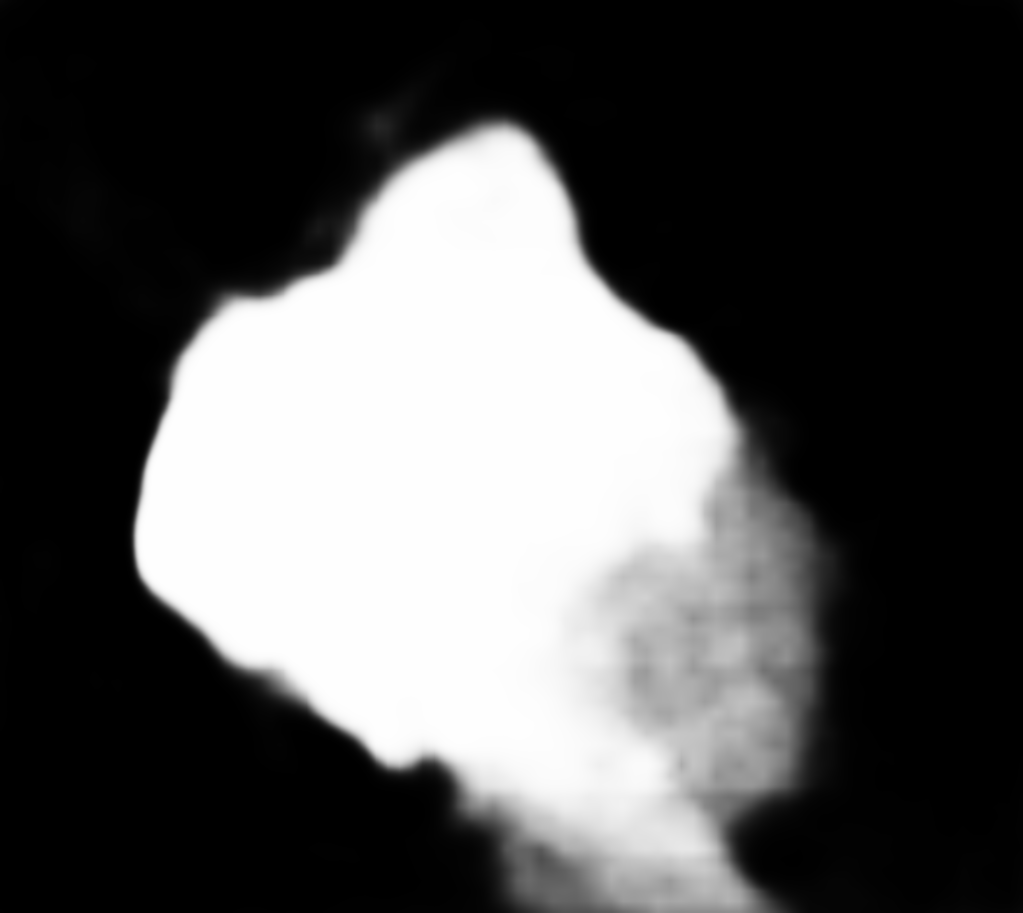}}&
     {\includegraphics[width=0.19\linewidth]{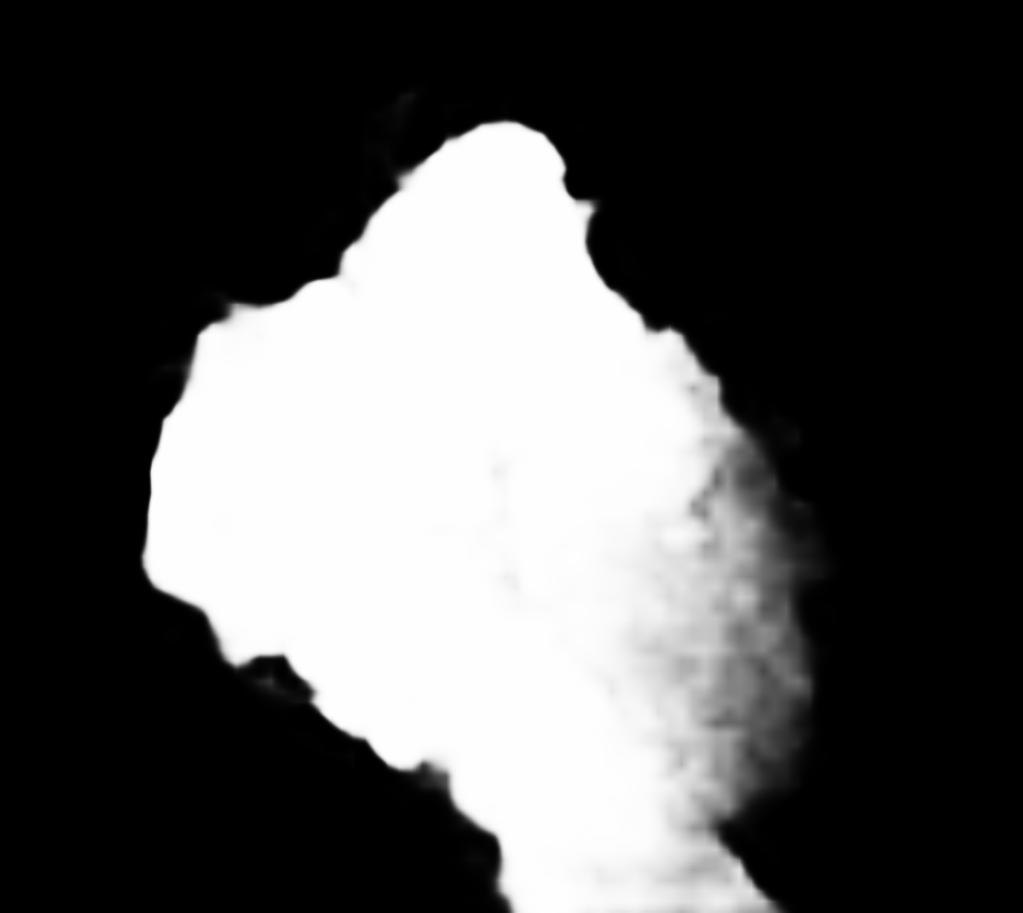}}&
     {\includegraphics[width=0.19\linewidth]{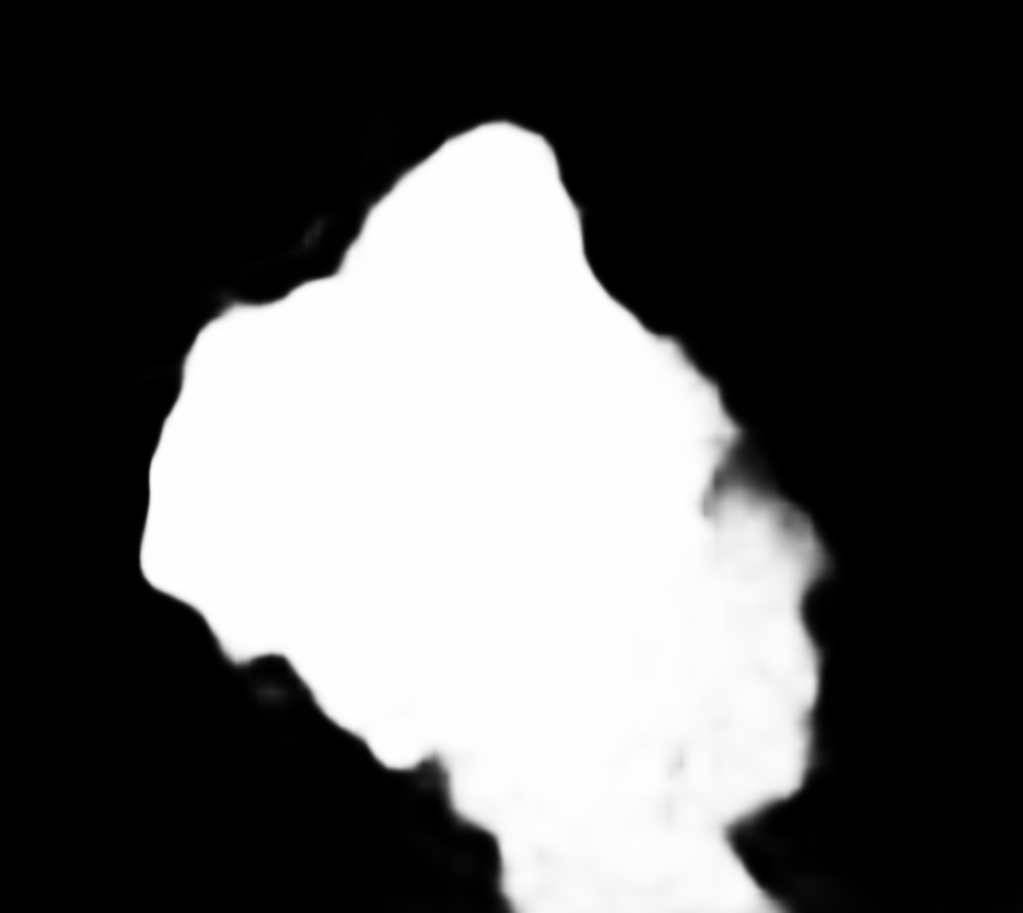}} \\
  {\includegraphics[width=0.19\linewidth]{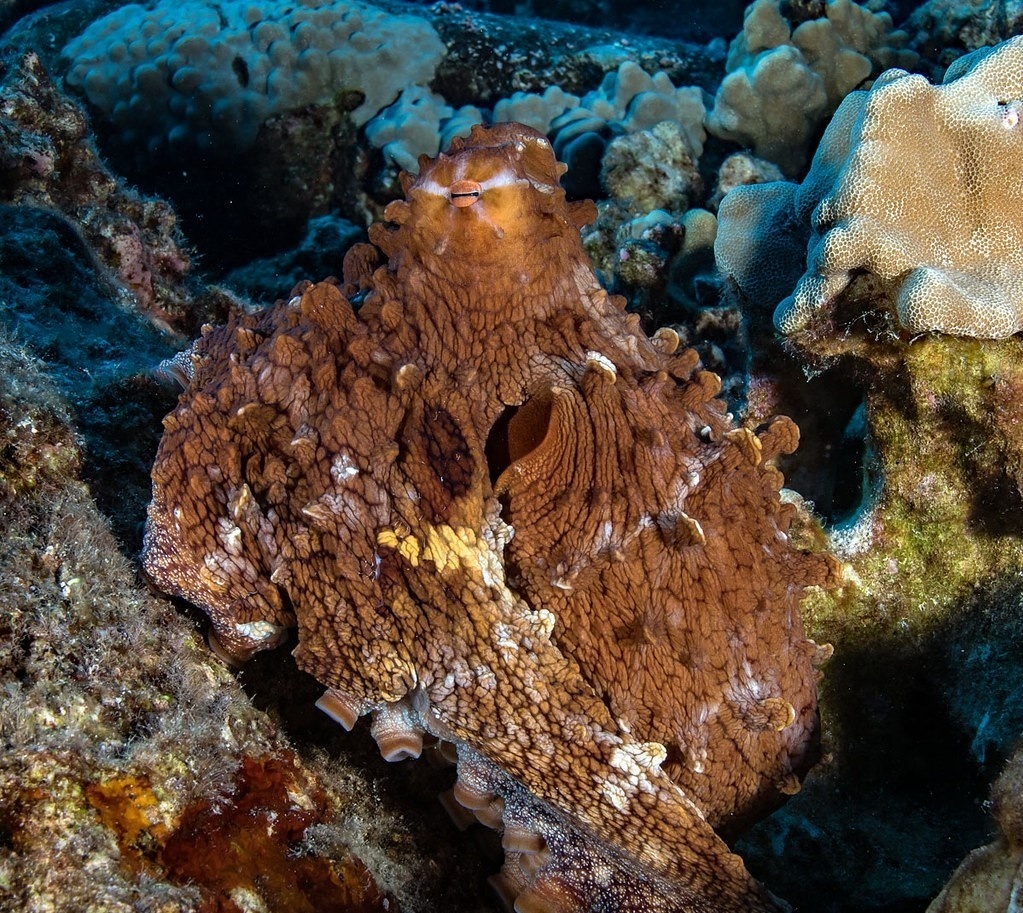}}&
     {\includegraphics[width=0.19\linewidth]{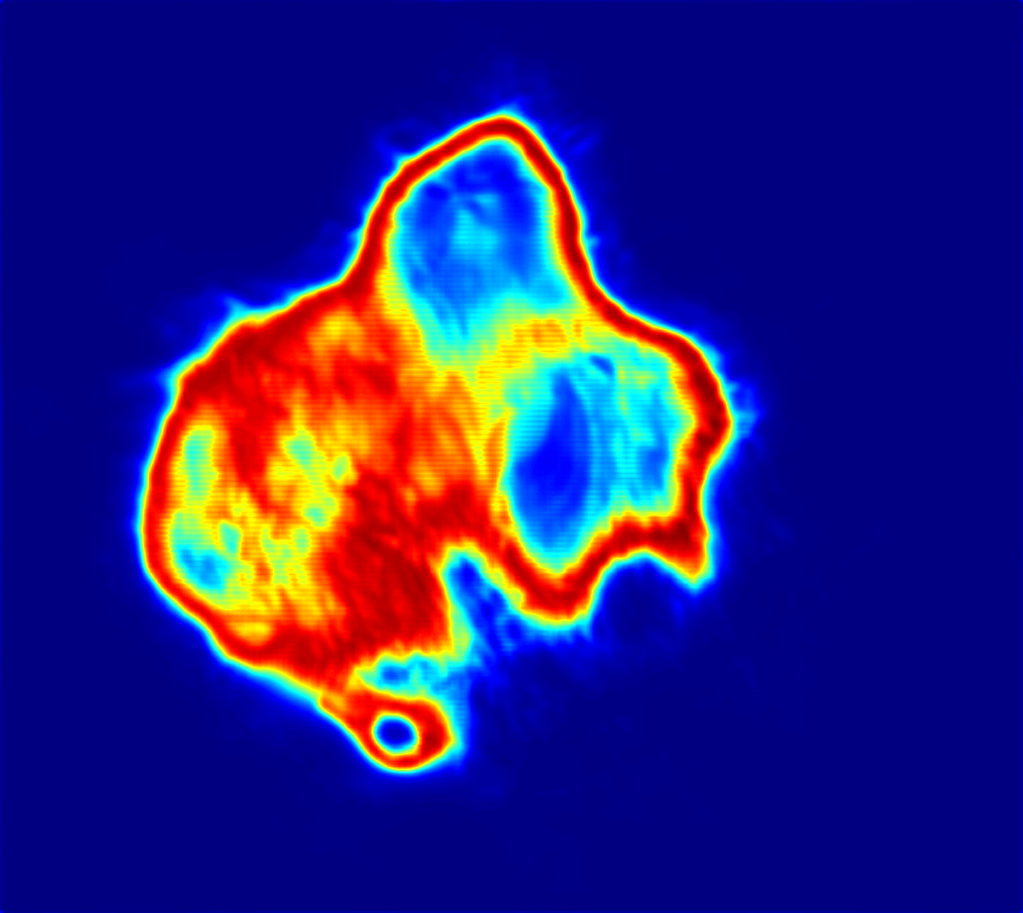}}&
     {\includegraphics[width=0.19\linewidth]{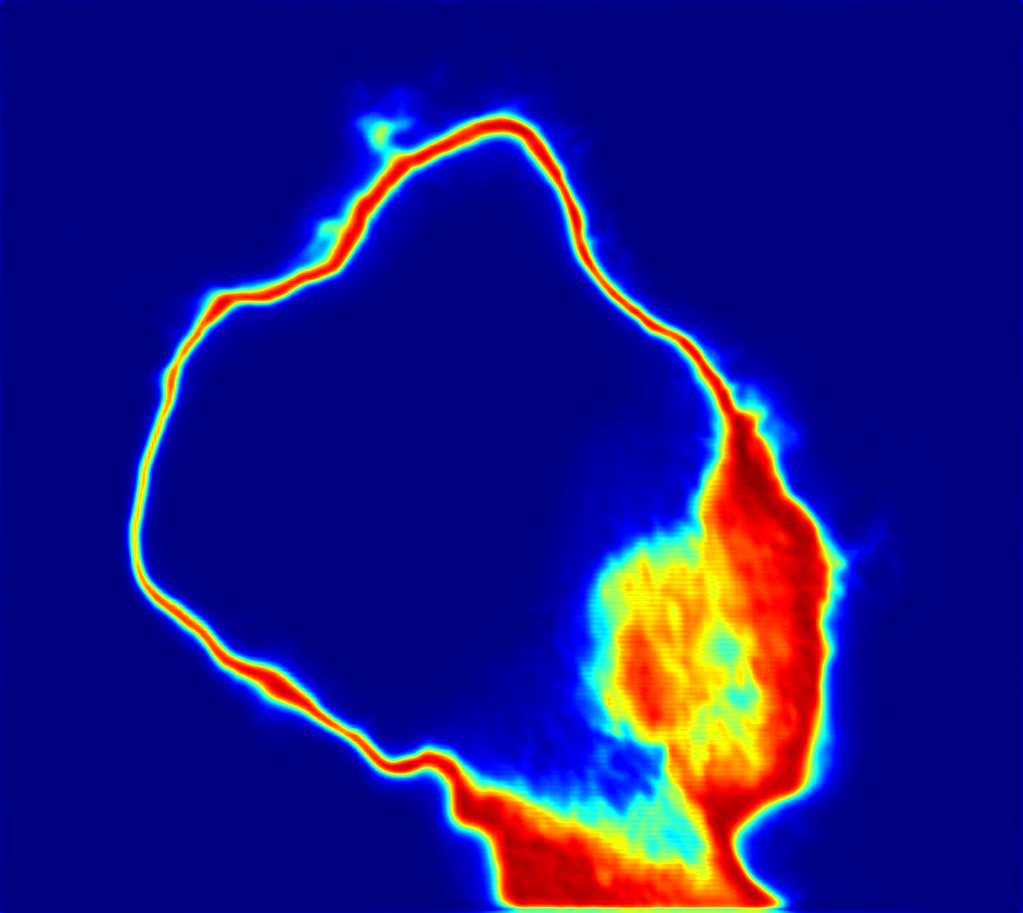}}&
     {\includegraphics[width=0.19\linewidth]{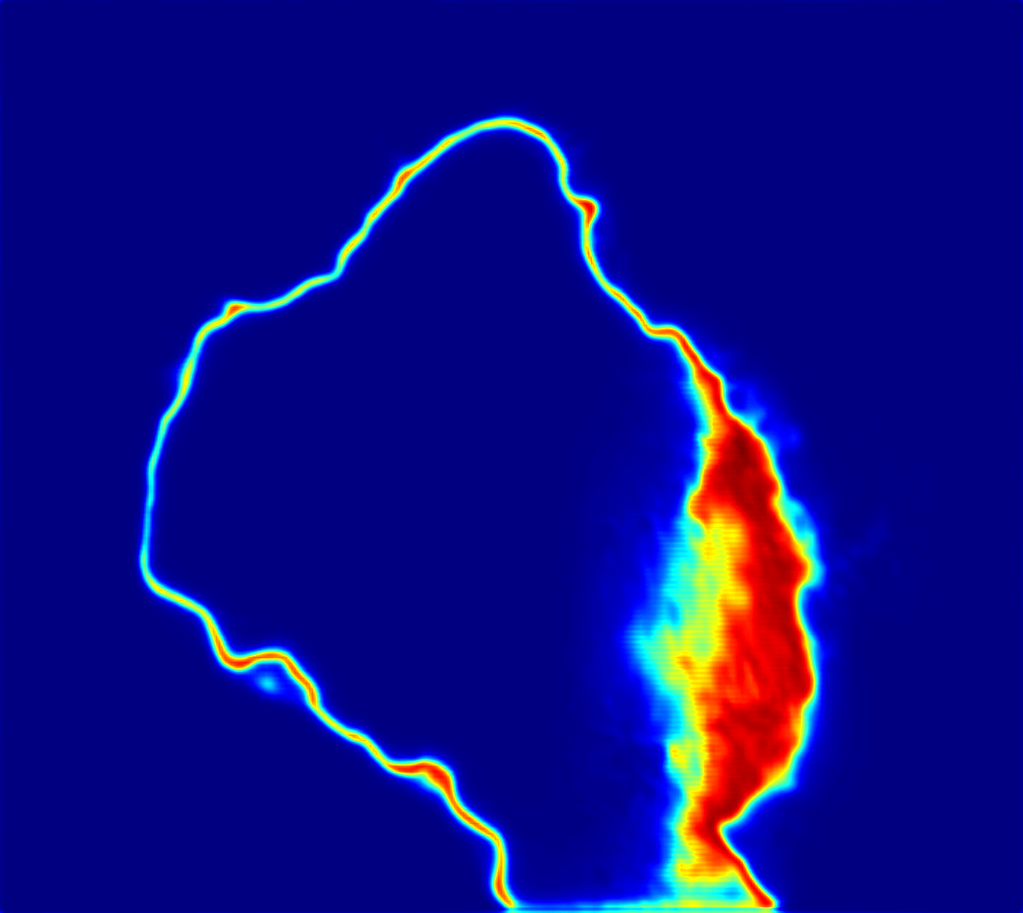}}&
     {\includegraphics[width=0.19\linewidth]{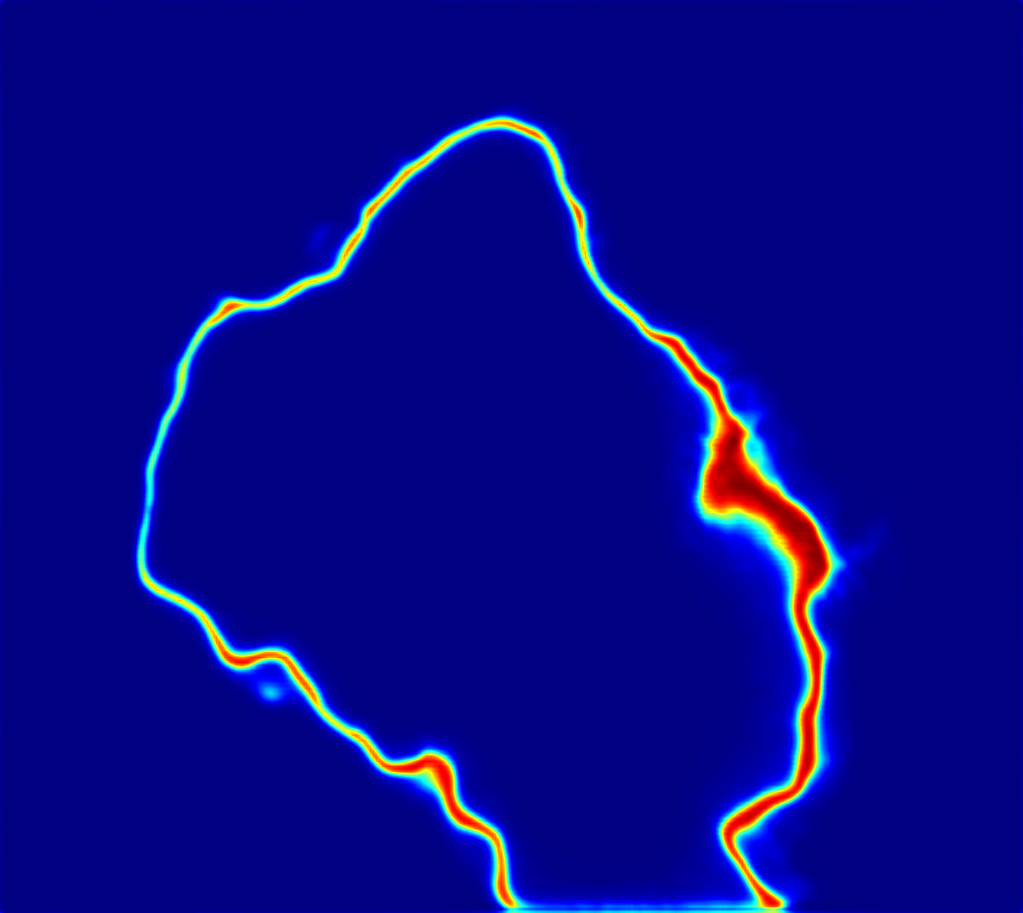}} \\
     \footnotesize{Image/GT} & \footnotesize{t = 2} & \footnotesize{t = 5} & \footnotesize{t = 10} & \footnotesize{t = 20} \\
  \end{tabular}
  \end{center}
    \caption{Using estimated confidence map as an indicator of the prediction quality. 
    The first column displays the ground truth and image. Predictions and corresponding uncertainty maps from different training stages are displayed from the second to the fifty column.
    Red indicates low confidence and blue indicates high confidence. $t$ indicates the training epoch.
    } 
    \label{fig: confidence as evaluation through training}
\end{figure}

\subsection{Discussion}
\label{Discussion section}
\noindent\textbf{Comparison with adversarial learning:}
Fig.~\ref{fig: confidence map comparisons} illustrates the difference between the confidence maps produced with the adversarial learning setting and dynamic supervision method. In general, the confidence map produced with adversarial learning setting is biased to have higher uncertainty values associated with the foreground predictions. These biased uncertainties are consistent across the foreground predictions although they are mostly correct. 
For example, in the samples presented in Fig.~\ref{fig: confidence map comparisons}, the strong foreground predictions inside the object are correct, thus should be regarded as confident. The desired uncertainty values are manifested in the confidence map produced with dynamic supervision. Correct foreground predictions are assigned with high confidence. The errors only occur at the boundary pixels, which correspond to high uncertainty values.

Confidence maps produced with the adversarial learning setting generate artifacts, a blob of uncertain area centred at weak foreground predictions. Although these artifacts locate the uncertain foreground predictions, they fail to provide spatially-accurate uncertainties. On the contrary, confidence map with dynamic supervision is able to delineate a more precise uncertainty structure. 
On the second row of Fig.~\ref{fig: confidence map comparisons}, it produces a thickened uncertainty prediction along the predicted object boundaries where most errors occur. Just inside these boundaries, it faintly traces the thin body parts of the caterpillar at the top-right corner and the left side. The structure-preserving property of the confidence map with dynamic supervision is best demonstrated on the third sample of Fig.~\ref{fig: confidence map comparisons}. Although the camouflaged object detection network fails to predict the webbed frog foot, its structure is picked up by the confidence map produced with dynamic supervision where the boundary of the webbed foot is delineated, forcing the camouflaged object detection network to focus on learning the missing parts.

When the uncertainty map is used as guidance to the structure loss, the adversarial learning version is able to direct attention to weak prediction areas where errors are prone to occur. However, despite its localisation capability, it is not pixel-wise accurate. On the contrary, the dynamic supervision version can 
discover object structure that the camouflaged object detection network fails to find. 
It complements the camouflaged object detection network, refining object structure and recovering initially lost object parts.

\noindent\textbf{Confidence module as a trained evaluation tool without relying on ground truth maps:}
Our confidence map can serve as a rough evaluation tool of the prediction quality of the camouflaged object detection network without relying on the ground-truth segmentation map. Fig.~\ref{fig: confidence as evaluation through training} illustrates the predicted camouflage map and its corresponding confidence map of a sample at different stages of training. The sample is regarded as hard and its initial prediction discovers only a small part of foreground object at the second epoch. This leads to large areas of high uncertainty values in its corresponding confidence map. As the prediction becomes more refined as the training progresses, the high-uncertainty areas in the confidence maps shrink as a result, eventually highlighting only the structures of the camouflaged objects where errors are prone to occur. In addition, Fig.~\ref{fig: confidence as evaluation through training} also validates that our estimated confidence map guides the camouflaged object detection network to gradually recover the initially lost object parts of the hard samples.

\noindent\textbf{Hyper-parameter analysis:}
The impact of selecting different values for $\lambda$, which is a factor controlling the uncertainty guidance in the structure loss, is demonstrated in Tab.~\ref{tab: hyperparameter analysis}. We choose $\lambda = {3, 5, 10, \lambda_{D}}$ where $\lambda_{D}$ is a dynamic factor defined as $\lambda_{D} = \min\{2 \times ReLU(t-5), 20\}$, where $t$ is the current training epoch. The results show that $\lambda = 10$ achieves better performance than $\lambda = {3, 5}$. Although the higher factor introduces a larger distraction in the early stage where the estimated confidence map of an untrained confidence estimation network consists of mainly just random noise, it has little effect on the learning of camouflaged object detection network owing to the sparseness
of the noise in the early epochs. However, with the confidence map becoming more refined as the training progresses, a higher factor is better able to guide the camouflaged object detection network to focus on learning uncertain prediction areas. The inferior performance of the dynamic factor $\lambda_{D}$ can be attributed to that it provides insufficient guidance in the early stage of training. It is difficult to tune it to precisely set the factor at right scale for different stages of training.




\section{Conclusion}
As an effective solution to prevent networks becoming overconfident, confidence learning has attracted more attention recently. In this paper, we propose a confidence-aware camouflaged object detection framework to produce both accurate camouflage prediction and a reasonable confidence map representing model awareness of the prediction. It is composed of an interdependent camouflaged object detection network and confidence estimation network. The dynamic confidence label is generated to train the confidence estimation network, which is derived from the prediction of the camouflaged object detection network and the ground truth map. The estimated confidence map from the confidence estimation network directs the camouflaged object detection network to place more emphasis on learning areas with uncertain predictions. Our proposed network performs favourably against existing camouflaged object detection methods
on four benchmark camouflaged object detection testing datasets.
Further, the generated confidence map provides an effective solution to explain the model prediction without relying on the ground truth map.

{\small
\bibliographystyle{ieee_fullname}
\bibliography{cam_ref}
}

\end{document}